\documentclass[12pt]{iopart}

\usepackage{iopams}
\usepackage{bm,color}  
\usepackage{graphicx}
\usepackage{amssymb}
\usepackage{amsthm}  
\usepackage{color}
\usepackage{graphicx}
\newcommand{\Part}[3]{ \frac{ \partial^{#3} #1 }{ \partial #2^{#3} } }
\newcommand{\V}[1]{\bm{#1} } 

\newcommand{\Ave}[1]{\left\langle {#1} \right\rangle} 

\newcommand{\Extr}[1]{ \mathop{\rm Extr}_{ #1 } }

\newcommand{\mR}{\mathbb{R}}
\newcommand{\mN}{\mathbb{N}}
\newcommand{\lb}{\left(}
\newcommand{\rb}{\right)}
\newcommand{\lbb}{\left\{}
\newcommand{\rbb}{\right\}}
\newcommand{\lsb}{ \left[ }
\newcommand{\rsb}{ \right] }

\newcommand{\T}[1]{\tilde{#1}}
\newcommand{\Req}[1]{(\ref{eq:#1})}
\newcommand{\NReq}[1]{(\ref{eq:#1})}
\newcommand{\Reqs}[2]{(\ref{eq:#1}) and (\ref{eq:#2})}

\newcommand{\Rfig}[1]{Fig.\ \ref{fig:#1}}
\newcommand{\Rfigs}[2]{Figs.\ \ref{fig:#1} and \ref{fig:#2}}
\newcommand{\Rfigss}[2]{Figs.\ \ref{fig:#1}-\ref{fig:#2}}
\newcommand{\Lfig}[1]{\label{fig:#1}}
\newcommand{\Leq}[1]{\label{eq:#1}}
\newcommand{\Rsec}[1]{sec.\ \ref{sec:#1}}
\newcommand{\Lsec}[1]{\label{sec:#1}}
\newcommand{\be}{\begin{eqnarray}}
\newcommand{\ee}{\end{eqnarray}}
\newcommand{\ba}{\begin{array}}
\newcommand{\ea}{\end{array}}
\newcommand{\no}{\nonumber}

\newcommand{\subbe}{\begin{subequations}}
\newcommand{\subee}{\end{subequations}}
\newcommand{\Bs}{\backslash}
\newcommand{\mc}[1]{\mathcal{#1}}

\newcommand{\argmin}{\mathop{\rm arg\, min}\limits}

\newcommand{\MSEx}{\epsilon_{x}}
\newcommand{\MSEy}{\epsilon_{y}}

\newcommand{\LOOE}{\epsilon_{\rm LOO}}
\newcommand{\GE}{\epsilon_{\mathrm{g}}}

\newcommand{\HESS}{\partial^2}

\newcommand{\Wc}{*}
\newcommand{\OBP}{L}
\newcommand{\LOOfactor}{\Theta}
\newcommand{\DM}{A}
\newcommand{\noise}{\Delta}

\begin{document}
\title[CV in sparse linear regression with nonconvex penalties and its acceleration]
{Cross validation in sparse linear regression with piecewise continuous nonconvex penalties and its acceleration}
\author{Tomoyuki Obuchi$^{1}$ and Ayaka Sakata$^{2,3}$ }

\address{$^1$Department of Mathematical \& Computing Science,
Tokyo Institute of Technology,
Ookayama, Meguro-ku, Tokyo 152-8552, Japan}
\address{$^2$Department of Statistical Inference \& Mathematics, The Institute of Statistical Mathematics, Midori-cho, Tachikawa, Tokyo 190-8562, Japan}
\address{$^3$Department of Statistical Science, The Graduate University for Advanced Science (SOKENDAI), Hayama-cho, Kanagawa 240-0193, Japan}
\ead{$^{1}$obuchi@c.titech.ac.jp,$^{2,3}$ayaka@ism.ac.jp}
\address{}
\vspace{10pt}

\begin{abstract}
We investigate the signal reconstruction performance of sparse linear regression in the presence of noise when piecewise continuous nonconvex penalties are used. Among such penalties, we focus on the smoothly clipped absolute deviation (SCAD) penalty. The contributions of this study are three-fold: We first present a theoretical analysis of a typical reconstruction performance, using the replica method, under the assumption that each component of the design matrix is given as an independent and identically distributed (i.i.d.) Gaussian variable. This clarifies the superiority of the SCAD estimator compared with $\ell_1$ in a wide parameter range, although the nonconvex nature of the penalty tends to lead to solution multiplicity in certain regions. This multiplicity is shown to be connected to replica symmetry breaking in the spin-glass theory, and associated phase diagrams are given. We also show that the global minimum of the mean square error between the estimator and the true signal is located in the replica symmetric phase. Second, we develop an approximate formula efficiently computing the cross-validation error without actually conducting the cross-validation, which is also applicable to the non-i.i.d. design matrices. It is shown that this formula is only applicable to the unique solution region and tends to be unstable in the multiple solution region. We implement instability detection procedures, which allows the approximate formula to stand alone and resultantly enables us to draw phase diagrams for any specific dataset. Third, we propose an annealing procedure, called nonconvexity annealing, to obtain the solution path efficiently. Numerical simulations are conducted on simulated datasets to examine these results to verify the consistency of the theoretical results and the efficiency of the approximate formula and nonconvexity annealing. The characteristic behaviour of the annealed solution in the multiple solution region is addressed. Another numerical experiment on a real-world dataset of Type Ia supernovae is conducted; its results are consistent with those of earlier studies using the $\ell_0$ formulation. A MATLAB package of numerical codes implementing the estimation of the solution path using the annealing with respect to $\lambda$ in conjunction with the approximate CV formula and the instability detection routine is distributed in~\cite{obuchi:SLRpackage}.
\end{abstract}

\section{Introduction}
Variable selection problems ubiquitously appear in statistics and machine learning tasks. Although traditional statistical approaches to variable selection work well in principle~\cite{Breiman1996}, difficulties in computational efficiency and stability emerge owing to the largeness and high dimensionality of the datasets. To overcome this, the possibility of sparse estimation has been pursued for decades. Naive methods for sparse estimation yet require solving discrete optimisation problems, involving a serious computational difficulty, even in the simplest case of linear models~\cite{natarajan1995sparse}. Hence, certain relaxations or approximations are required for handling such large high-dimensional datasets.

A breakthrough was made with the least absolute shrinkage and selection operator (LASSO) \cite{tibshirani1996regression}. Its basic idea is to relax the sparsity constraint by using $\ell_1$ regularisation. The success of LASSO motivated the usage of $\ell_1$ regularisation in many different contexts and models~\cite{meinshausen2004consistent,banerjee2006convex,friedman2008sparse}, leading to an ongoing innovation in signal and information processing~\cite{Rish:2014:SMT:2695516,Mairal:2014:SMI:2747300.2747301,Hastie:2015:SLS:2834535}. 

Although LASSO has many attractive properties, the shrinkage introduced by the $\ell_1$ regularisation results in a significant bias in regression coefficients. To solve this, some nonconvex penalties, such as the smoothly clipped absolute deviation (SCAD) penalty \cite{SCAD} and the minimax concave penalty (MCP) \cite{MCP}, have recently been proposed. Although the estimators under these regularisations have desirable properties such as unbiasedness and continuity \cite{SCAD}, there exist some concerns about the stability and interpretability of the estimators because of the potential local minimums owing to the lack of convexity. Hence, investigations of typical performance of those estimators are desired.

Under this situation, one of the present authors recently analyzed the SCAD estimator performance in~\cite{sakata2018approximate}, using the replica method and the message passing technique~\cite{mezard1987spin,nishimori2001statistical,dotsenko2005introduction}. It was shown that there exist two regions in the space of regularization parameters, and in one of them the estimator is uniquely and stably obtained. It was also shown that the estimator outperforms LASSO in the fit quality, and that the emergence of the two regions can be viewed as a phase transition involving replica symmetry breaking (RSB). The latter finding yields a nontrivial insight to the behaviour of local search algorithms, and it was demonstrated that the convergence limit of the coordinate descent (CD) algorithm is closely related to the RSB transition and that a sufficient condition of the convergence, derived in \cite{Breheny2011}, is not tight.

The above-mentioned analysis was limited to the data compression context, in which only the fit quality to a given dataset was considered important. However, with regard to certain applications of sparse estimation, such as compressed sensing~\cite{Donoho2006,DMM_IEEE}, the reconstruction performance of the true signal embedded in the data generation process is more important. The current study addresses this problem and conducts a quantitative analysis for the case where noise exists, while the noiseless limit is investigated in a separate study~\cite{sakata2019perfect}. We provide phase diagrams with respect to regularization parameters derived using the replica method and discuss their implications to the reconstruction performance and the behaviour of local search algorithms. Moreover, we develop an approximate formula for efficiently computing the cross-validation (CV) error, which can be identified with the reconstruction error in our setting. The key results are summarised as follows:
\begin{enumerate}
\item{In the replica symmetric (RS) phase, a unique solution is stably obtained also in the signal reconstruction context. }
\item{The global minimum of the CV error is (presumably always) obtained in the RS phase. }
\item{Our approximate formula efficiently estimates the CV error, without actually conducting CV.}
\end{enumerate} 
These imply that we need not to care about the RSB phase as long as our purpose is to obtain the model best reconstructing the true signal, and in the RS phase we can benefit from the proposed approximate CV formula enabling an efficient estimation of the reconstruction error. Below, we show the theoretical results supporting these messages.

The remaining of the paper is organised as follows: In the next section, our problem setting and an overview of the SCAD penalty are given; in \Rsec{Macroscopic}, the replica analysis result is shown without the derivation because the essential part is already given in~\cite{sakata2018approximate,sakata2019perfect}, and phase diagrams and plots of relevant quantities are shown; in \Rsec{Approximate}, the approximate formula of the CV error is derived; in \Rsec{Numerical}, numerical experiments are carried out on both simulated and real-world datasets to check the accuracy of the replica result and the approximate formula. 
The last section concludes the paper.

\section{Problem settings}\Lsec{Problem}
Suppose a data vector $\V{y}\in \mR^{M}$ is generated by the following linear process with a design matrix $\DM\in \mR^{M\times N}$ and a signal vector $\V{x}^0 \in \mR^{N}$:
\be
\V{y}=\DM\V{x}^0+\V{\noise},
\Leq{generative}
\ee
where $\V{\noise}$ is a noise vector, the component of which is assumed to be an independent and identically distributed (i.i.d.) variable from the normal distribution with zero mean and variance $\sigma_{\noise}^2$, $\mc{N}(0,\sigma_{\noise}^2)$. We denote our dataset as $D_M=\lbb \V{y},\DM \rbb$. In the context of compressed sensing, the design matrix $\DM$ represents the measurement process, and we try to infer $\V{x}^0$ given $\DM$ and $\V{y}$. The inference is herein formulated as a regularised linear regression, and the concrete form of our estimator is given by:
\be
\hat{\V{x}}(\eta,D_M)=\argmin_{\V{x}}\lbb \frac{1}{2}||\V{y}-\DM\V{x}||_2^2+J(\V{x};\eta) \rbb,
\Leq{original}
\ee
where $J(\V{x};\eta)=\sum_{i=1}^NJ(x_i;\eta)$ is the regularisation inducing the estimator sparsity, and $\eta$ is a set of regularisation parameters with a concrete form shown below. To quantify the fit quality of the estimator $\hat{\V{x}}$ to the data $\V{y}$, we introduce:
\be
&&
\MSEy(\hat{\V{x}}|D_M)=\frac{1}{2M}||\V{y}-\DM\hat{\V{x}} ||_2^2,
\Leq{output MSE}
\ee
and call it the output mean squared error (MSE). We also introduce a MSE between the estimator and the true signal as:
\be
&&
\MSEx(\hat{\V{x}}|\V{x}^0)
=
\frac{1}{2N}||\hat{\V{x}}-\V{x}^0||_2^2,
\Leq{input MSE}
\ee
which is termed input MSE, and these characterise the goodness of fit of our estimator $\hat{\V{x}}$.

The purpose of this study is to compute the typical behaviour of $\MSEx$ and $\MSEy$ to obtain insights into the estimator behaviour; meanwhile, some other relevant quantities are also evaluated. The analytical techniques for achieving this purpose are explained in \Rsec{Macroscopic}, with a more detailed description on $\V{x}^0$ and $\DM$.

\subsection{SCAD regularisation}\Lsec{SCAD regularization}
As a representative piecewise continuous nonconvex penalty, we investigate SCAD regularisation in this study. The parameter set consists of $\eta=\{\lambda,a\}~(a>1)$, and the functional form is:
\be
J(\theta;\eta)=
\left\{
\begin{array}{ll}
\lambda|\theta| & (|\theta|\leq \lambda) \\
-\displaystyle\frac{\theta^2-2a\lambda|\theta|+\lambda^2}{2(a-1)} & (\lambda<|\theta|\leq a\lambda) \\
\displaystyle\frac{(a+1)\lambda^2}{2} & (|\theta|>a\lambda)
\end{array}
\right..
\ee 
An illustration of this form is given as the left panel of \Rfig{SCAD}. 
\begin{figure}[htbp]
\begin{center}
\includegraphics[width=0.32\columnwidth]{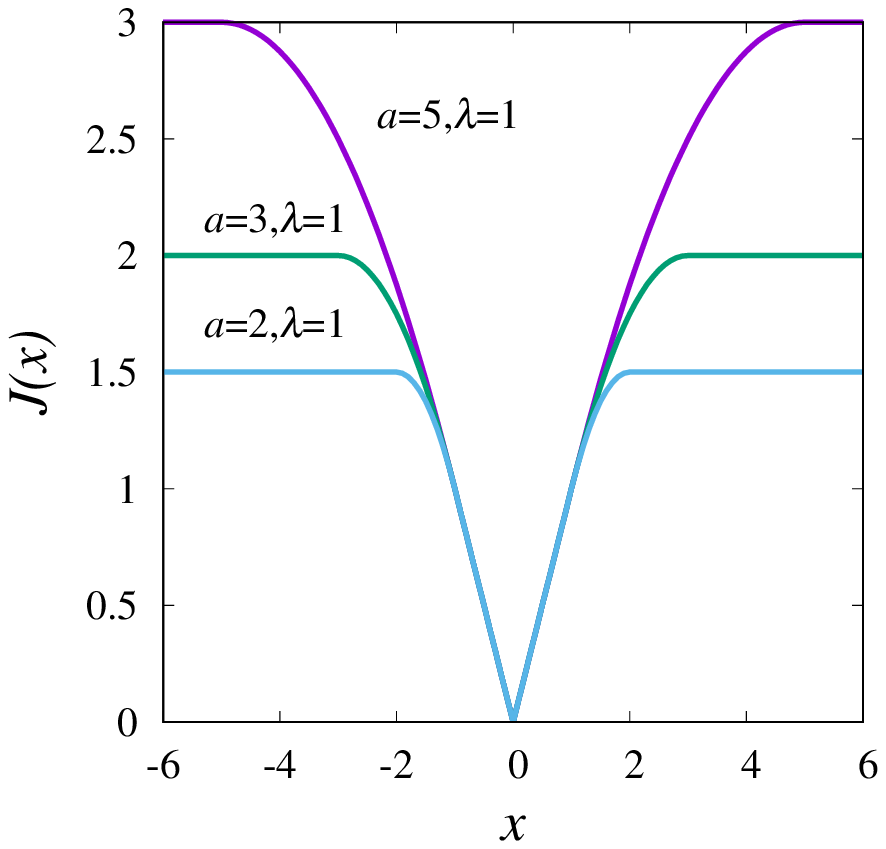}
\includegraphics[width=0.32\columnwidth]{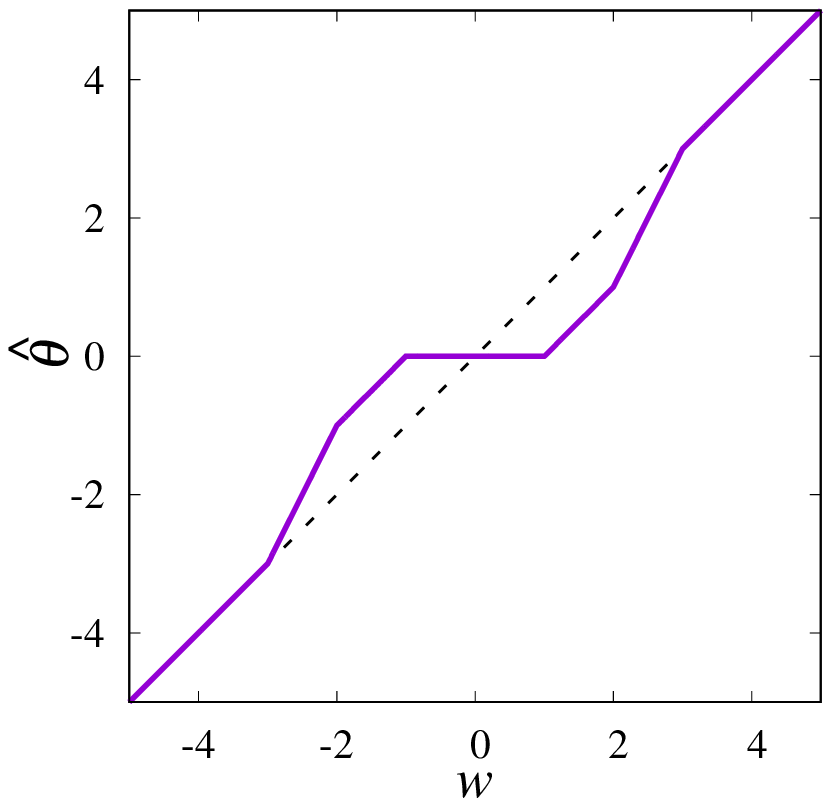}
\includegraphics[width=0.32\columnwidth]{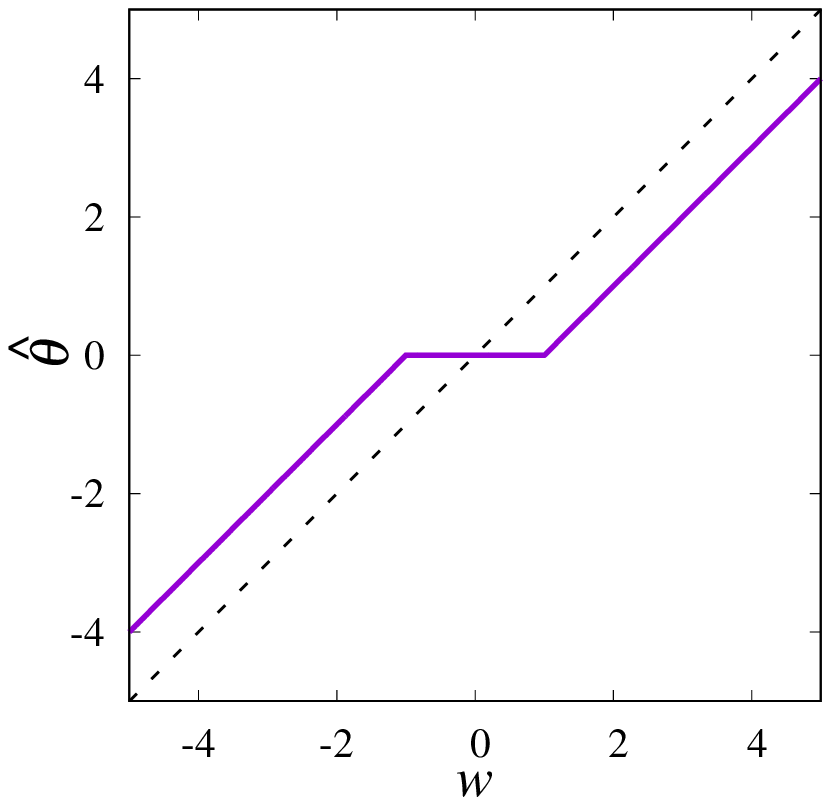}
\caption{
(Left) Shapes of the SCAD regularisations for some parameters. 
(Middle) Behaviour of the SCAD estimator \NReq{estimator-one_dim} at $a=3,\lambda=1,\sigma^2_w=1$; the diagonal dashed line represents the OLS estimator. 
(Right) Behaviour of the LASSO estimator at $\lambda=1$ for comparison with the SCAD; a shrinkage bias is clearly seen for a large $w$.
}
\Lfig{SCAD}
\end{center}
\end{figure}
In the limit $a\to \infty$, the SCAD regularisation tends to be the $\ell_1$ regularisation $J(\theta;\{\lambda,a\to\infty \}) \to \lambda |\theta|$, and correspondingly, the SCAD estimator converges to the LASSO one, allowing the comparison in a continuous manner. For later convenience, we term $a$ the switching parameter, and $\lambda$ the amplitude parameter.

To obtain an intuitive view for the SCAD estimator behaviour, we compute the one-dimensional case:
\be
\hat{\theta}(w;\sigma_w^2,\eta)=\argmin_\theta\left\{\frac{1}{2\sigma_w^2}(\theta-w)^2+J(\theta;\eta)\right\}.
\Leq{one_dim}
\ee
The solution is given by:
\be
\hat{\theta}(w;\sigma_w^2,\eta)
=
V_{\mathrm{SCAD}}(w\slash\sigma_w^2;\sigma_w^2,\eta)S_{\mathrm{SCAD}}(w\slash\sigma_w^2;\sigma_w^2,\eta),
\Leq{estimator-one_dim}
\ee
where 
\be
 S_{\mathrm{SCAD}}(x;\sigma^2,\eta)&=\left\{\begin{array}{ll}
x-\mathrm{sgn}(x)\lambda & \mathrm{for}~\lambda(1+\sigma^{-2})\geq|x|>\lambda\\
x-\mathrm{sgn}(x)\frac{a\lambda}{a-1} & \mathrm{for}~a\lambda\sigma^{-2}\geq|x|>\lambda(1+\sigma^{-2})\\
x & \mathrm{for}~|x|>a\lambda\sigma^{-2}\\
0 & \mathrm{otherwise}
 \end{array}
 \right.,
 \\
 V_{\mathrm{SCAD}}(x;\sigma^2,\eta)&=\left\{\begin{array}{ll}
 \sigma^2 & \mathrm{for}~\lambda(1+\sigma^{-2})\geq|x|>\lambda\\
 \left(\sigma^{-2}-\frac{1}{a-1}\right)^{-1}& \mathrm{for}~a\lambda\sigma^{-2}\geq|x|>\lambda(1+\sigma^{-2})\\
 \sigma^2 & \mathrm{for}~|x|>a\lambda\sigma^{-2}\\
 0 & \mathrm{otherwise}
 \end{array}
\right..
\ee
The middle panel of \Rfig{SCAD} presents an illustration of the estimator at $a=3,\lambda=1,\sigma^2_w=1$, which behaves as the LASSO estimator when $\lambda(1+\sigma^{-2}_w)\geq|w|>\lambda$, and as the ordinary least square (OLS) estimator when $|w|>a\lambda\sigma^{-2}_w$. In the region $a\lambda\sigma^{-2}_w\geq|w|>\lambda(1+\sigma^{-2}_w)$, the estimator linearly transits between LASSO and OLS estimators.

The one-dimensional case estimator plays a key role in our analysis, because our original problem with high dimensionality is, eventually, reduced to an effective one-dimensional problem in the limit $N \to \infty$, termed {\it decoupling principle} in~\cite{guo2005randomly}.

\section{Macroscopic analysis}\Lsec{Macroscopic}
In this section, we provide the order parameters and their determining equations of state (EOS). Associated phase diagrams are shown, and their implications on the performance and the computational stability of the SCAD estimator are also discussed.

For proceeding with the analysis, the ensemble of $\DM$ and $\V{x}^0$ is required to be fixed. We assume that $\DM$ is a random matrix whose component is i.i.d. from $\mc{N}(0,M^{-1})$. The true signal $\V{x}^0$ is also assumed to be a random number drawn from the independent Bernoulli--Gaussian distribution:
\be
P(\V{x}^0)=\prod_{i=1}^{N}\left\{(1-\rho_0)\delta(x_i^0)+\frac{\rho_0}{\sqrt{2\pi\sigma_x^2}}\exp\left(-\frac{(x_i^0)^2}{2\sigma_x^2}\right)\right\}.
\Leq{Bernoulli-Gauss}
\ee
We note that the i.i.d. assumption on $\DM$ is crucial for completing the computation, while the choice of the distribution of $\V{x}^0$ does not matter for the analytical tractability. We admit that this i.i.d. assumption on $\DM$ is not necessarily realistic, but it provides a sufficiently nontrivial setup for our purpose. Although it is possible to extend the present analysis to certain other ensembles~\cite{opper2001adaptive,opper2001tractable,opper2005expectation,cakmak2014s,kabashima2014signal,cespedes2014expectation,rangan2016vector,ma2017orthogonal}, we leave this as a future study.

In the following discussion, we consider the so-called thermodynamic limit $N\to \infty$, while keeping $\alpha\equiv M/N=O(1)$.

\subsection{Outline of analysis}\Lsec{Outline of}
In order to avoid duplication with \cite{sakata2018approximate,sakata2019perfect}, an outline of the analysis is presented here, instead of the EOS derivation.

Our analysis starts from defining Hamiltonian $\mc{H}$, partition function $Z$, and free energy density $f$ as follows:
\be
&&
\mc{H}\lb\V{x}|D_M \rb \equiv \frac{1}{2}||\V{y}-\DM\V{x}||_2^2+J(\V{x};\eta),
\\ && 
Z(\beta|D_M)\equiv \int d\V{x}~e^{-\beta \mc{H}(\V{x}|D_M)},
\\ && 
f(\beta|D_M)\equiv -\frac{1}{N \beta} \ln Z(\beta|D_M).
\ee
As seen from \Req{original}, the minimiser or the ground state of $\mc{H}$ corresponds to our estimator and, hence, we are interested in the $\beta \to \infty$ limit of the free energy density. The input and output MSEs can be computed from the free energy in this limit, by following a standard prescription. The free energy density becomes the primary object to be computed, and enjoys the self-averaging property, and the typical value thus converges to the averaged one $f(\beta|D_M)\to E_{\V{y},\DM}\lsb f(\beta|D_M) \rsb\equiv f(\beta)$, where $E_{\V{y},\DM}\lsb \cdots \rsb$ denotes the average over $\V{y}$ and $\DM$. 

Unfortunately, the average density $E_{\V{y},\DM}\lsb f(\beta|D_M) \rsb$ is not analytically tractable. To overcome this, we employ the following identity:
\be
E_{\V{y},\DM}[\ln Z(\beta|D_M)]=\lim_{n\to 0}\frac{E_{\V{y},\DM}[Z^n(\beta|D_M)]-1}{n}.
\Leq{replica}
\ee
However, the computation for general $n \in \mR$ is still intractable; thus, we additionally assume that $n$ is a positive integer, because $E_{\V{y},\DM}[Z^n(\beta|D_M)]$ is analytically computable for $n \in \mN$. Then, using an analytically continuable expression of $E_{\V{y},\DM}[Z^n(\beta|D_M)]$ from $\mN$ to $\mR$, we evaluate $\lim_{n\to 0}E_{\V{y},\DM}[Z^n(\beta|D_M)]$ at the final step. These procedures are termed replica method.  

The final expression of $f(\beta)$ is given as an extremisation problem with respect to a number of parameters, called order parameters. The extremisation condition appears because of the limit $N\to \infty$, and yields EOS determining the values of the order parameters. The explicit formulas are given below. It should be noted that the following analysis is conducted only under the RS assumption, although RSB occurs in some parameter regions. This is because the RS analysis is sufficient for the present purpose of obtaining insights on the stability of the SCAD estimator. Beyond this purpose, the RSB analysis will provide further quantitative information about the estimator when many local minimums exist, which will be an interesting future direction.

\subsection{Order parameters, equations of state, and stability condition}\Lsec{Order parameters}
Here, we summarise the order parameters and EOS. In the RS level, our system is characterised by the following three-order parameters:
\be
\Leq{order parameters}
&&
m=\frac{1}{N}\sum_i E_{\V{y},\DM }\lsb \Ave{ x^0_{i}  x_i } \rsb,
\\ &&
Q=\frac{1}{N}\sum_i E_{\V{y},\DM }\lsb  \Ave{ x_i^2 } \rsb,
\\ &&
q=\frac{1}{N}\sum_i  E_{\V{y},\DM }\lsb \Ave{x_i}^2\rsb,
\ee
where the angular brackets, $\Ave{\cdots}$, denote the average over the Boltzmann distribution $P(\V{x}|\beta,D_M)=e^{-\beta \mc{H}(\V{x}|D_M)}/Z(\beta|D_M)$. $m$ is the overlap with the true signal $\V{x}^0$ and is relevant to the reconstruction performance. $Q$ and $q$ both describe the powers (per element) of the estimator, but the latter takes into account the `thermal' fluctuation that results from the introduction of $\beta$. These two quantities fall within the limit $\beta \to \infty$, but their infinitesimal difference yields an important contribution:
\be
\chi=\beta(Q-q).
\Leq{limit-chi}
\ee
This is $O(1)$, even in the limit $\beta \to \infty$. Besides, we introduce the conjugate parameters of $Q,\chi,m$ as $\T{Q},\T{\chi},\T{m}$, respectively, and denote their sets as $\Omega=\{Q,\chi,m\}$ and $\T{\Omega}=\{\T{Q},\T{\chi},\T{m}\}$. The RS free-energy density in the limit $\beta \to \infty$ takes the following extremisation problem, with respect to $\Omega$ and $\T{\Omega}$:
\be
\hspace{-1.5cm}
f(\beta \to \infty)=\Extr{\Omega,\T{\Omega}}
\lbb
\frac{Q-2m+\rho_0 \sigma_x^2 +\alpha \sigma_{\noise}^2}{2(1+\chi/\alpha)}+m\T{m}-\frac{\T{Q}Q-\T{\chi}\chi}{2}+\frac{\overline{\xi(\sigma;\T{Q})}}{2}
\rbb,
\Leq{f_RS}
\ee
where:
\be
\OBP(h;\T{Q})&\equiv 
\min_x 
\lbb 
 \frac{\T{Q}}{2}x^2-h x+J(x;\eta)
\rbb.
\Leq{one_body}
\\
\int Dz(\cdots)&\equiv \int_{-\infty}^{\infty}\frac{dz}{\sqrt{2\pi}}\exp\lb -\frac{1}{2}z^2 \rb (\cdots),
\\
\xi(\sigma;\T{Q})&\equiv 2\int Dz~\OBP(\sigma z;\T{Q}),
\Leq{xi_def}
\ee
and $\overline{\cdots}$ represents the average over $\sigma$, whose distribution is:
\be
&&
P_\sigma(\sigma)=(1-\rho)\delta(\sigma-\sigma_{-})+\rho\delta(\sigma-\sigma_+),
\Leq{sigma_dist}
\\ &&
\sigma_{-}=\sqrt{\T{\chi}},
\\ &&
\sigma_+=\sqrt{\T{\chi}+\T{m}^2\sigma_x^2}.
\ee
The minimiser of \Req{one_body} is the solution of the one-dimensional problem \NReq{estimator-one_dim}, with $\sigma_{w}^2\to \T{Q}^{-1}$ and $ w\to h/\T{Q}$, thus we can denote it as:
\be
x^{*}(h;\T{Q}^{-1})=V_{\rm SCAD}(h;\T{Q}^{-1},\eta)S_{\rm SCAD}(h;\T{Q}^{-1},\eta).
\Leq{x^*_SCAD}
\ee
The extremisation condition in \Req{f_RS} yields EOS as:
\numparts
\be
&& \hspace{-0.5cm}
\chi=\int Dz 
\overline{\frac{\partial x^*(h;\T{Q}^{-1})}{\partial h}\Biggr|_{h=\sigma z}  }
=\frac{1}{\T{Q}}
  \lbb 
   \hat{\rho}+\frac{\frac{1}{a-1}}{\T{Q}-\frac{1}{a-1}}\overline{\xi_4(\sigma)}
  \rbb,
\Leq{RS_chi_gen}
\\ && \hspace{-0.5cm}
Q
=\int Dz \overline{(x^*(\sigma z;\T{Q}^{-1}))^2 }
=\overline{\left\{\frac{\xi_1(\sigma)}{\T{Q}}+\frac{\xi_2(\sigma)}{\T{Q}-\frac{1}{a-1}}+\frac{\xi_3(\sigma)}{\T{Q}}\right\}},
\Leq{RS_Q_gen}
\\ && \hspace{-0.5cm}
m
=\rho\T{m}\sigma_x^2
\int Dz
\frac{\partial x^*(h;\T{Q}^{-1})}{\partial h}\Biggr|_{h=\sigma_+ z}
=
\rho_0 \sigma_x^2
\lbb 
\mathrm{erfc}(\theta_1(\sigma_+))+\frac{\frac{1}{a-1}\xi_4(\sigma_+)}{\T{Q}-\frac{1}{a-1}}
\rbb,
\Leq{RS_m_gen}
\\ && \hspace{-0.5cm}
\T{\chi}=\frac{1}{\alpha} \frac{Q-2m+\rho_0 \sigma_x^2+\alpha\sigma_{\noise}^2}{(1+\chi/\alpha)^2},
\Leq{RS_chih}
\\ && \hspace{-0.5cm}
\T{Q}=\frac{1}{1+\chi/\alpha},
\Leq{RS_Qh}
\\ && \hspace{-0.5cm}
\T{m}=\frac{1}{1+\chi/\alpha},
\Leq{RS_Qh}
\ee
\endnumparts
where
\numparts
\Leq{EOSaux}
\be
&&
\theta_1(\sigma)={\lambda}\slash(\sqrt{2}\sigma),
\\ &&
\theta_2(\sigma)=\lambda(1+\T{Q})\slash(\sqrt{2}\sigma),
\\ &&
\theta_3(\sigma)=a\lambda\T{Q}\slash(\sqrt{2}\sigma),
\\ &&
\hat{\rho}=\overline{\mathrm{erfc}(\theta_1(\sigma))},
\\ &&
\xi_1(\sigma)=\frac{\sigma^2}{\T{Q}}\Big[-\frac{2\theta_{1}(\sigma)}{\sqrt{\pi}}\Big(e^{-\theta_{1}^2(\sigma)}+(\T{Q}-1)e^{-\theta_{2}^2(\sigma)}\Big)
\no \\
&&+(1+2\theta_{1}^2(\sigma))\{\mathrm{erfc}(\theta_1(\sigma))-{\rm erfc}(\theta_{2}(\sigma))\}\Big],
\\
&&
\xi_2(\sigma)=
\frac{
\sigma^2
}{  \T{Q}-\frac{1}{a-1}
}
\Big[
 \frac{2}{\sqrt{\pi}}
  \Big\{
  \theta_{2}(\sigma)e^{-\theta_{2}^2(\sigma)}-\theta_{3}(\sigma)e^{-\theta_{3}^2(\sigma)}
\no \\ 
  &&
  -\frac{2\theta_{3}(\sigma)}{\T{Q}(a-1)}\left(e^{-\theta_{2}^2(\sigma)}-e^{-\theta_{3}^2(\sigma)} \right)
  \Big\}
   +\Big\{1+2\Big(\frac{\theta_{3}(\sigma)}{\T{Q}(a-1)}\Big)^2\Big\}\xi_4(\sigma)
\Big],
\\
&&
\xi_3(\sigma)=\frac{\sigma^2}{\T{Q}}\Big[\frac{2\theta_{3}(\sigma)}{\sqrt{\pi}}e^{-\theta_{3}^2(\sigma)}+{\rm erfc}(\theta_{3}(\sigma))\Big],
\\
&&
\xi_4(\sigma)={\rm erfc}(\theta_{2}(\sigma))-{\rm erfc}(\theta_{3}(\sigma)).
\ee
\endnumparts
The SCAD regularisation divides the domain of definition into some analytic components, and $\{\theta_i \}_{i=1}^{3}$ are the corresponding boundary values for $z$ for the integration $\int Dz(\cdots)$. The parameter $\hat{\rho}$ is the density of non-zero components in the estimate.

Using the solution of EOS, the input and output MSEs can be expressed as:
\be
&&
\MSEx=\frac{1}{2}\lb \rho_0 \sigma_x^2-2m +Q \rb,
\\ &&
\MSEy=\frac{1}{2}\T{\chi}.
\ee
Furthermore, we additionally quantify the reconstruction performance of the support of the true signal. Denoting the support or {\em active set} of $\V{x}$ as $S_{A}(\V{x})=\{i|x_i\neq 0\}$, we introduce the true positive rate $TP(\V{x}|\V{x}_0)=\frac{|S_{A}(\V{x}) \cap S_{A}(\V{x}^0)|}{|S_{A}(\V{x}^0)|}$ and the false positive rate $FP(\V{x}|\V{x}_0)=\frac{|S_{A}(\V{x}) \cap S^{c}_{A}(\V{x}^0)|}{|S^{c}_{A}(\V{x}^0)|}$, where $S^c$ denotes the complement set of $S$.
These are expressed by using the solution of EOS as:
\be
&&
TP=\int Dz ~\left| x^*(\sigma_+ z;\T{Q}^{-1}) \right|_0=\mathrm{erfc}(\theta_1(\sigma_+)),~
\Leq{TP}
\\ &&
FP=\int Dz ~\left| x^*(\sigma_{-} z;\T{Q}^{-1}) \right|_0=\mathrm{erfc}(\theta_1(\sigma_{-})),
\Leq{FP}
\ee
where $| x |_0$ expresses $\ell_0$ operator giving $0$ if $x=0$ and $1$ otherwise. Following the standard analysis~\cite{sakata2018approximate}, we can derive the stability condition of the RS solution, called de Almeida--Thouless (AT) condition~\cite{AT}. The derivation of our specific case is already given in~\cite{sakata2018approximate} and we just quote the resultant expression: 
\be
&&
\frac{1}{\alpha \lb 1+\chi/\alpha \rb^2}
\overline{
\int Dz \lb \Part{x^{*}(h;\T{Q}^{-1})}{h}{}\Biggr|_{h=\sigma z}\rb^2
}
\no \\ &&
=
\frac{1}{\alpha \lb 1+\chi/\alpha \rb^2}
\left[\frac{\hat{\rho}}{\T{Q}^2}+\left\{\left(\frac{1}{\T{Q}\!-\!\frac{1}{a-1}}\right)^2-\frac{1}{\T{Q}^2}\right\}\overline{\xi_4(\sigma)}\right]
<1.
\Leq{AT}
\ee
Apart from the AT condition, we also notice that the RS solution does not exist when the switching nonconvexity parameter $a$ is small. This is because \Req{one_body} tends to have no solution in the small $a$ limit, leading to the following existence condition: 
\be
\T{Q}- \frac{1}{a-1} \geq 0.
\Leq{RS feasible}
\ee
These provide sufficient information for the following analyses.

\subsection{Phase diagram}\Lsec{Phase diagram}
In this subsection, we show the phase diagrams in the $\lambda$--$a$ plane for a wide range of parameters. We introduce three boundaries: The first one, derived from \Req{AT}, is the AT line $a_{\rm AT}(\lambda)$ below which the RS solution is unstable; the second, derived from \Req{RS feasible}, is the existence limit of the RS solution $a_{\rm RS}(\lambda)$, below which the RS solution does not exist; and the third, $a_{\rm IMSE}(\lambda)$ represents the minimum point of the input MSE $\MSEx$, when sweeping $\lambda$ given $a$. For clarity, the variance of the non-zero components of $\V{x}^0$ is fixed as $\sigma_x^2=1/\rho_0$, setting the signal power per component unity, in average, $\sum_{i=1}^N\lb x^0_i \rb^2/N\approx 1$. 

First, we compare the phase diagrams for different noise strengths $\sigma_{\noise}^2$ at $\alpha=0.5$ and $\rho_0=0.2$ in \Rfig{PD-noise}. 
\begin{figure}[htbp]
\begin{center}
\includegraphics[width=0.32\columnwidth]{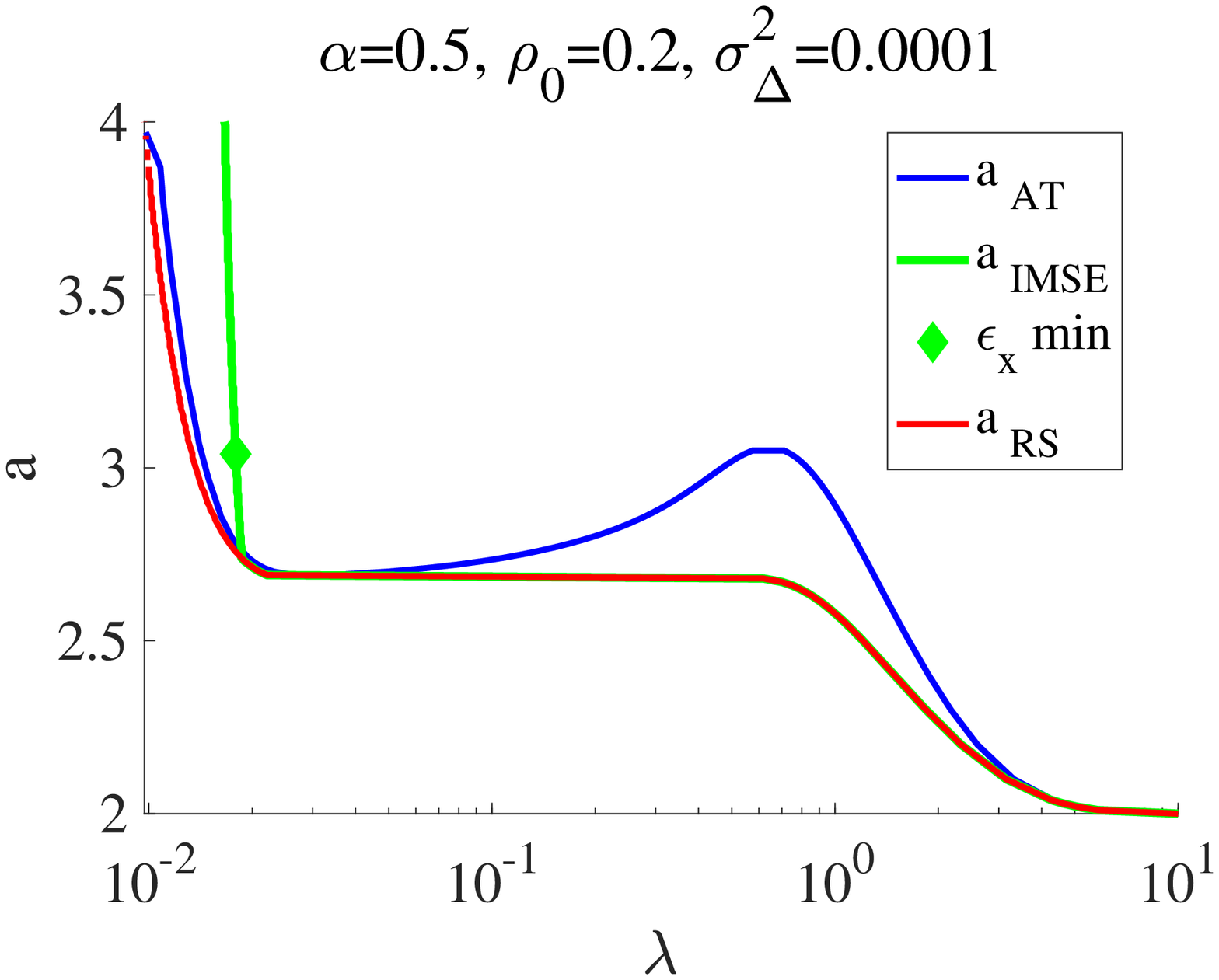}
\includegraphics[width=0.32\columnwidth]{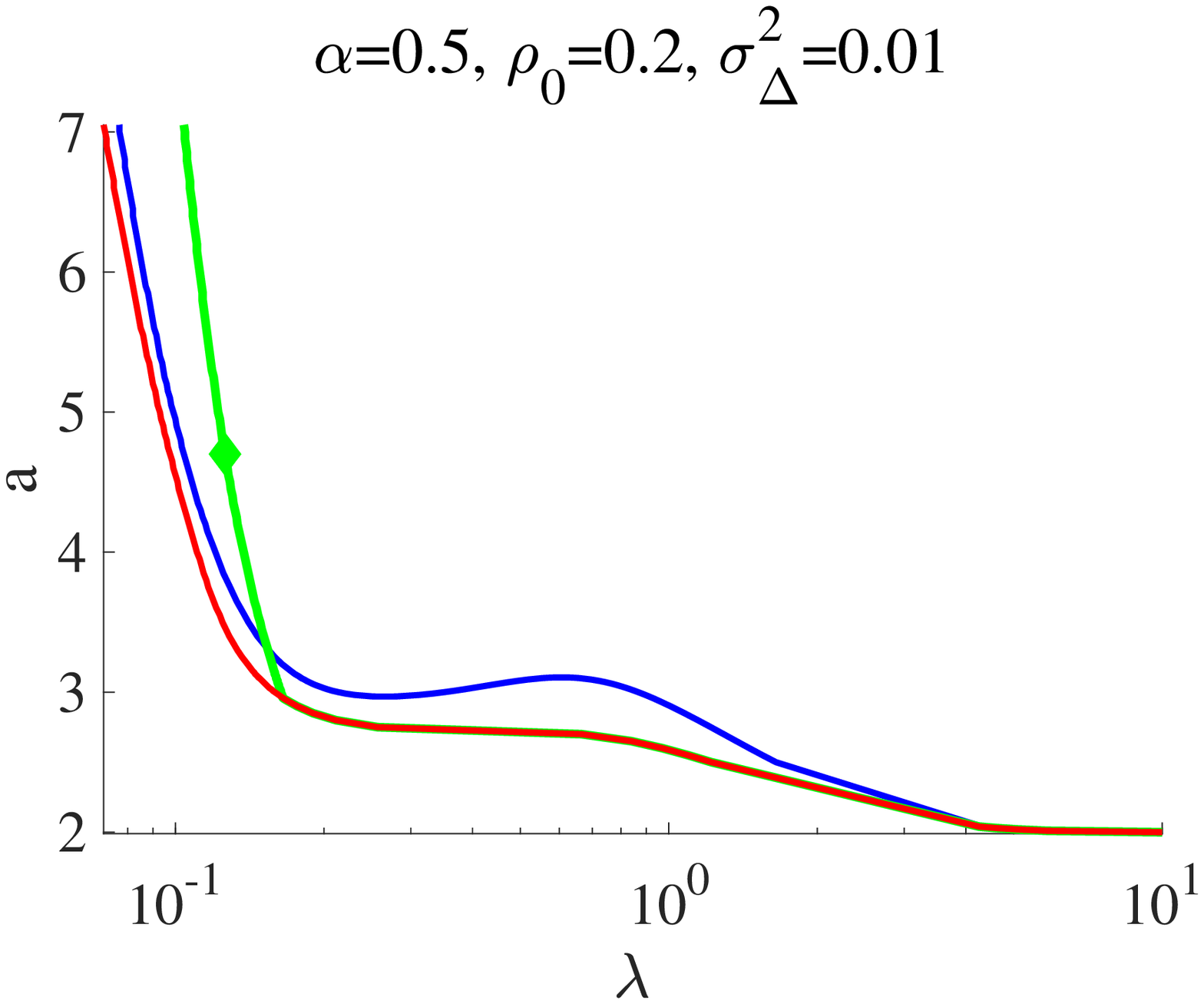}
\includegraphics[width=0.32\columnwidth]{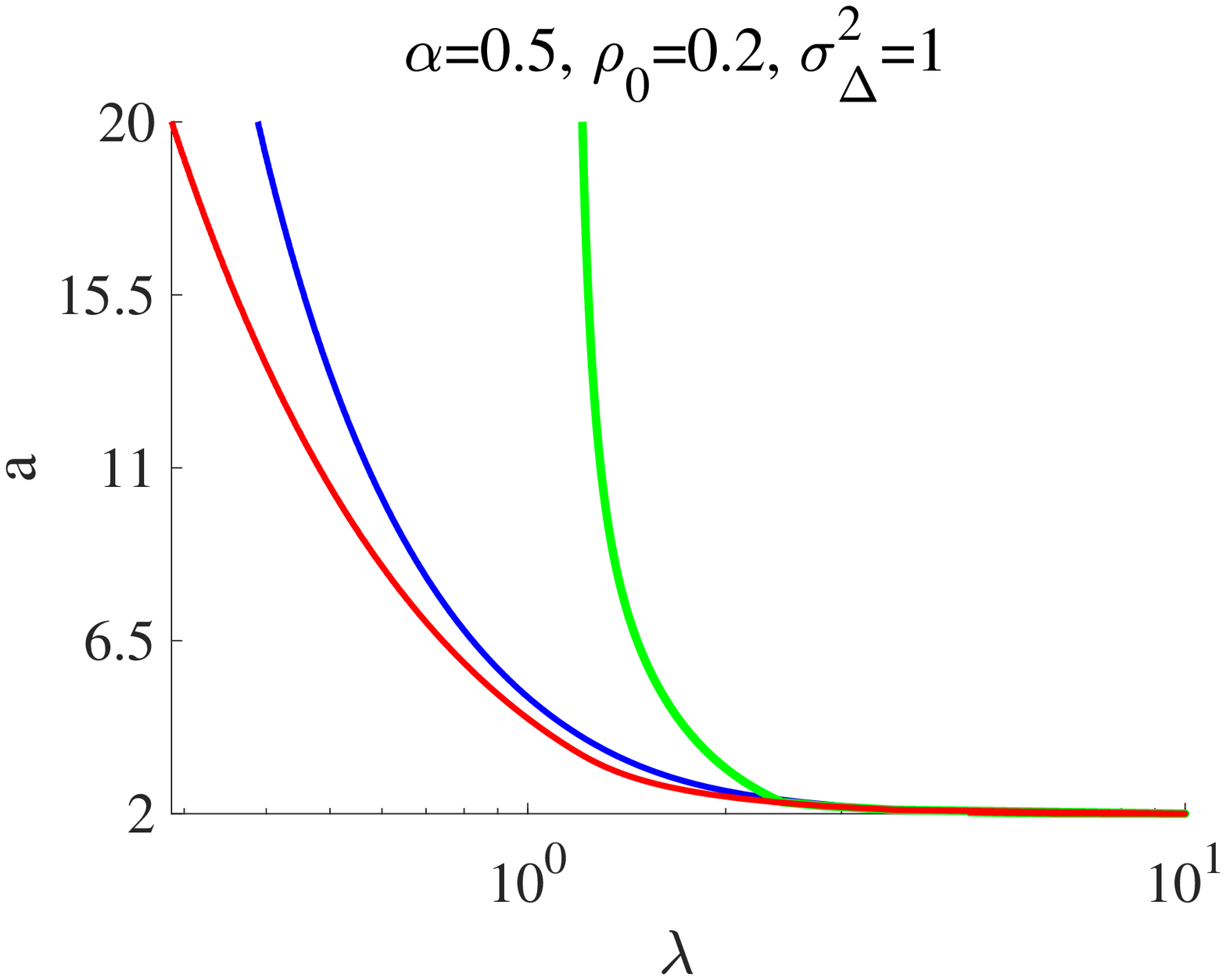}
\caption{
Phase diagrams in the $\lambda$--$a$ plane for different noise strengths $(\sigma_{\noise}^2=10^{-4}$ (left), $10^{-2}$ (middle),  $1$ (right)) at $\alpha=0.5$ and $\rho_0=0.2$. The blue, green, and red lines denote $a_{\rm AT},a_{\rm IMSE},$ and $a_{\rm RS}$, respectively. The green diamond represents the location of the global minimum of the input MSE $\MSEx$. For the right panel of the largest noise case $\sigma_{\noise}^2=1$, there seems to be no global minimum of $\MSEx$ for finite $a$ (located in the LASSO limit $a\to \infty$). 
}
\Lfig{PD-noise}
\end{center}
\end{figure}
We plot $a_{\rm AT},a_{\rm RS},$ and $a_{\rm IMSE}$ by blue, red, and green lines, respectively. The green diamond represents the location of the global minimum of $\MSEx$ under the RS assumption. A useful finding concerning this is that the location is above the AT line, which is always the case as far as we have examined, and some additional evidences are later given in \Rfigs{PD-rho}{PD-alpha}. In the right panel of \Rfig{PD-noise}, the green diamond is not shown, because the input MSE continuously decreases as $a$ grows, implying that the global minimum of $\MSEx$ is obtained at the LASSO limit $a\to \infty$. These imply that the best reconstruction performance of the true signal is always obtained in the RS phase, which is one of main claims of this study. Admittedly, there is a possibility that the true global minimum exists in the RSB phase, and the green diamond just represents a local minimum. Our present analysis does not exclude this possibility. To clarify this point, further quantitative analysis in the RSB framework is required, but this is beyond the scope of this study. 

Another interesting observation in \Rfig{PD-noise} is the re-entrant phase transition concerning $\lambda$ in relatively small $a$ regions for the weak noise cases (left and middle panels). 
For example at $a=2.8$ in the left panel, when decreasing $\lambda$ from a large enough value, we first go across the rightmost branch of $a_{\rm AT}$ around $\lambda\approx 1$ and enter into the RSB phase from the RS phase; further decreasing $\lambda$ we meet the middle branch of $a_{\rm AT}$ around $\lambda \approx 0.1$ and thus re-enter into the RS phase; still decreasing $\lambda$ we hit the leftmost branch of $a_{\rm AT}$ around $\lambda \approx 0.01$ and we are eventually in the RSB phase. Although the physical reason of the emergence of the re-entrance is not clear, it seems to only exist in the weak noise region.  We also note that the AT line, $a_{\rm AT}$, is always located above $a_{\rm RS}$. The solution vanishment in the low $\lambda$ region is thus an artefact of the RS assumption, and the corresponding parameter regions should be described by the RSB solution. 

Next, we check the $\rho_0$ dependence of the phase structures. Phase diagrams at $\alpha=0.5$ and a moderate noise level $\sigma_{\noise}^2=0.1$ are shown in \Rfig{PD-rho}. 
\begin{figure}[htbp]
\begin{center}
\includegraphics[width=0.32\columnwidth]{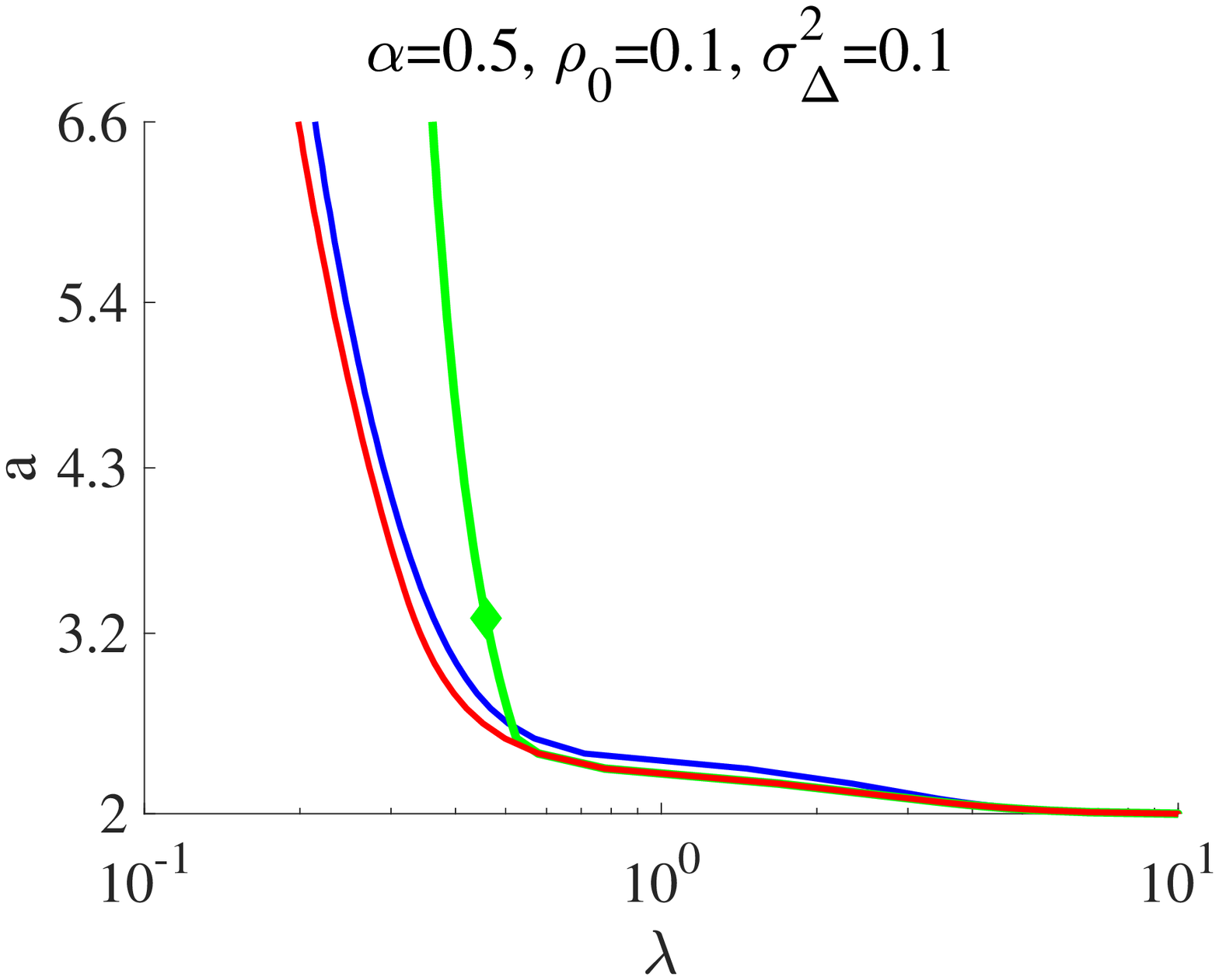}
\includegraphics[width=0.32\columnwidth]{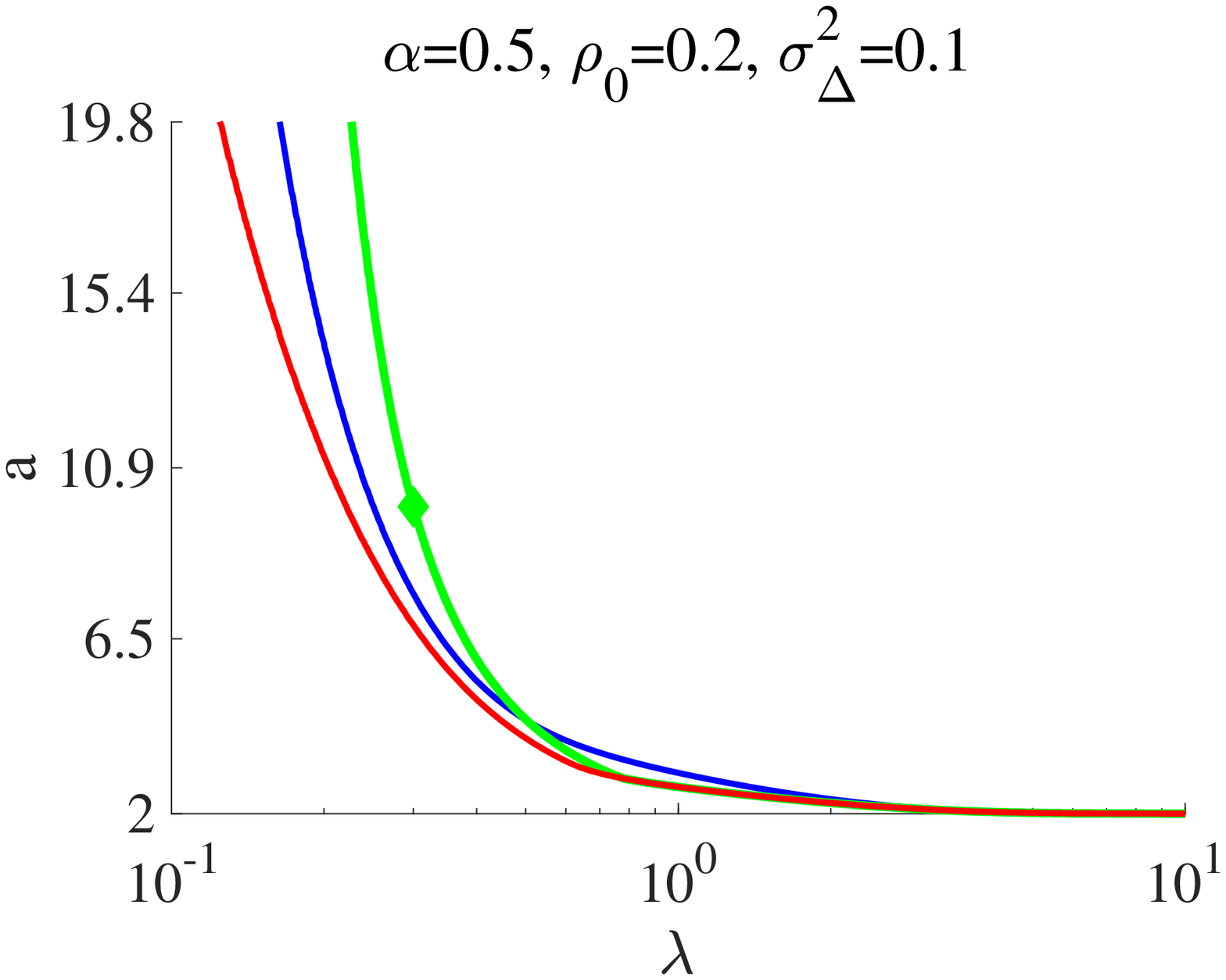}
\includegraphics[width=0.32\columnwidth]{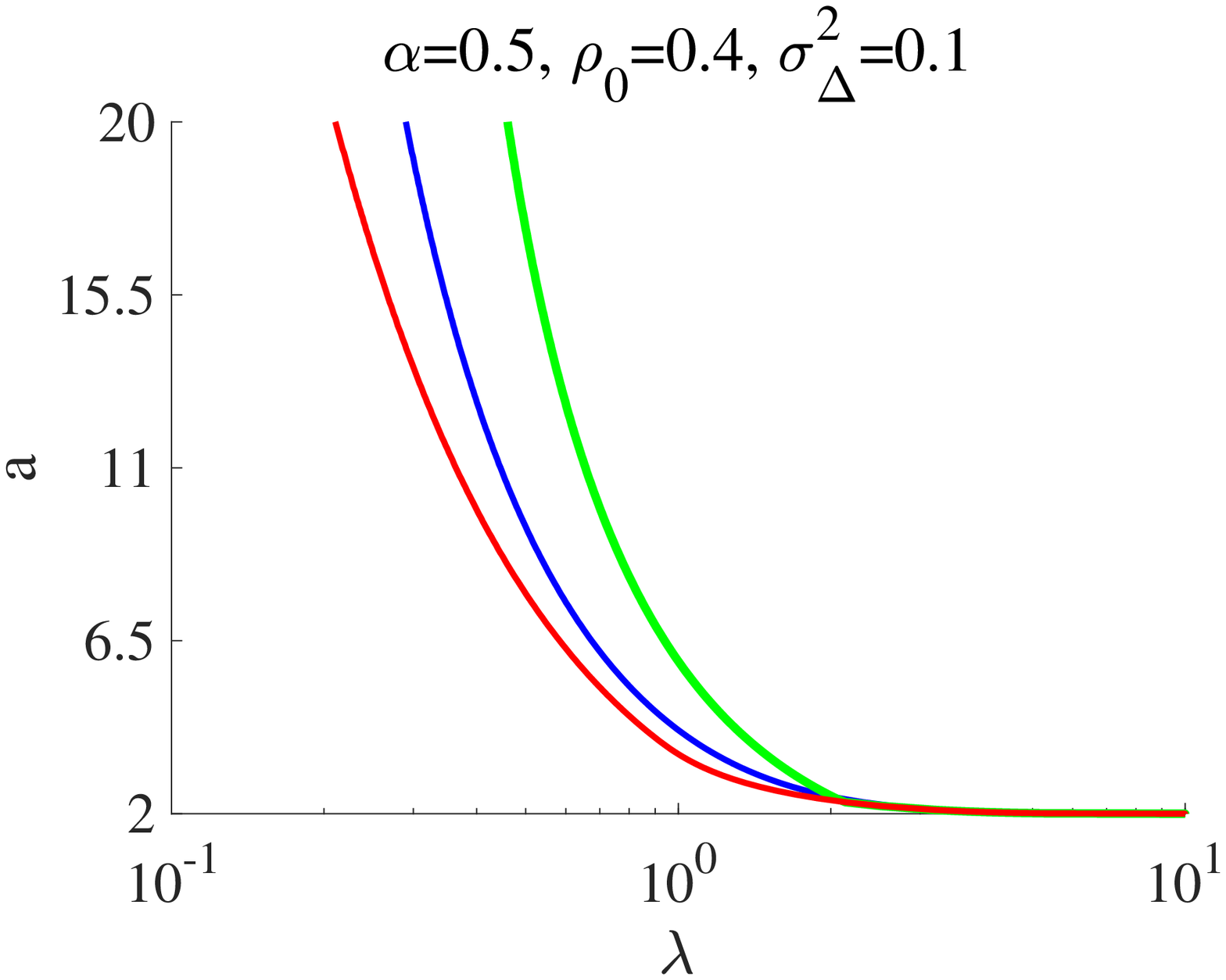}
\caption{
$\lambda$--$a$ phase diagrams for different densities of non-zero components $(\rho_0=0.1$ (left), $0.2$ (middle), $0.4$ (right) at $\alpha=0.5$ and $\sigma_{\noise}^2=0.1$. The lines and markers have the same meaning as \Rfig{PD-noise}. As $\rho_0$ increases, the location of the minimum of $\MSEx$ tends to be at larger values of $a$. 
}
\Lfig{PD-rho}
\end{center}
\end{figure}
Although the basic structure does not change from \Rfig{PD-noise}, the re-entrant transitions in the weak noise cases disappear. As $\rho_0$ increases, the minimum location of $\MSEx$ increases along the $a$--axis, and for the large $\rho_0$ (right panel) the green diamond tends to disappear at finite values of $a$, implying the LASSO limit yields the minimum of $\MSEx$ as the strong noise case.

The last phase diagrams are given for checking the $\alpha$ dependence. Phase diagrams for $\alpha=0.3,0.8,1.5$ at $\rho_0=0.2$ and $\sigma_{\noise}^2=0.1$ are shown in \Rfig{PD-alpha}.
\begin{figure}[htbp]
\begin{center}
\includegraphics[width=0.32\columnwidth]{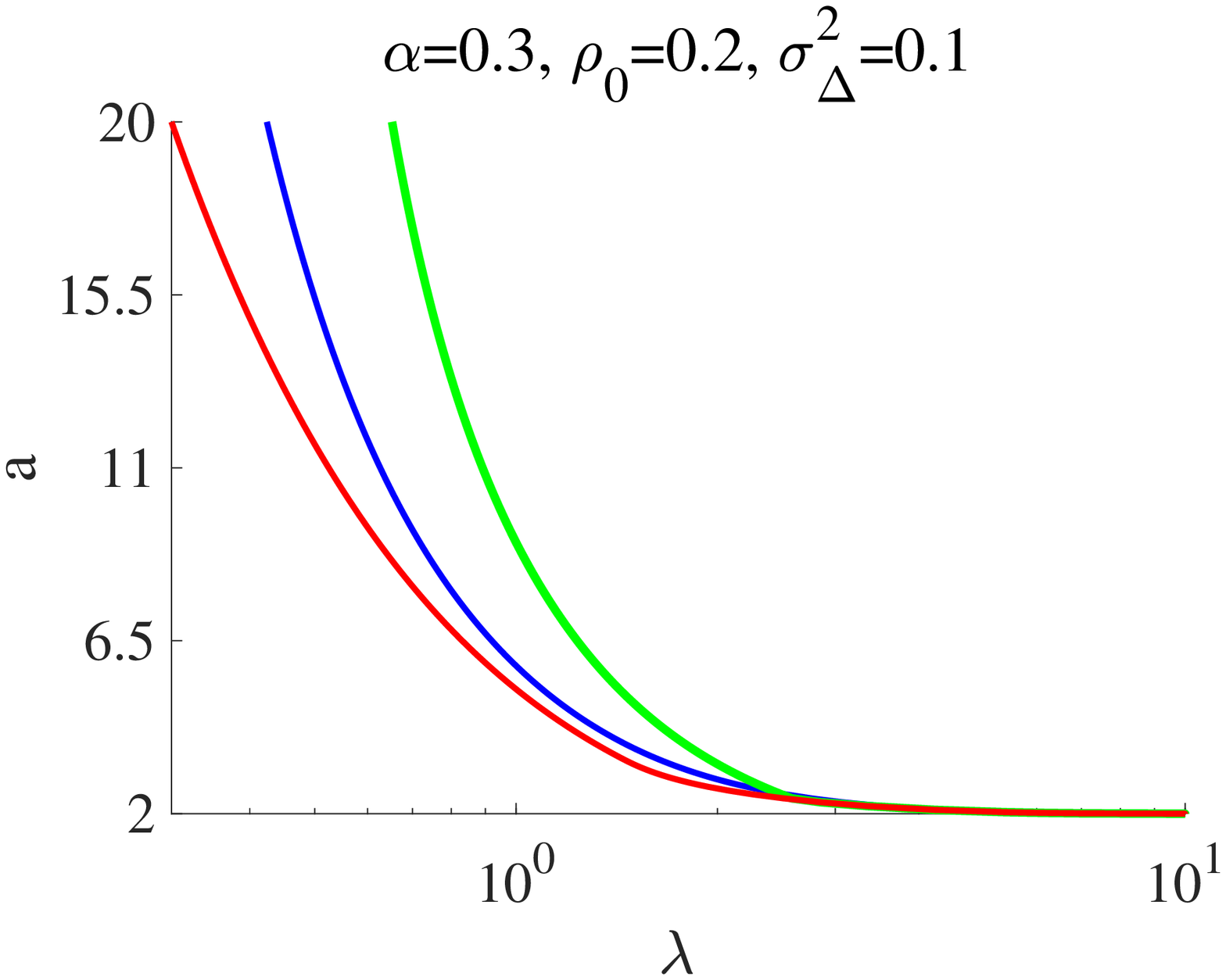}
\includegraphics[width=0.32\columnwidth]{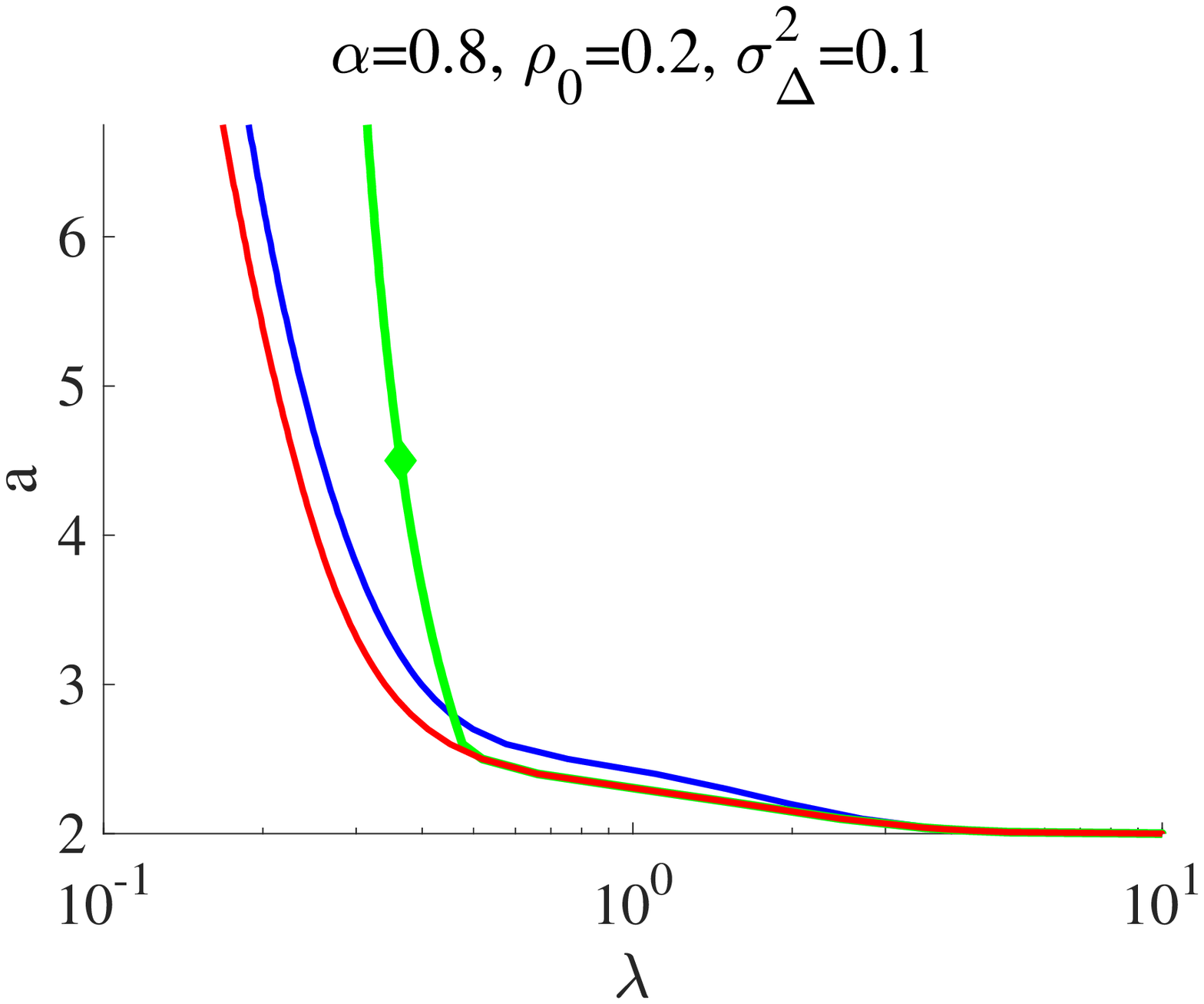}
\includegraphics[width=0.32\columnwidth]{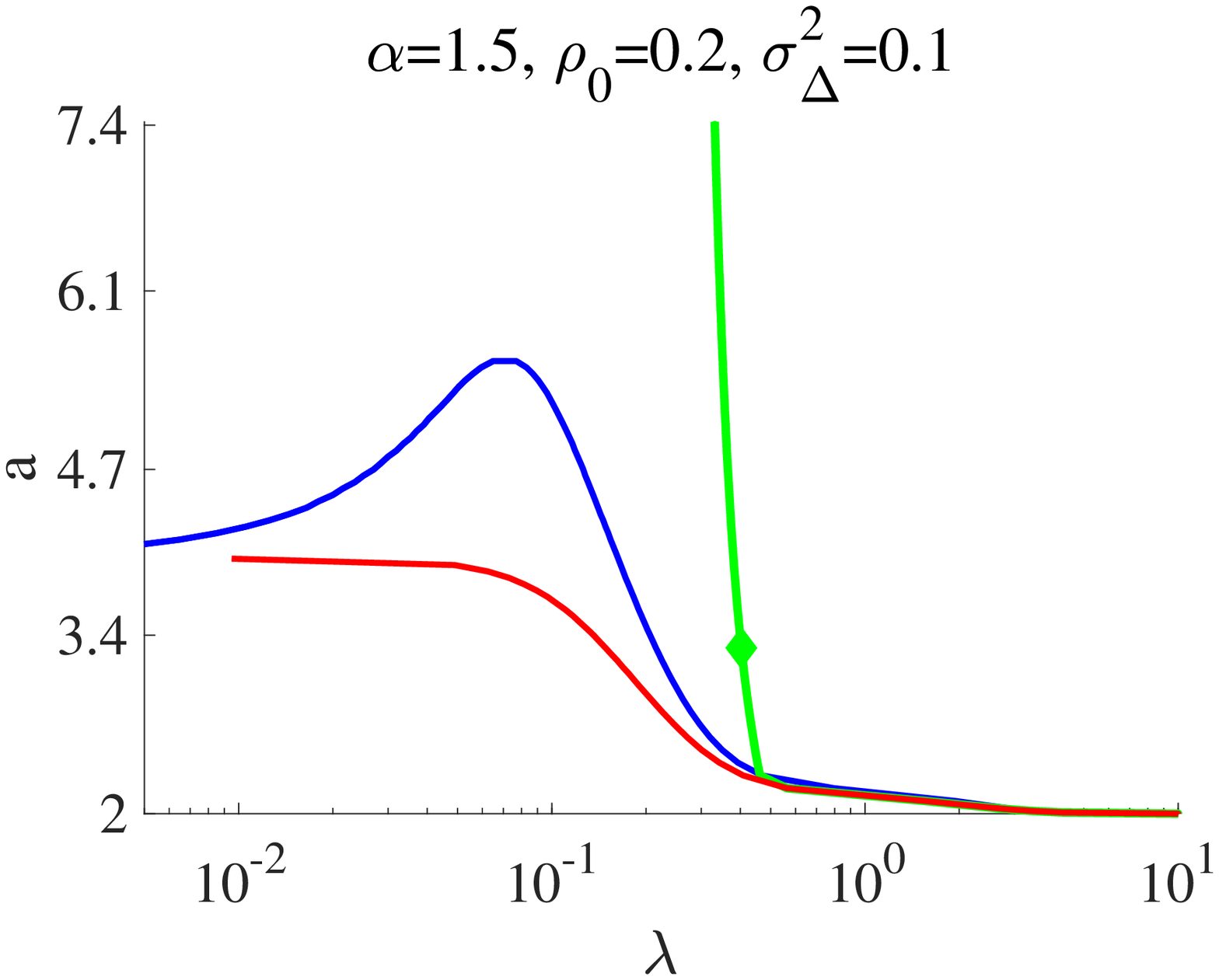}
\caption{
Phase diagrams for different ratios of the dataset size to the model dimensionality $(\alpha=0.3$ (left), $0.8$ (middle), $1.5$ (right)) at $\rho_0=0.2$ and $\sigma_{\noise}^2=0.1$. The lines and markers have the same meaning as \Rfig{PD-noise}. 
}
\Lfig{PD-alpha}
\end{center}
\end{figure}
As seen in the left panel, if the value of $\alpha$ is close to that of $\rho_0$, the larger $a$ tends to give smaller values of the input MSE, as in the right panel of \Rfig{PD-rho}. In contrast to the other diagrams of the underdetermined case $(\alpha<1)$, the right panel of the $\alpha=1.5$ case shows a particular behaviour, as both the AT line and the RS existence limit tend to converge to certain finite values of $a$ in the $\lambda \to 0$ limit. Hence, the whole $\lambda$ region becomes RS at sufficiently large but finite $a$ values.

In all the phase diagrams shown above, the minimum value of $a$ is fixed to be $2$. This is because the RS solution cannot describe the region $a < 2$. There is a simple reason for this. According to the argument of the approximate message passing technique~\cite{sakata2018approximate,sakata2019perfect}, the effective one-dimensional problem \NReq{one_body} corresponds to the following marginal distribution:
\be
P_i(x_i)\propto \lim_{\beta \to \infty} e^{-\beta \lbb \frac{1}{2}\lb \sum_{\mu=1}^{M}\frac{A_{\mu i}^2}{1+\chi_{\mu}}\rb x_i^2 -h_i x_i +J(x_i;\eta) \rbb },
\ee
where the minimiser of \Req{one_body} corresponds to the location of $x_i$, at which the measure concentrates in the limit $\beta\to \infty$. The factor $\lb \sum_{\mu=1}^{M}\frac{A_{\mu i}^2}{1+\chi_{\mu}}\rb$ corresponds to $\T{Q}$, while $\chi_{\mu}$ is a non-negative quantity related to $\chi$ in the RS solution. This means that, under the assumption of $A_{\mu i}\sim \mc{N}(0,1/M)$, $\T{Q}$ is bounded as:
\be
\T{Q}=\lb \sum_{\mu=1}^{M}\frac{A_{\mu i}^2}{1+\chi_{\mu}}\rb 
\leq 
\sum_{\mu=1}^{M} A_{\mu i}^2 \approx 1.
\Leq{Qleq1}
\ee 
Combining this with \Req{RS feasible}, we find that the condition $a\geq 2$ is necessary for the existence of the RS solution. The merging behaviour of the three lines to the $a=2$ line, as $\lambda$ grows in the phase diagrams, well matches to this condition. Although $a=2$ is a known critical value~\cite{fan2001variable,lee2015strong}, the above argument provides another perspective from a different viewpoint. We also note that this non-existence of the RS solution does not mean the non-existence of the SCAD estimators. Actually, numerical experiments easily show that the estimators take non-trivial values in the region $a<2$, and they tend to show strong multiplicity and dependency on the initial condition. To analyse the behaviour of those estimators, we need to consider the RSB solution, but it is beyond the present purpose as already declared.

\subsection{Receiver operating characteristic curve}\Lsec{Receiver operating}
To characterise the reconstruction performance of the true signal's support, we employ the so-called receiver operating characteristic (ROC) curve. The ROC curve is a plot of $TP$ \NReq{TP} against $FP$ \NReq{FP}. The best ROC curve goes through the point $(TP,FP)=(1,0)$. Accordingly, to quantify `optimality' of the points on a ROC curve, we use the following quantity:
\be
R=(TP-1)^2+(FP-0)^2.
\ee
Thus, the smallest value of $R$ defines the `optimal' point of the ROC curve. This easy-to-use quantity is commonly applied as a criterion, followed here.

First, ROC curves when sweeping $\lambda$ at given values of $a$ are plotted in the left panel of \Rfig{ROC-a}. The other parameters are $(\alpha,\rho_0,\sigma_{\noise}^2)=(0.5,0.2,0.1)$. 
\begin{figure}[htbp]
\begin{center}
\includegraphics[width=0.48\columnwidth]{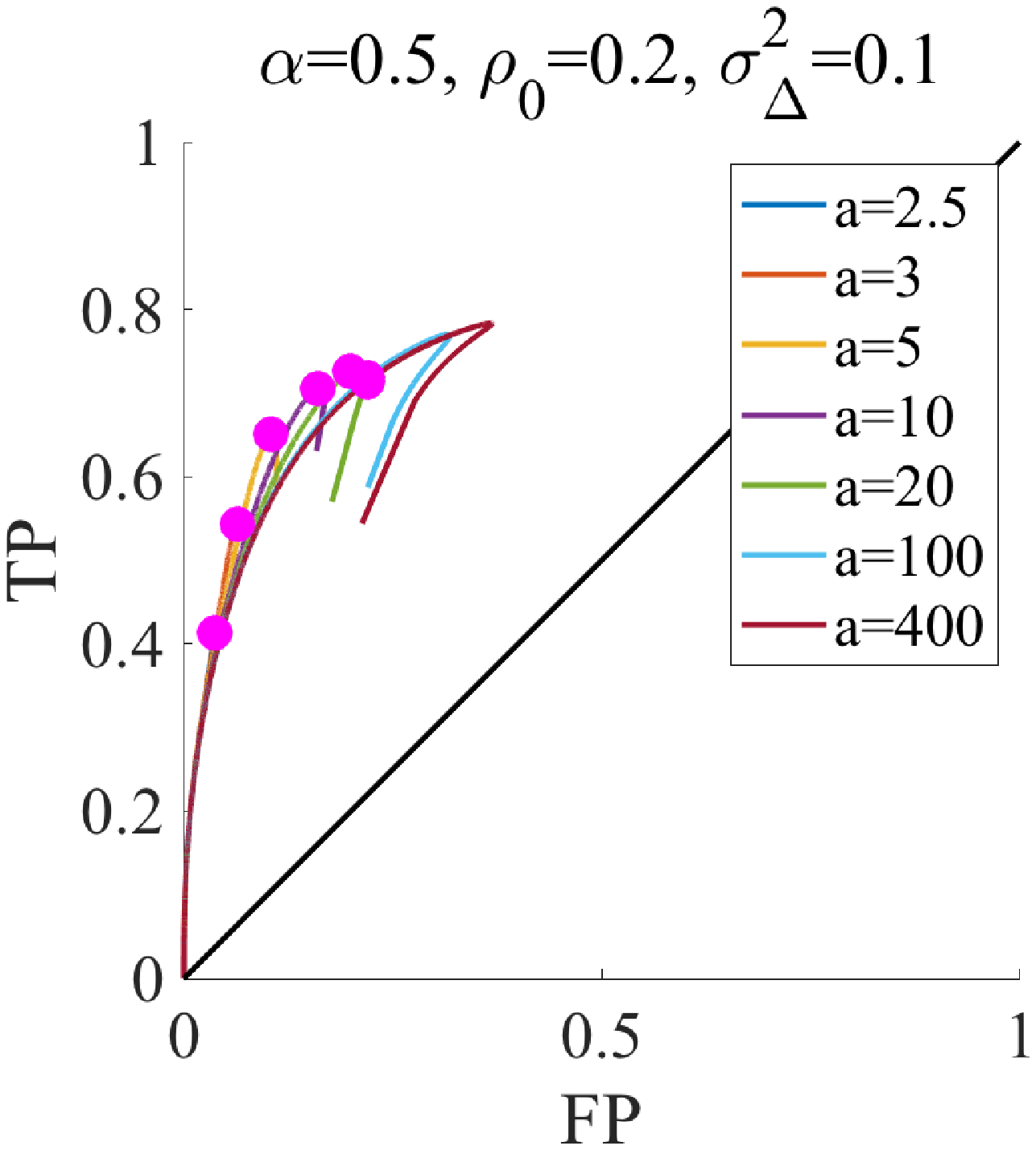}
\includegraphics[width=0.45\columnwidth]{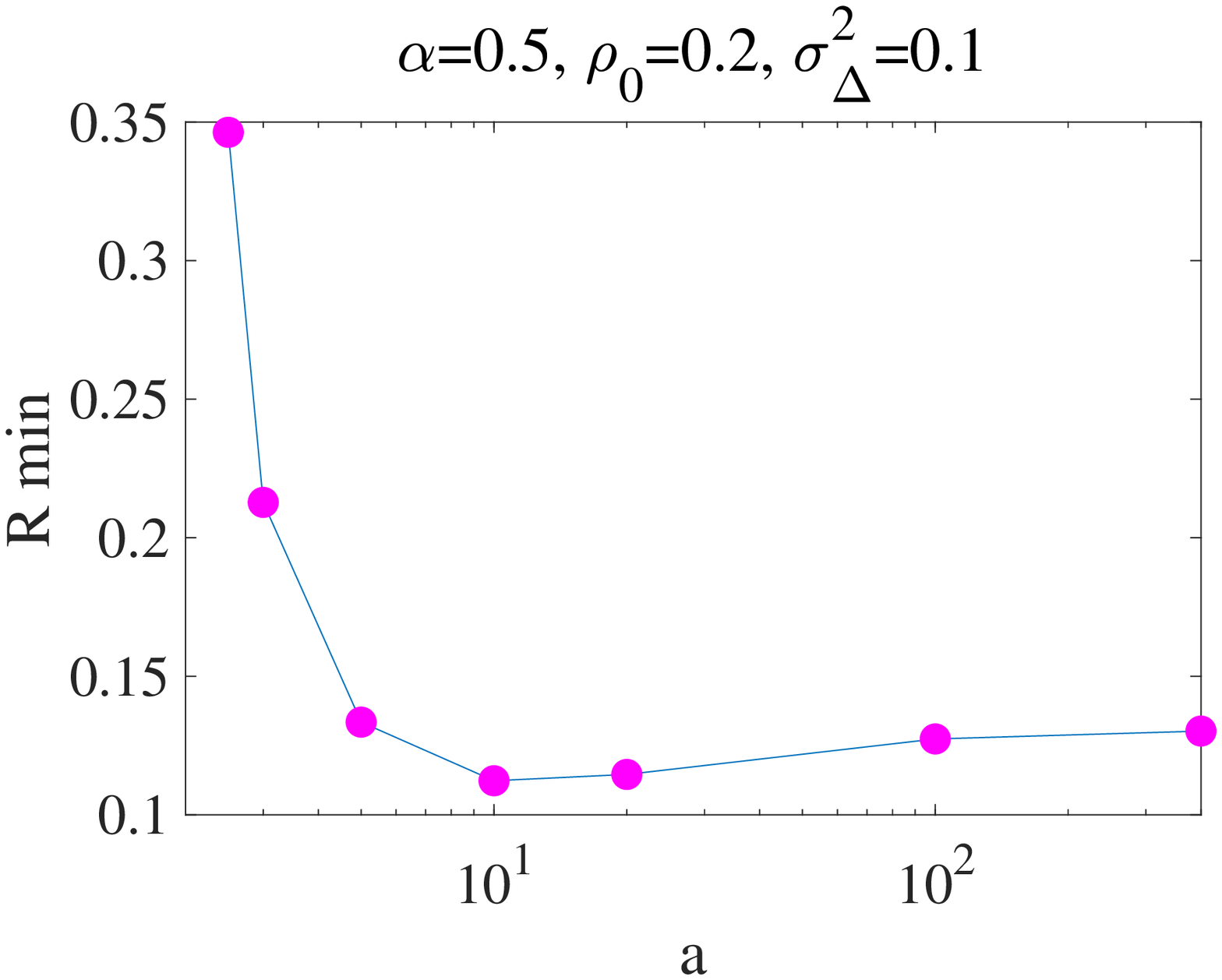}
\caption{
(Left) ROC curves when sweeping $\lambda$ at different values of $a$ for $(\alpha,\rho_0,\sigma_{\noise}^2)=(0.5,0.2,0.1)$. The minimums of $R$ on the curves, for given values of $a$ are plotted by filled magenta circles. 
(Right) The minimum value of $R$, when sweeping $\lambda$ given $a$, is plotted against $a$. The global minimum tends to be located around $a\approx 10$, which matches to the minimum location of $\MSEx$ depicted by the green diamond in the middle panel of \Rfig{PD-rho}.   
}
\Lfig{ROC-a}
\end{center}
\end{figure}
The curves are not monotonic and tend to change in the small $\lambda$ region sharply, but the locations of the minimums of $R$, depicted by filled magenta circles, tend to be in the monotonic region. To compare the values of the $R$ minimums, we plot them against $a$ in the right panel. The global minimum is located at $a\approx 10$, which matches to the minimum location of $\MSEx$, depicted by the green diamond in the middle panel of \Rfig{PD-rho}. This suggests a possibility that minimising $\MSEx$ also approximately minimises the error in the variable selection. 

To scrutinise the possibility, we show ROC curves when adaptively changing the nonconvexity parameters along the $a_{\rm IMSE}(\lambda)$ line in the $\lambda$--$a$ phase diagrams: The upper panels of \Rfig{ROC-aIMSE} are the ROC curves for $(\alpha,\sigma_{\noise}^2)=(0.5,0.0001),(0.5,0.1)$, and $(1.5,0.1)$ at $\rho_0=0.2$.
\begin{figure}[htbp]
\begin{center}
\includegraphics[width=0.32\columnwidth]{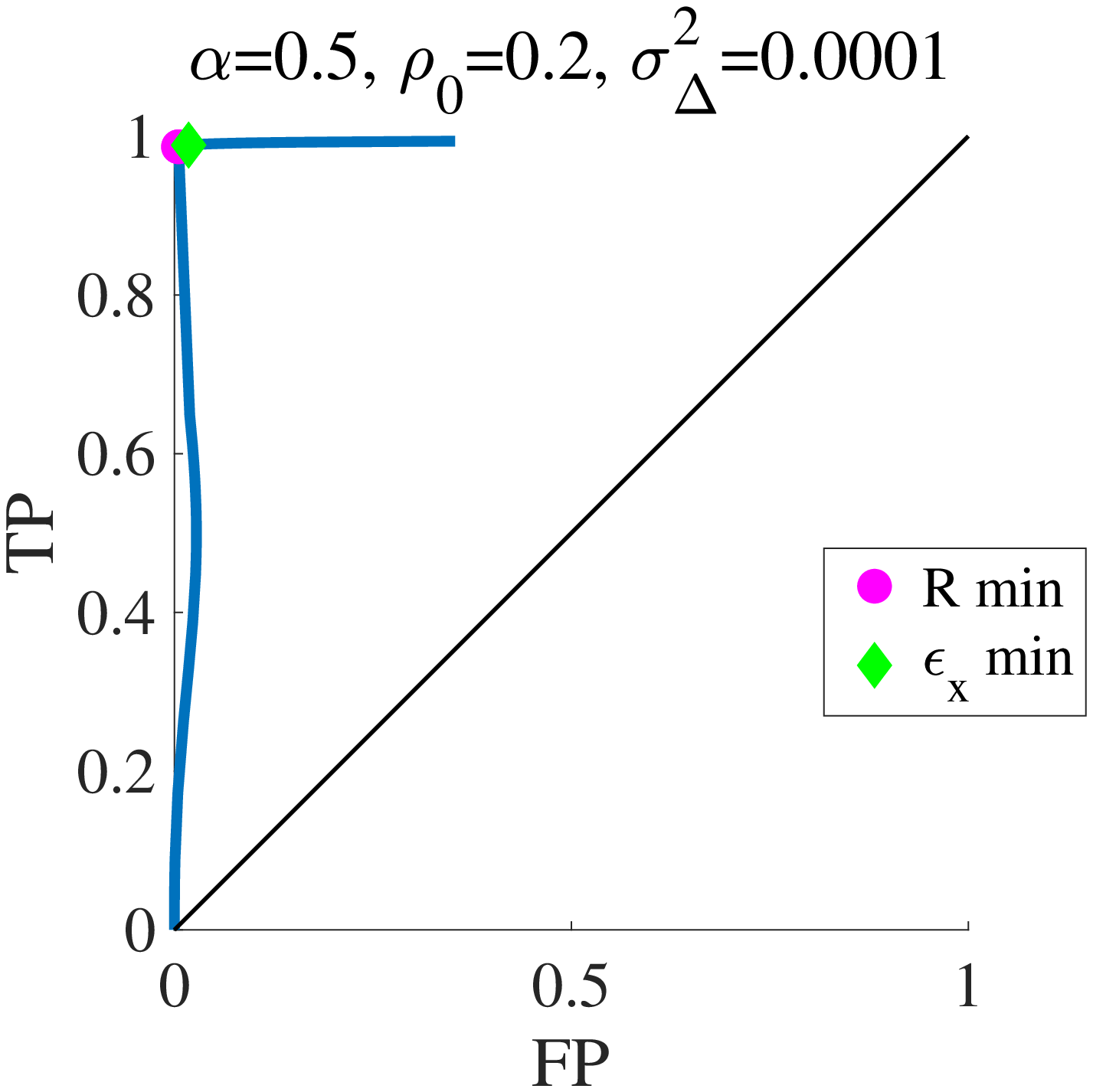}
\includegraphics[width=0.32\columnwidth]{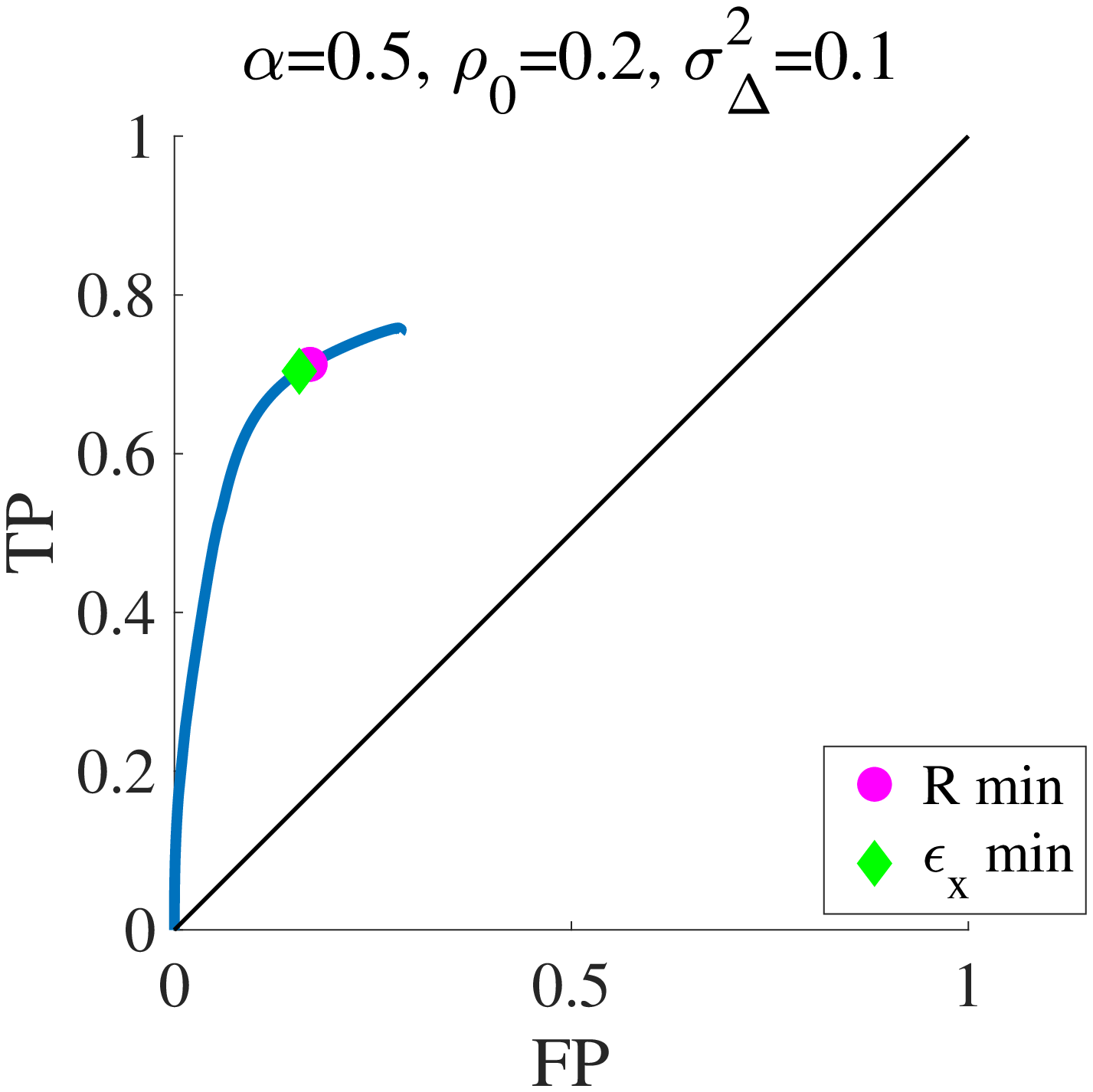}
\includegraphics[width=0.32\columnwidth]{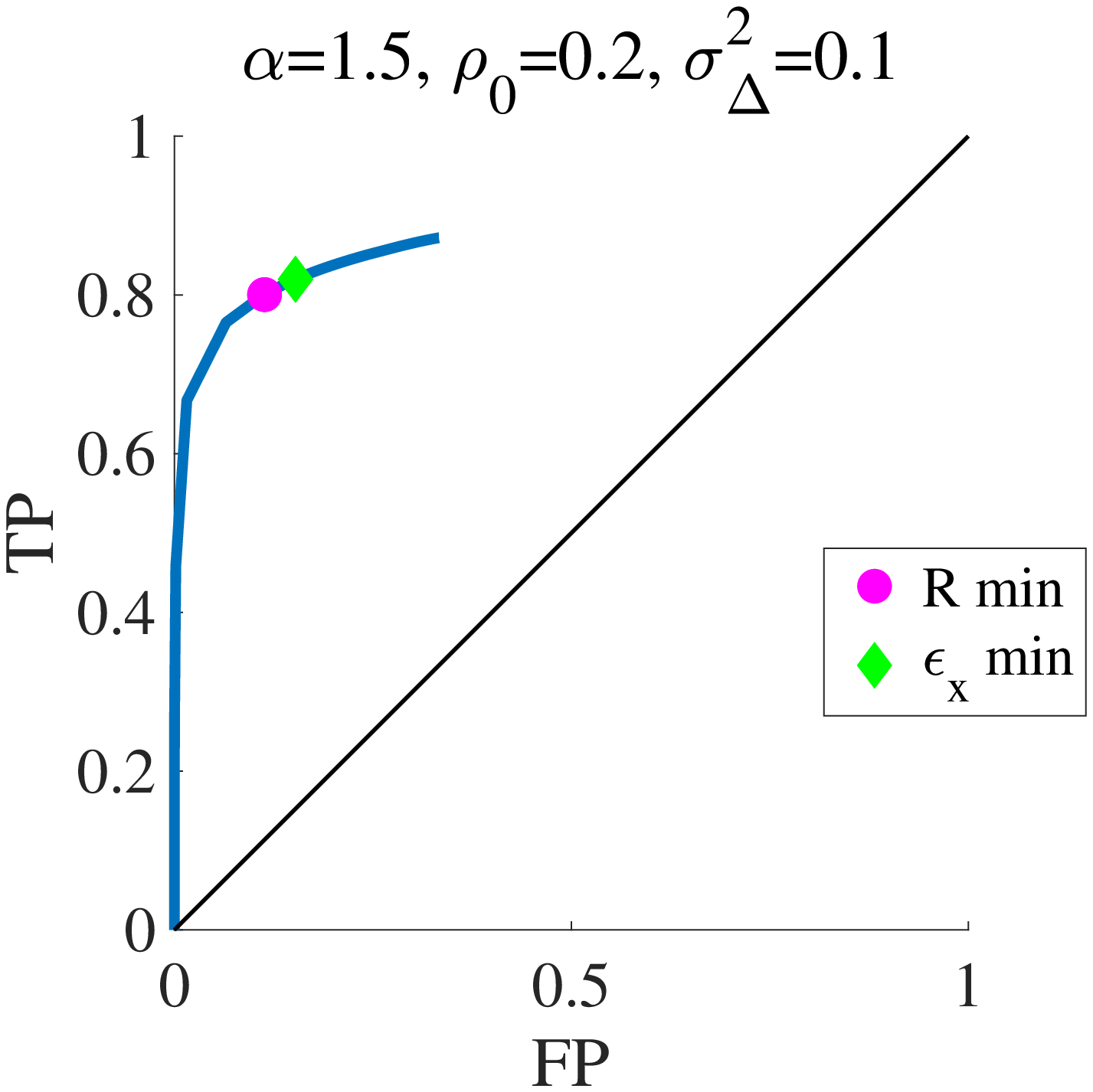}
\includegraphics[width=0.32\columnwidth]{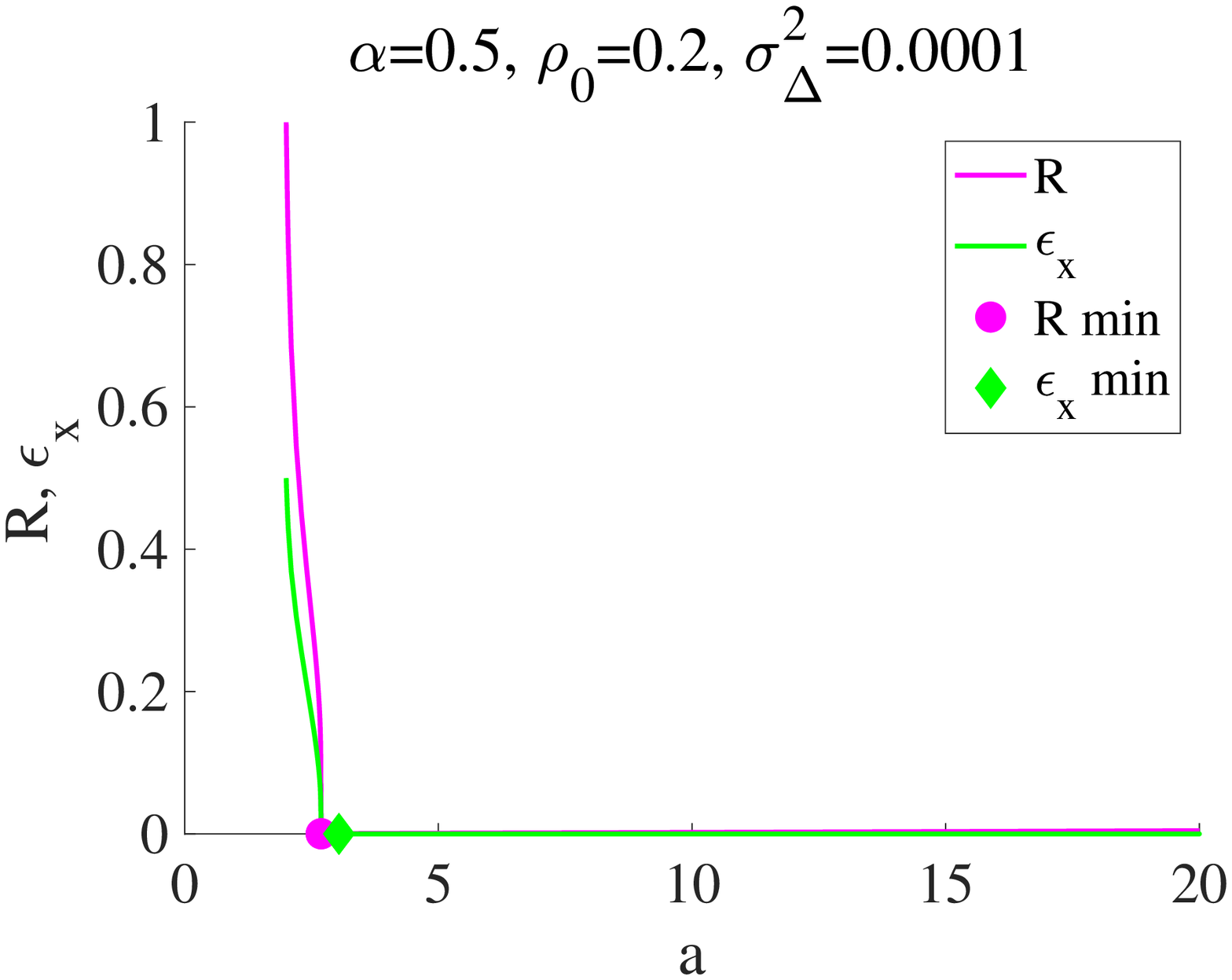}
\includegraphics[width=0.32\columnwidth]{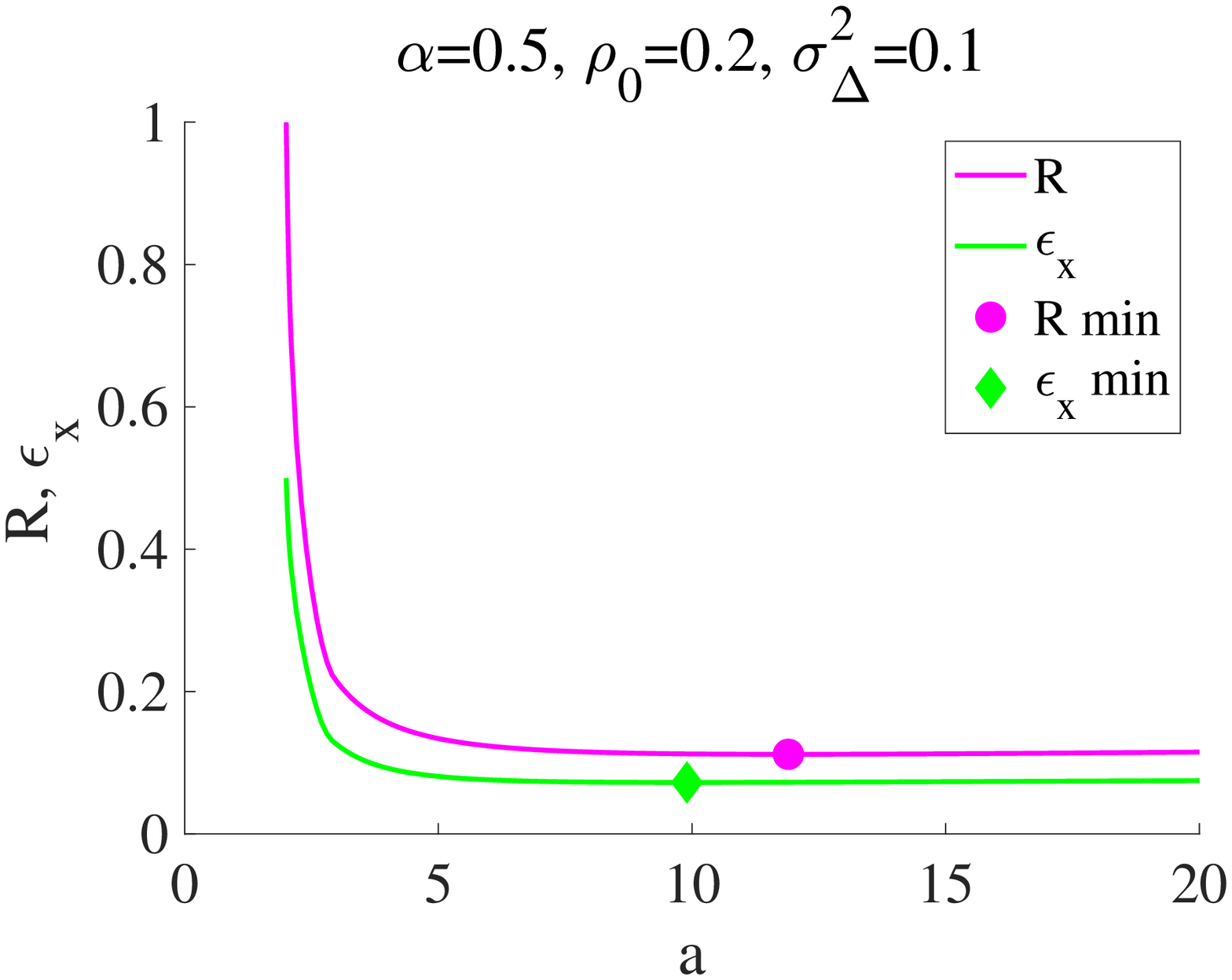}
\includegraphics[width=0.32\columnwidth]{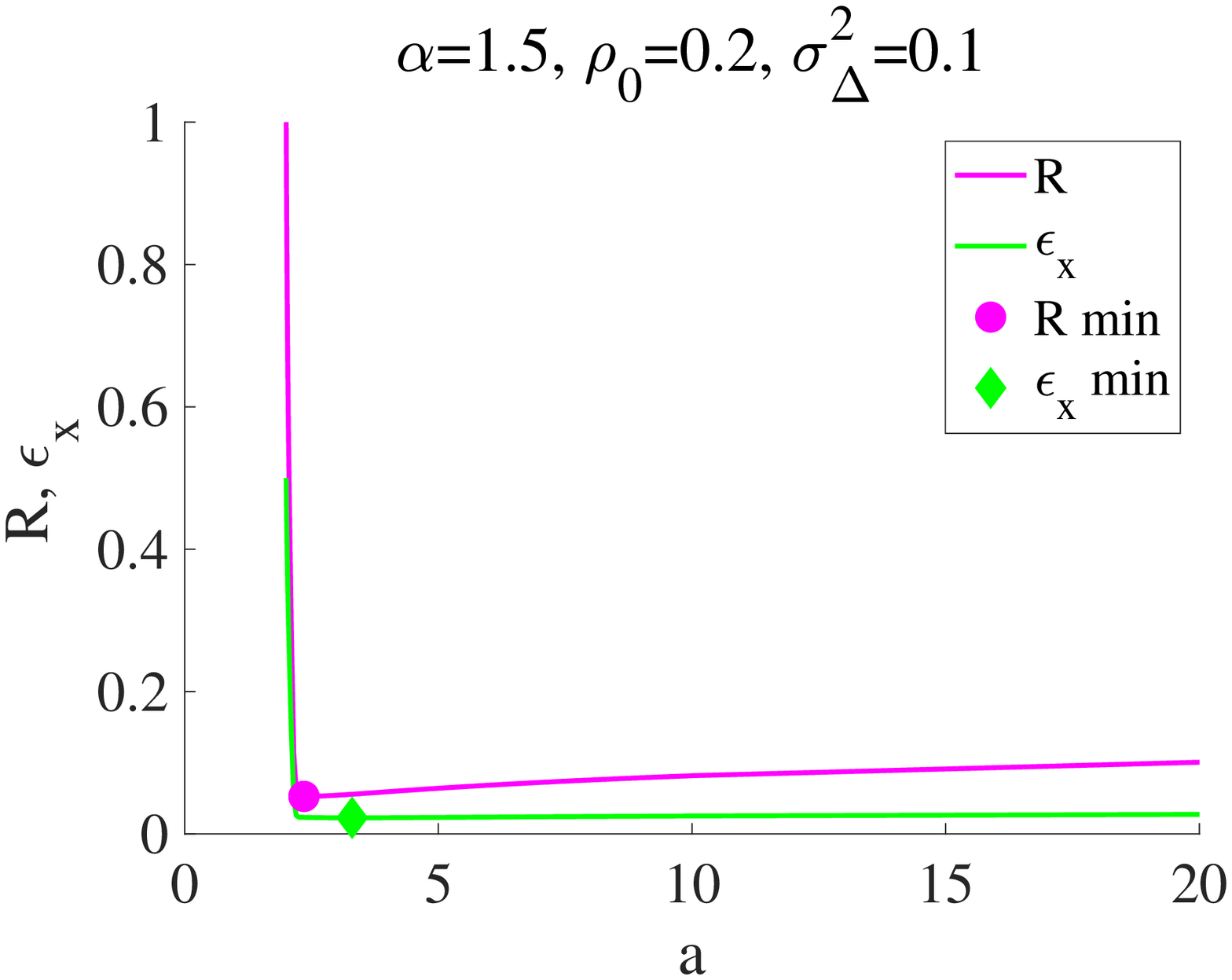}
\caption{
(Upper) 
ROC curves when changing the nonconvexity parameters along the $a_{\rm IMSE}(\lambda)$ line in the $\lambda$--$a$ phase diagrams for $(\alpha,\sigma_{\noise}^2)=(0.5,0.0001)$ (left), $(0.5,0.1)$ (middle), and $(1.5,0.1)$ (right) at $\rho_0=0.2$. The other parameters are $(\rho_0,\sigma_{\noise}^2)=(0.2,0.1)$. The minimum values of $R$ and $\MSEx$ are depicted by filled magenta circle and green diamond, respectively.
(Lower) 
Plots of $R$ and $\MSEx$ against $a$ along the $a_{\rm IMSE}$ line. The parameters are identical to the corresponding upper panels.}
\Lfig{ROC-aIMSE}
\end{center}
\end{figure}
The corresponding plots of $R$ and $\MSEx$ against $a$ are also shown in the lower panels. These figures show that the minimums of $\MSEx$ are actually close to that of $R$. As far as we have searched, similar tendency holds in other parameters. These fully support the above-mentioned possibility. Such a nice property is absent in LASSO~\cite{obuchi2016cross}, and the minimum point of $\MSEx$ in LASSO tends to give a solution with rather large FP~\footnote{In \cite{obuchi2016cross}, essentially the same analysis is done for LASSO, but a wrong terminology is used. The quantity $R$ is termed Youden's index in that study, but it is contradictory to the conventional terminology. Youden's index is another similar but different criterion for choosing an `optimal' point on ROC curve.}. Hence in the reconstruction performance of the true model, the SCAD estimator is superior to the LASSO one. 

Readers may doubt the effectiveness of this statement, because the input MSE $\MSEx$ cannot be computed for realistic settings with unknown true signals. As explained later, the input MSE has a simple linear relation to the generalisation error estimated by CV, when rows of the design matrix are uncorrelated with each other. Hence, we may minimise the CV error instead of the input MSE. 

\Rfigss{PD-noise}{PD-alpha} show that in some parameter regions there seems to be no global minimum of the input MSE at finite $a$, as for the strong noise case of $(\alpha,\rho_0,\sigma_{\noise}^2)=(0.5,0.2,1)$ in \Rfig{PD-noise} and the dense signal case $(\alpha,\rho_0,\sigma_{\noise}^2)=(0.5,0.4,0.1)$ in \Rfig{PD-rho}. To examine those cases, we plot relevant quantities when changing $a$ and $\lambda$ again along the $a_{\rm IMSE}(\lambda)$ line in \Rfig{ROC-aIMSE-nomin}. The left panels show the plots of $TP$ and $FP$ against $a$, the middle panels display the plots of $R$ and $\MSEx$ against $a$, and the right panels give the associated ROC curves.  
\begin{figure}[htbp]
\begin{center}
\includegraphics[width=0.32\columnwidth]{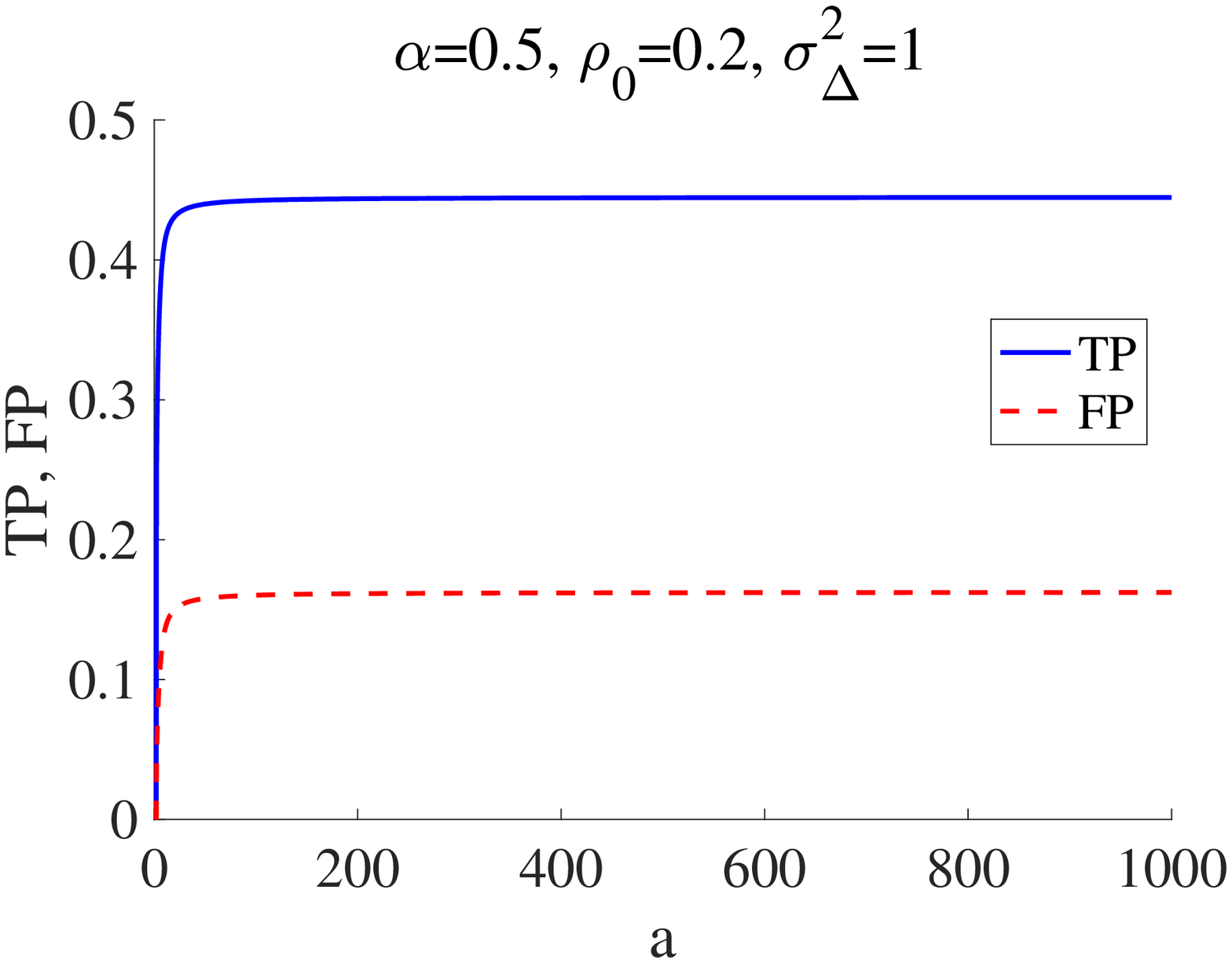}
\includegraphics[width=0.32\columnwidth]{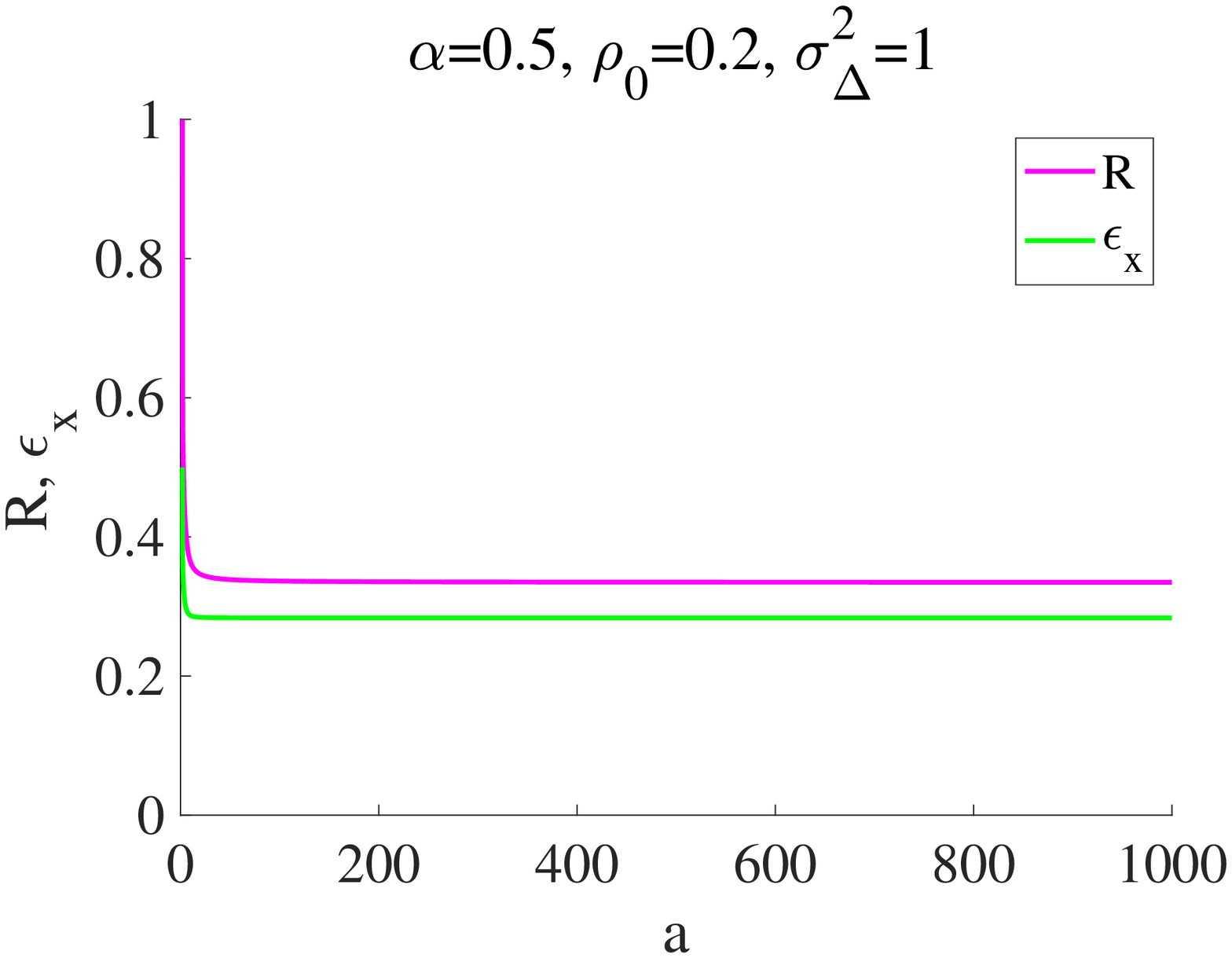}
\includegraphics[width=0.32\columnwidth]{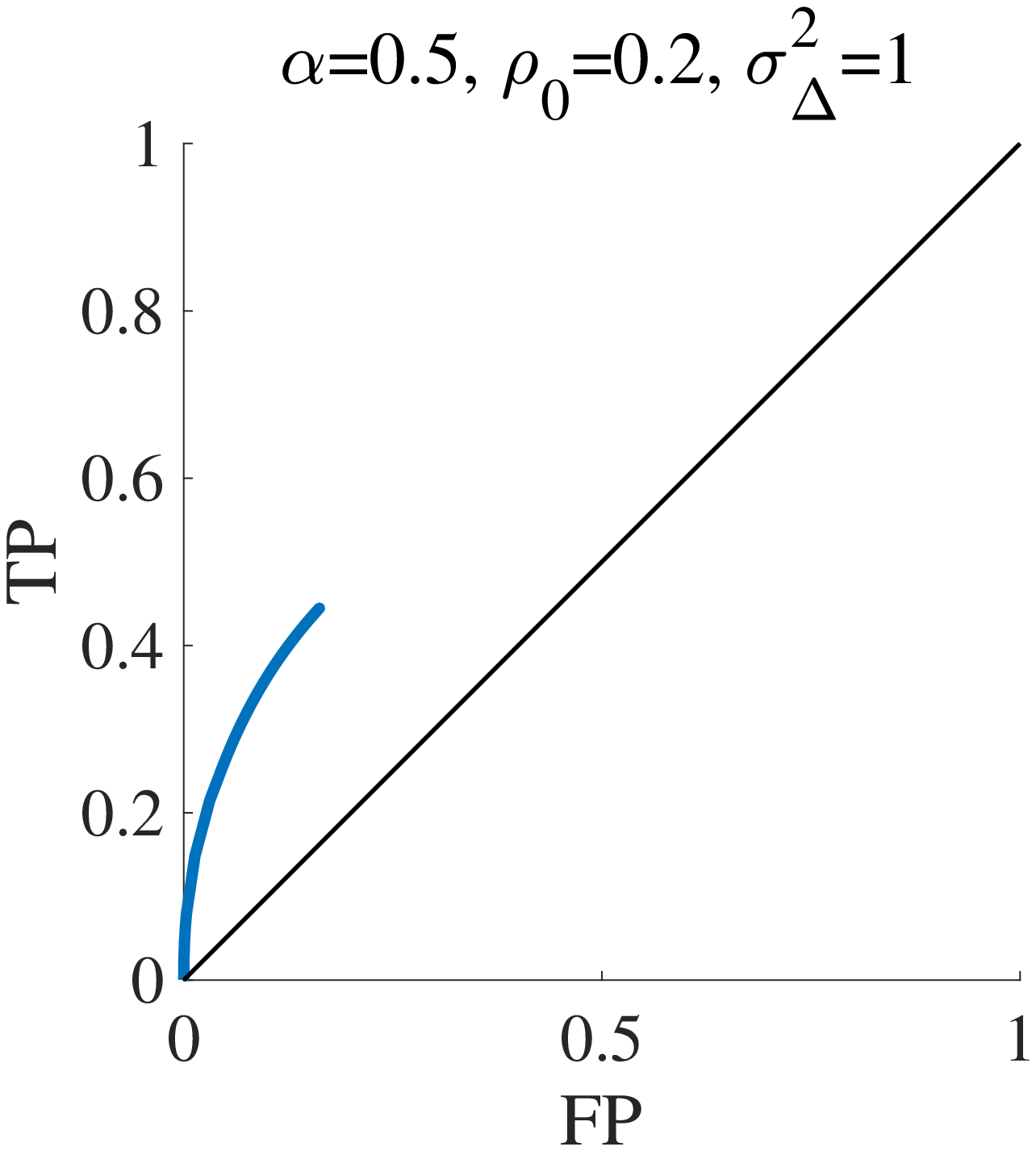}
\includegraphics[width=0.32\columnwidth]{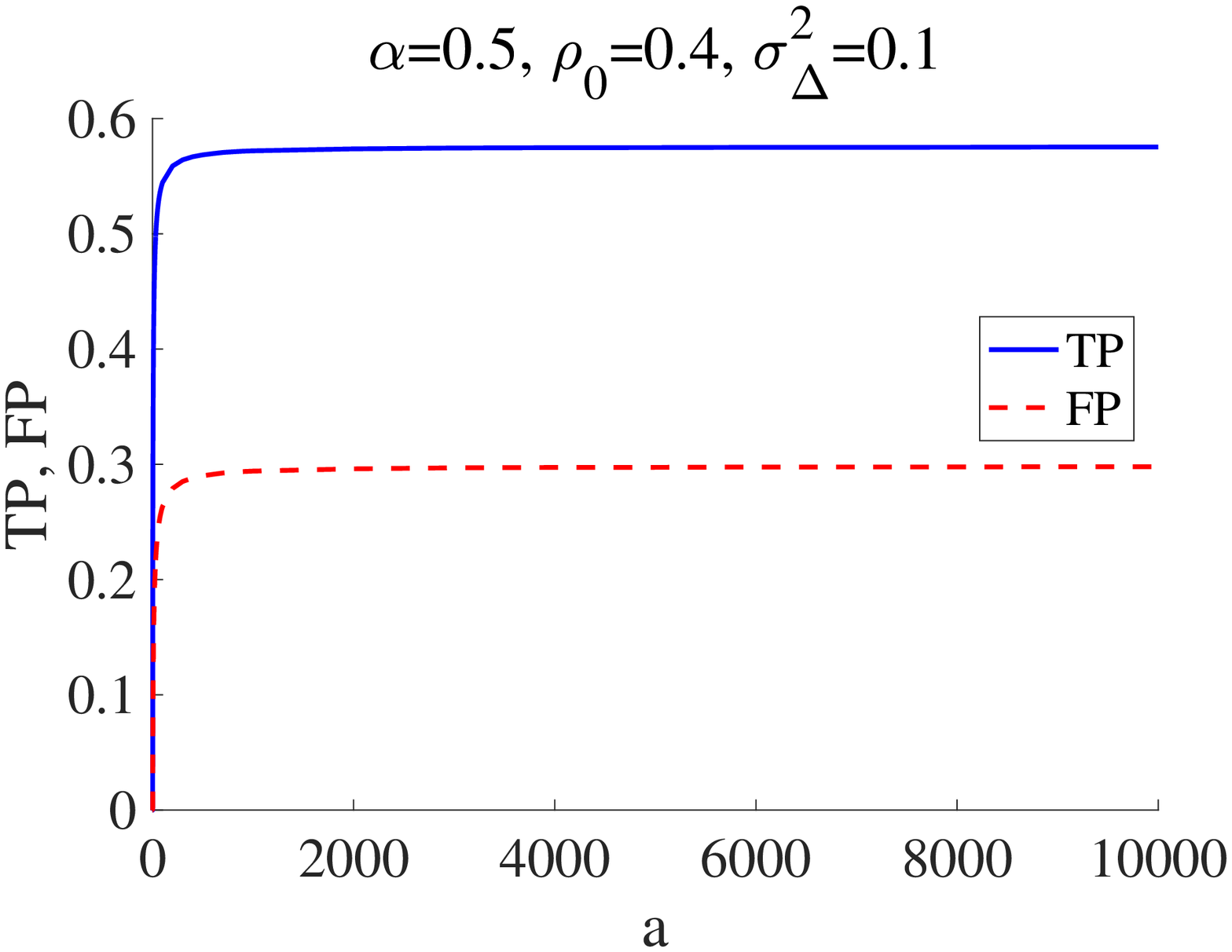}
\includegraphics[width=0.32\columnwidth]{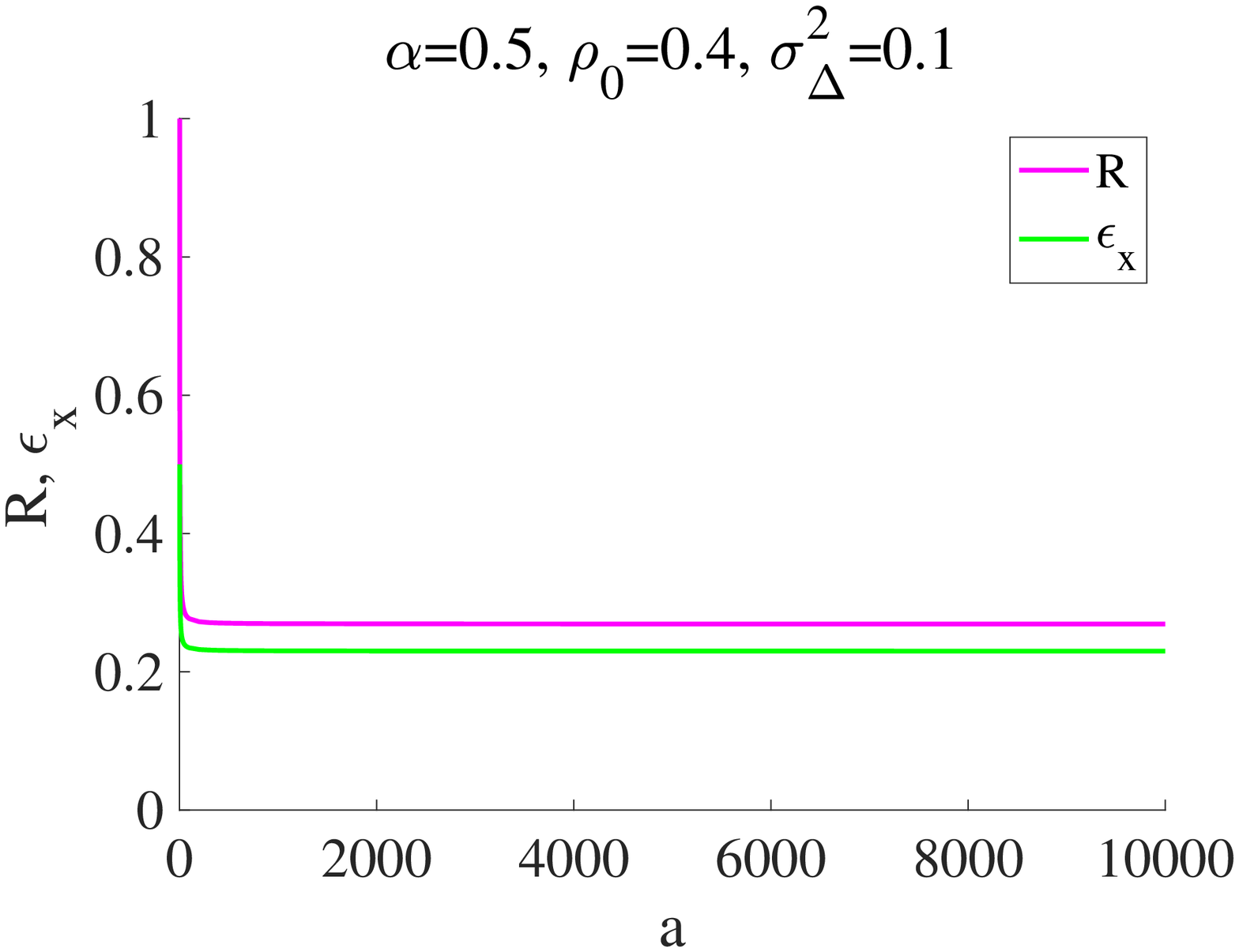}
\includegraphics[width=0.32\columnwidth]{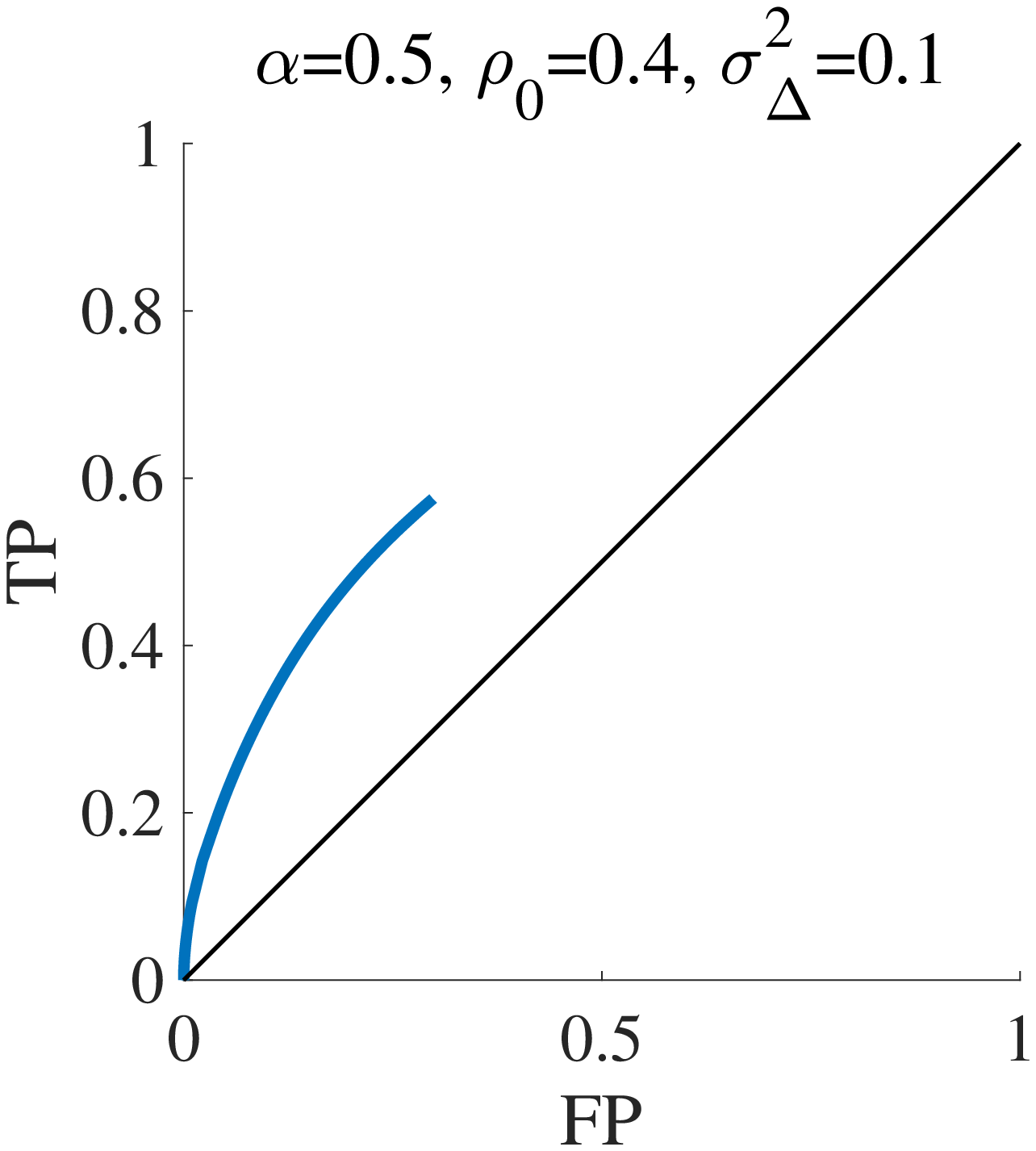}
\caption{
(Left) Plots of $TP$ and $FP$ against $a$ along the $a_{\rm IMSE}(\lambda)$ line.
(Middle) Plots of $R$ and $\MSEx$ against $a$ along the $a_{\rm IMSE}(\lambda)$ line.
(Right) The associated ROC curves.
The upper row is for the strong noise case of $(\alpha,\rho_0,\sigma_{\noise}^2)=(0.5,0.2,1)$ corresponding to the right panel of \Rfig{PD-noise}, while the lower row is for the dense signal case $(\alpha,\rho_0,\sigma_{\noise}^2)=(0.5,0.4,0.1)$ corresponding to the right panel of  \Rfig{PD-rho}.
In both the cases, all the quantities of $TP,FP,R$, and $\MSEx$ behave monotonically with respect to $a$, and seem to converge to finite values in the LASSO limit $a\to \infty$.
}
\Lfig{ROC-aIMSE-nomin}
\end{center}
\end{figure}
All the quantities of $TP,FP,R$, and $\MSEx$ show monotonic behaviours with respect to $a$, and seem to converge to finite values in the LASSO limit $a\to \infty$. The minimums of $R$ and $\MSEx$ would be thus obtained by LASSO, with the optimised $\lambda$. These observations imply that LASSO is sufficient for difficult cases with strong noises or dense signals. This also implies that it is difficult to determine a good value of $a$ to find the least $\MSEx$ solution given a dataset prior to actual analyses, because it strongly depends on the noise strength or the signal density.

\section{Approximate formula for cross-validation}\Lsec{Approximate}
In this section, we derive an approximate formula for the leave-one-out (LOO) CV error. If the dataset size $M$ is large enough, the difference between the estimators of the full and LOO datasets is considered to be small, and it is expected that those two estimators can be connected in a perturbative manner. We concretise this idea below.

The estimator without the $\mu$th data in \Req{original} is, hereafter, termed $\mu$th LOO estimator, and the explicit formula is given by:
\be
&&
\hat{\V{x}}^{\Bs \mu}(\eta,D_M)=
\argmin_{\V{x}}
\lbb 
\frac{1}{2}\sum_{\nu (\neq \mu)} \lb y_{\nu}-\sum_{i}A_{\nu i}x_i \rb^2 +J(\V{x};\eta)
\rbb
.
\Leq{LOO estimator}
\ee
The LOO CV error (LOOE) is accordingly defined as:
\be
\LOOE(\eta,D_M)
=
\frac{1}{2M}\sum_{\mu=1}^{M}( y_{\mu} - \V{a}_{\mu}^{ \top } \hat{\V{x}}^{\Bs \mu}(\eta,D_M))^2,
\Leq{LOOE}
\ee
where $\V{a}_{\mu}^{\top}=(A_{\mu 1},\cdots,A_{\mu N})$ is the $\mu$th row vector of $A$. The LOOE is an estimator for the generalisation error or extra-sample error, defined as:
\be
\GE(\eta,D_M) \equiv \int d y_{\rm new}d \V{a}_{\rm new}
P(y_{\rm new},\V{a}_{\rm new})
\frac{1}{2}( y_{\rm new} - \V{a}_{\rm new}^{ \top } \hat{\V{x}}(\eta,D_M))^2,
\Leq{GE}
\ee
where $\{ y_{\rm new},\V{a}_{\rm new} \}$ represents a new data sample, and $P(y_{\rm new},\V{a}_{\rm new})$ denotes its distribution. In our setting, the distribution corresponds to the i.i.d. process described around \Req{generative}, and it is analytically shown that $\GE$ has a direct connection to $\MSEx$ as:
\be
\GE(\eta,D_M)=\frac{1}{\alpha }\MSEx\lb \hat{\V{x}} \lb \eta,D_M \rb|\V{x}^0 \rb+\frac{1}{2}\sigma_{\noise}^2.
\Leq{GE-LOOE}
\ee
Hence, we can estimate the input MSE from the LOOE. Note that the sufficient condition for \Req{GE-LOOE} is that both the noise components and the rows of the design matrix are zero-mean and uncorrelated; correlations in the signal vector $\V{x}^0$ may exist because they do not affect \Req{GE-LOOE}.

Owing to sparse priors, the variables in the estimator are separated in two types. Some variables are set to zero and the others take non-zero values. We call the former {\em inactive} variables and the latter {\em active} variables. The index set of the inactive variables, or inactive set, is denoted by $S_I=\{i|\hat{x}_{i}=0\}$, while one of the active variables, or active set, is $S_A=\{i| \hat{x}_{i} \neq 0\}$. The active (inactive) components of a vector $\V{x}$ are formally expressed as $\V{x}_{S_A}(\V{x}_{S_I})$. For any matrix $X$, we use double subscripts in the same manner and introduce the symbol $\Wc$ meaning all the components in the respective dimension. For example, for an $N\times N$ matrix $X$, $X_{S_A S_I}$ and $X_{\Wc S_I}$ denote $X$'s sub-matrices having row components of $S_A$ and all, respectively, while their column components are commonly of $S_I$.

\subsection{Derivation}\Lsec{Derivation}
The basic assumption to derive the approximate formula is that the active set is `common' between the full and LOO estimators. Although this assumption is literally not true, we numerically confirmed that this approximately holds in the RS region. In other words, the change of the active set is small enough compared to the size of the active set itself, when considering the LOO operation under the situation with large $N$ and $M$. Moreover, in the LASSO case, it has been shown that the contribution of the active set change vanishes in a limit $N, M\to \infty$, keeping $\alpha=M/N=O(1)$~\cite{obuchi2016cross}. It is expected that the same holds in the present problem. Hence, we adopt this assumption in the following derivation. We also note that this assumption is not applicable to the RSB region.

Once assuming the active set is known and common between the full and LOO systems, we can easily get the determining equations for the coefficients in the active set, by differentiating the cost function with respect to $\V{x}_{S_A}$, yielding:
\be
\lb \lb A_{\Wc S_{A}} \rb^{^{\top}} A_{\Wc S_{A}} \rb \hat{\V{x}}_{S_A}-\lb A_{\Wc S_{A}}\rb^{T}\V{y}+\nabla J(\hat{\V{x}}_{S_A};\eta)=0,
\Leq{full eq}
\ee
for the full system and:
\be
\lb \lb A^{\Bs \mu}_{\Wc S_{A}} \rb^{\top} A^{\Bs \mu}_{\Wc S_{A}} \rb \hat{\V{x}}^{\Bs \mu}_{S_A}-\lb A^{\Bs \mu}_{\Wc S_{A}}\rb^{T} \V{y}^{\Bs \mu}+\nabla J(\hat{\V{x}}^{\Bs \mu}_{S_A};\eta)=0,
\Leq{LOO eq}
\ee
for the LOO system. 

Let us denote $\V{d}=\hat{\V{x}}- \hat{\V{x}}^{\Bs \mu}$ and expand \Req{LOO eq} with respect to $\V{d}$ up to the first order. Erasing some terms using \Req{full eq}, and solving the remaining expression with respect to $\V{d}$, we obtain an equation of $\V{d}_{S_A}$ as:
\be
\V{d}_{S_A} \approx 
(y_{\mu}-\V{a}_{\mu}^{\top} \hat{\V{x}})\lb  
\lb  A^{\Bs \mu}_{\Wc S_{A}} \rb^{\top} A^{\Bs \mu}_{\Wc S_{A}}
+ \lb \HESS  J (\hat{\V{x}}_{S_A};\eta)\rb_{S_{A}S_{A}} 
\rb^{-1}
\V{a}_{\mu},
\Leq{d}
\ee
Using \Req{d}, we can connect the residuals of the LOO and full systems as:
\be
&&
\hspace{-2cm}
y_{\mu}-\V{a}_{\mu}^{\top}\hat{\V{x}}^{\Bs \mu}
=
y_{\mu}-\V{a}_{\mu}^{\top}(\hat{\V{x}}-\V{d})
\no \\ &&
\hspace{-2cm}
\approx
\lb 1+\lb \V{a}_{\mu} \rb_{S_A}^{\top} \lb  \lb  A^{\Bs \mu}_{\Wc S_{A}} \rb^{\top} A^{\Bs \mu}_{\Wc S_{A}}+ \lb \HESS  J (\hat{\V{x}}_{S_A};\eta)\rb_{S_{A}S_{A}} \rb^{-1}   \lb \V{a}_{\mu} \rb_{S_A}\rb
(y_{\mu}-\V{a}_{\mu}^{\top}\hat{\V{x}}).
\Leq{residuals}
\ee
Using the relation $\lb \lb A^{\Bs \mu} \rb^{\top} A^{\Bs \mu} \rb=\lb A^{\top}A-\V{a}_{\mu}\V{a}_{\mu}^{\top} \rb$ and the Woodbury matrix inversion formula, we finally get
\be
&&
\LOOE
\approx
\frac{1}{2M}
\sum_{\mu=1}^{M}
\LOOfactor_{\mu}
\lb y_{\mu}-\V{a}_{\mu}^{\top}\hat{\V{x}} \rb^2
,
\Leq{CVformula}
\ee
where
\be
\LOOfactor_{\mu}=
\lb 1-\lb \V{a}_{\mu} \rb_{S_A}^{\top} \lb  \lb  A_{\Wc S_{A}} \rb^{\top} A_{\Wc S_{A}}
+ 
\lb \HESS  J (\hat{\V{x}}_{S_A};\eta)\rb_{S_{A}S_{A}} \rb^{-1}   \lb \V{a}_{\mu} \rb_{S_A}\rb^{-2}.
\Leq{denominator}
\ee
The righthand side of \Req{CVformula} can be computed only from the full solution, enabling an approximate evaluation of the LOOE, without literally conducting CV. The error bar can be put as the standard deviation among all the terms in \Req{CVformula} divided by $\sqrt{M}$\footnote{Note that the main text description about the error bar of the approximate CV error in \cite{obuchi2016cross} is inconsistent with this, but this one is the correct one. Although the text description is incorrect, the experimentally reported error bars in~\cite{obuchi2016cross} are correct and consistent with this.}. This is convincing because each term of \Req{CVformula} gives an independent estimator to the generalisation error \Req{GE} and hence its error bar can be given as the standard error. Numerical experiments below show that this definition gives a reasonable error bar.

In the case of LASSO, the Hessian of the penalty term is identically zero, $\HESS  J_{\rm LASSO}=0$, meaning that \Req{CVformula} comes back to the `approximation 1' in~\cite{obuchi2016cross}. For SCAD, the Hessian takes the following form:
\be
&&
\lb \HESS  J_{\rm SCAD}(\hat{\V{x}};\eta=\{\lambda,a\})\rb_{ij}=\frac{1}{1-a}\delta_{ij}I(\lambda < |\hat{x}_i| \leq a\lambda),
\ee
where $I(\mathrm{statement})$ denotes the indicator function, giving $1$ if the statement is true and $0$ otherwise.

\section{Numerical experiments and numerical codes}\Lsec{Numerical}
Here we present numerical experiments. To obtain the SCAD estimator, we use the CD algorithm because it is common and stable. We implement this using C language, while the approximate CV formula is implemented as a raw code in MATLAB\textsuperscript{\textregistered}. Hence, it is not necessarily fair to compare the computational time in the literal and approximate CVs, which are computed by \Req{LOOE} and \Req{CVformula} respectively, conducted below as a part of experiments. However, even in this comparison there is a meaningful difference in the computational time. When showing the computational time, we fix our experimental environment which uses a single CPU of 3.3 GHz Intel Core i7.

In a single step of the CD algorithm, we update all the components of $\V{x}$ in a random order. To judge the convergence of the CD algorithm, we monitor the difference between the estimate $\hat{\V{x}}^{(t)}$ at the step $t$ and the previous one $\hat{\V{x}}^{(t-1)}$. If all the component-wise differences $\{ d_{i}=|\hat{x}_i^{(t)}-\hat{x}_i^{(t-1)}| \}_i$ are smaller than a threshold value $\delta$, then the algorithm stops; otherwise it continues. We set the threshold value as $\delta=10^{-10}$ in all the experiments below.

\subsection{Simulated dataset}\Lsec{Simulated dataset}
In this subsection, we conduct experiments using simulated datasets. The main purpose is to confirm analytical predictions and to examine the accuracy of the approximate formula.  Our simulated datasets are generated by the process described around \Reqs{generative}{Bernoulli-Gauss}. The signal power is set to be $\sigma_x^2=1/\rho_0$, matching with \Rsec{Phase diagram}. 

\subsubsection{Consistency check of the replica solution}\Lsec{Consistency check of}
To check the accuracy of the replica result and to examine the finite-size effect, we first plot the input and output MSEs against $\lambda$, given $a$ for different sizes. The plots for $(\alpha,\rho_0,\sigma_{\noise}^2,a)=(0.5,0.2,0.1,3)$ and $(1.5,0.2,0.1,4)$ are given in \Rfig{numcheck-eps} as example cases.
\begin{figure}[htbp]
\begin{center}
\includegraphics[width=0.45\columnwidth]{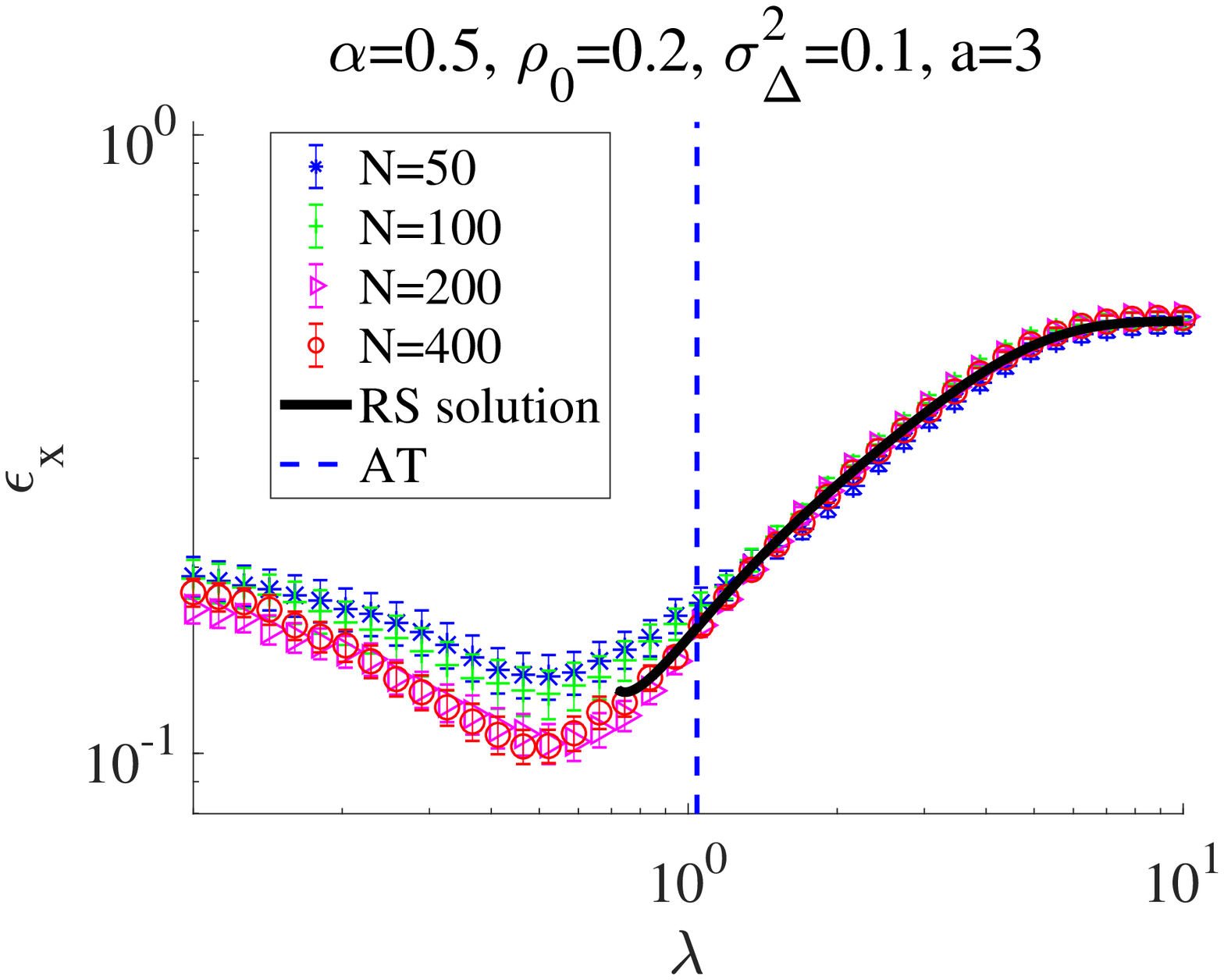}
\includegraphics[width=0.45\columnwidth]{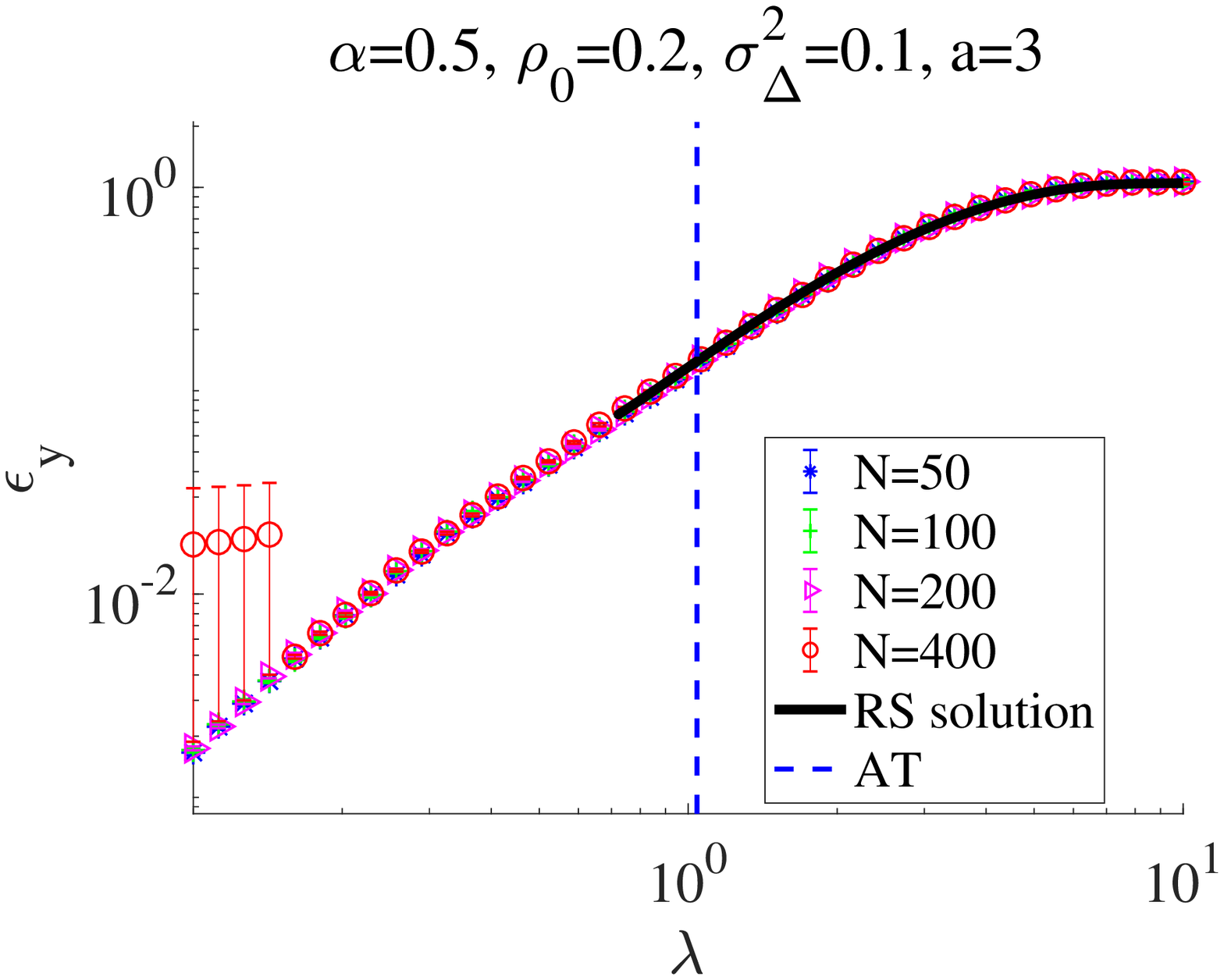}
\includegraphics[width=0.45\columnwidth]{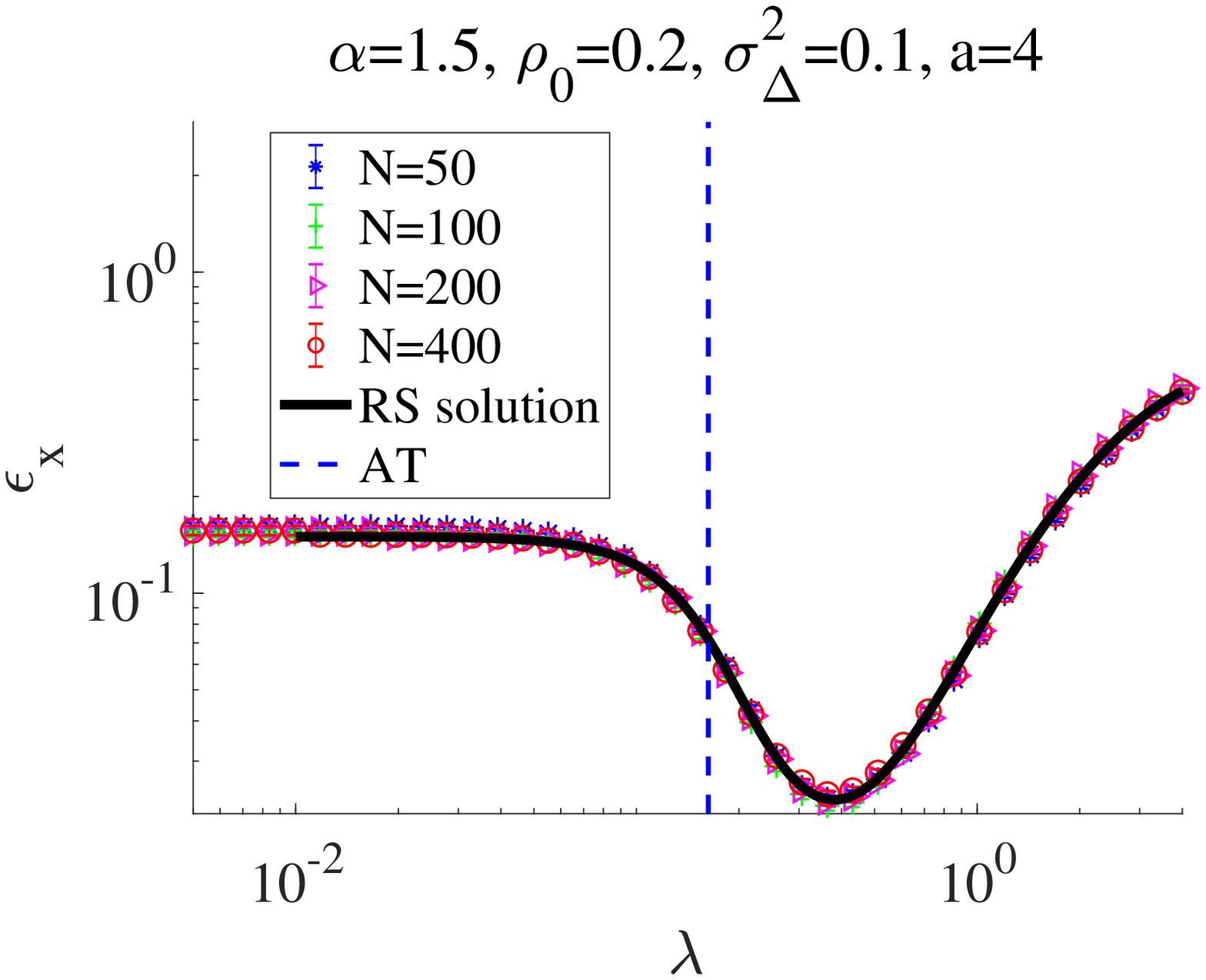}
\includegraphics[width=0.45\columnwidth]{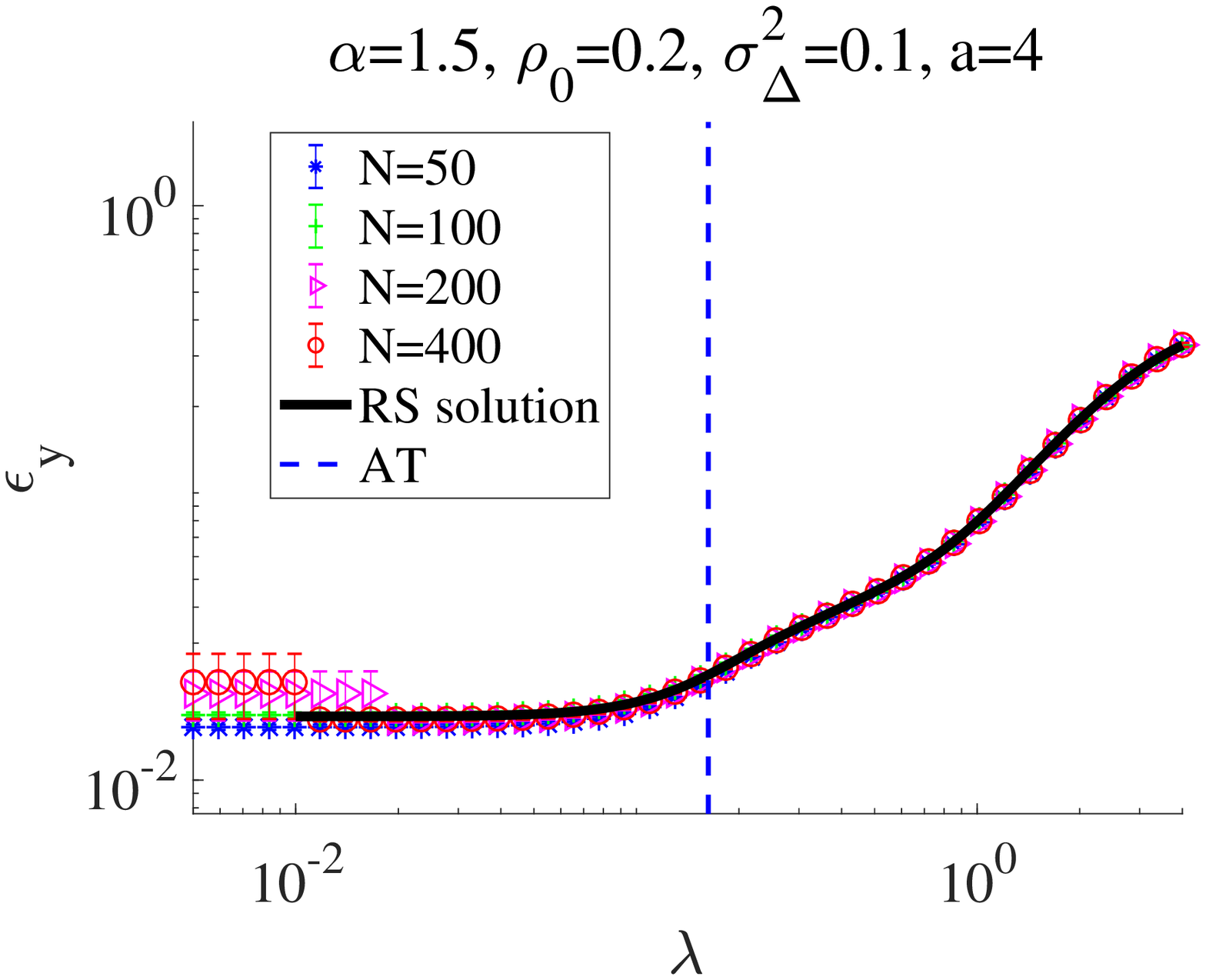}
\caption{
Plots of the input MSE (left) and the output MSE (right) against $\lambda$ for $(\alpha,\rho_0,\sigma_{\noise}^2,a)=(0.5,0.2,0.1,3)$ (upper) and $(\alpha,\rho_0,\sigma_{\noise}^2,a)=(1.5,0.2,0.1,4)$ (lower). The black thick curves and the colour markers denote the analytical and numerical results, respectively. The left end point of the analytical curve corresponds to the existence limit of the RS solution. The vertical blue dashed line represents the AT instability point below which the RS solution is unstable. The agreement between the analytical and numerical results is excellent in the RS region. The numerical results are obtained by the annealing with respect to the amplitude parameter $\lambda$, explaining the regularity of the numerical results even below the AT instability point. 
}
\Lfig{numcheck-eps}
\end{center}
\end{figure}
The black thick curves and the colour markers denote the analytical and numerical results, respectively. For the numerical data, the average over different samples of the set $\{\DM,\V{x}^0,\V{\noise}\}$ is taken; the sample numbers are $200,100,100,50$ for $N=50,100,200,400$, respectively; the error bars are given as the standard error among the samples. The vertical dashed line represents the AT instability point below which the RS solution is unstable. Focusing only on the RS region, we can find that the finite-size effect is quite weak, and the numerical results show an excellent agreement with the analytical ones, justifying our analytical solutions. In this experiment, even below the AT point, the numerical results show a strong regularity. This is because the solutions are obtained by gradually changing $\lambda$ from large to small values, and hence these solutions below the AT point are, in some sense, continuously connected to the ones above the AT point. We term this scheme {\it $\lambda$ annealing}, which can be considered as a part of the {\it nonconvexity control} proposed in~\cite{sakata2019perfect}. We warn that the solution path obtained by the $\lambda$ annealing is very atypical below the AT point, as implied in \Rsec{Accuracy of the}.

As another check of the RS solution's consistency, we also draw ROC curves by numerical experiments. In \Rfig{numcheck-ROC}, we give the ROC curves along the $a_{\rm IMSE}$ line for $(\alpha,\rho_0,\sigma_{\noise}^2)=(0.5,0.2,0.1)$ and $(\alpha,\rho_0,\sigma_{\noise}^2)=(0.5,0.4,0.1)$, which correspond to the upper middle panel of \Rfig{ROC-aIMSE} and the lower middle panel of \Rfig{ROC-aIMSE-nomin}, respectively. 
\begin{figure}[htbp]
\begin{center}
\includegraphics[width=0.48\columnwidth]{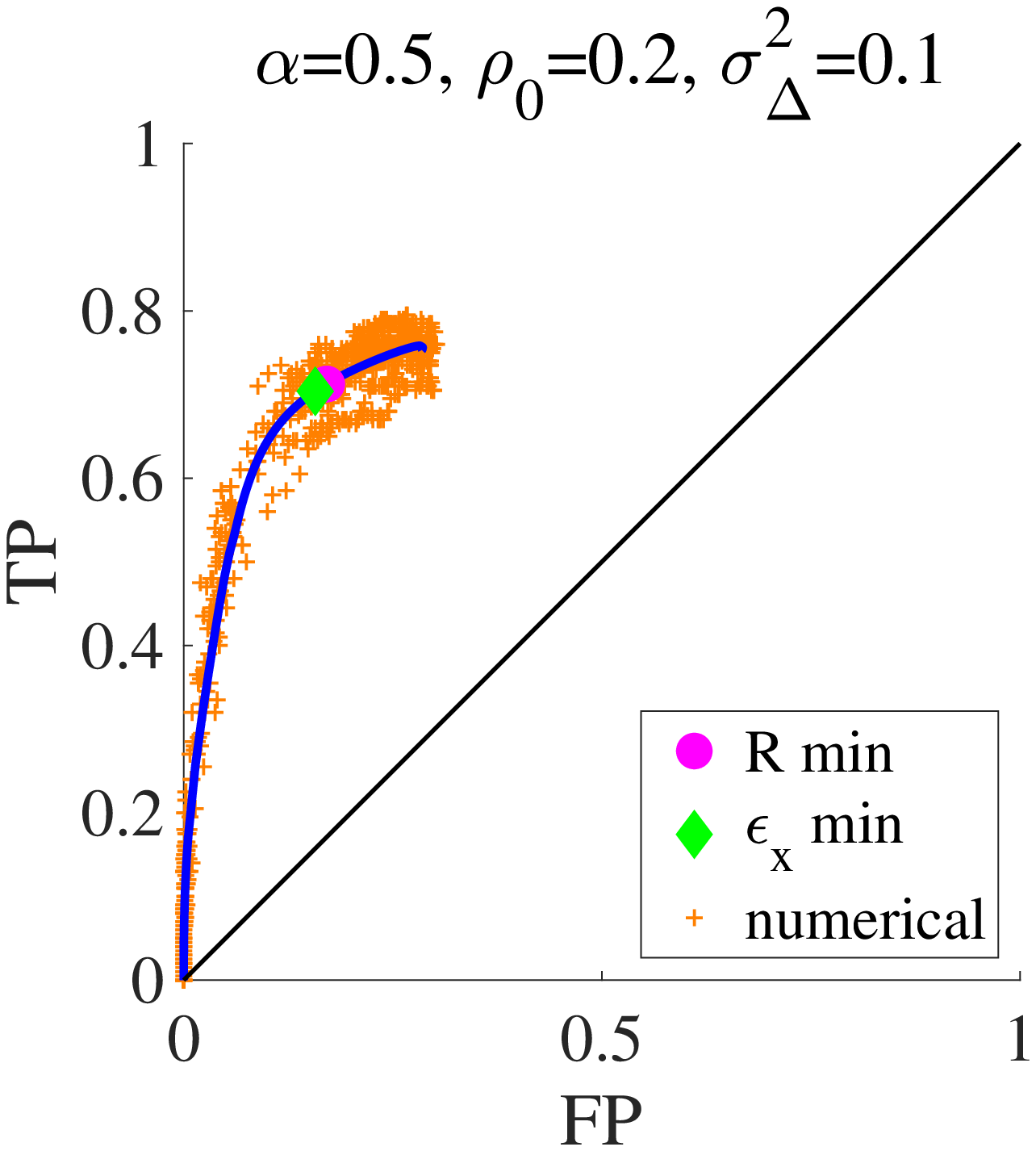}
\includegraphics[width=0.48\columnwidth]{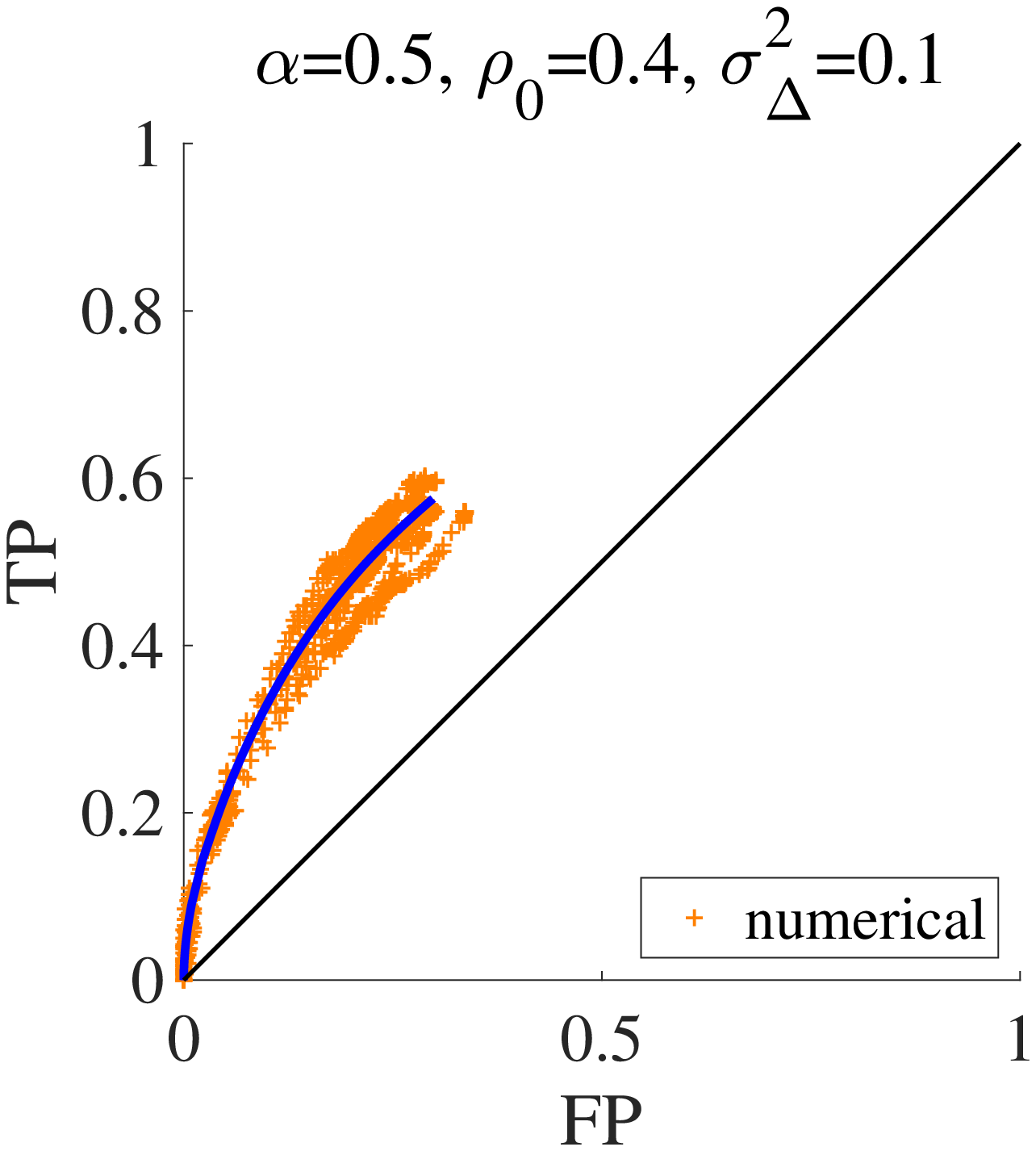}
\caption{
ROC curves evaluated by the analytical (blue thick curve) and numerical (orange cross point) for $(\alpha,\rho_0,\sigma_{\noise}^2)=(0.5,0.2,0.1)$ (left) and $(\alpha,\rho_0,\sigma_{\noise}^2)=(0.5,0.4,0.1)$ (right). The numerical data is obtained by the experiments of 10 different samples at $N=1000$. The agreement between the two are fairly good. 
}
\Lfig{numcheck-ROC}
\end{center}
\end{figure}
The numerical result is displayed as the scatter plots (orange cross points) of $TP$ and $FP$, for the experiments of 10 different samples at $N=1000$. The numerical plots show a fairly good agreement with the analytical curve, which again justifies our analytical solutions.  

\subsubsection{Accuracy of the approximate CV formula}\Lsec{Accuracy of the}
To check the accuracy of the approximate CV formula, in \Rfig{CV-directcomp} we compare the CV errors between the literal (by \Req{LOOE}) and approximate (by \Req{CVformula}) CVs for two specific samples of the system size $N=100$. The other parameters are $(\alpha,\rho_0,\sigma_{\noise}^2,a)=(0.5,0.2,0.1,3)$ and $(\alpha,\rho_0,\sigma_{\noise}^2,a)=(1.5,0.2,0.1,4)$, which correspond to \Rfig{numcheck-eps}. Here, all the results are obtained using the $\lambda$ annealing and they show regular behaviours, even below the AT point.  
\begin{figure}[htbp]
\begin{center}
\includegraphics[width=0.48\columnwidth]{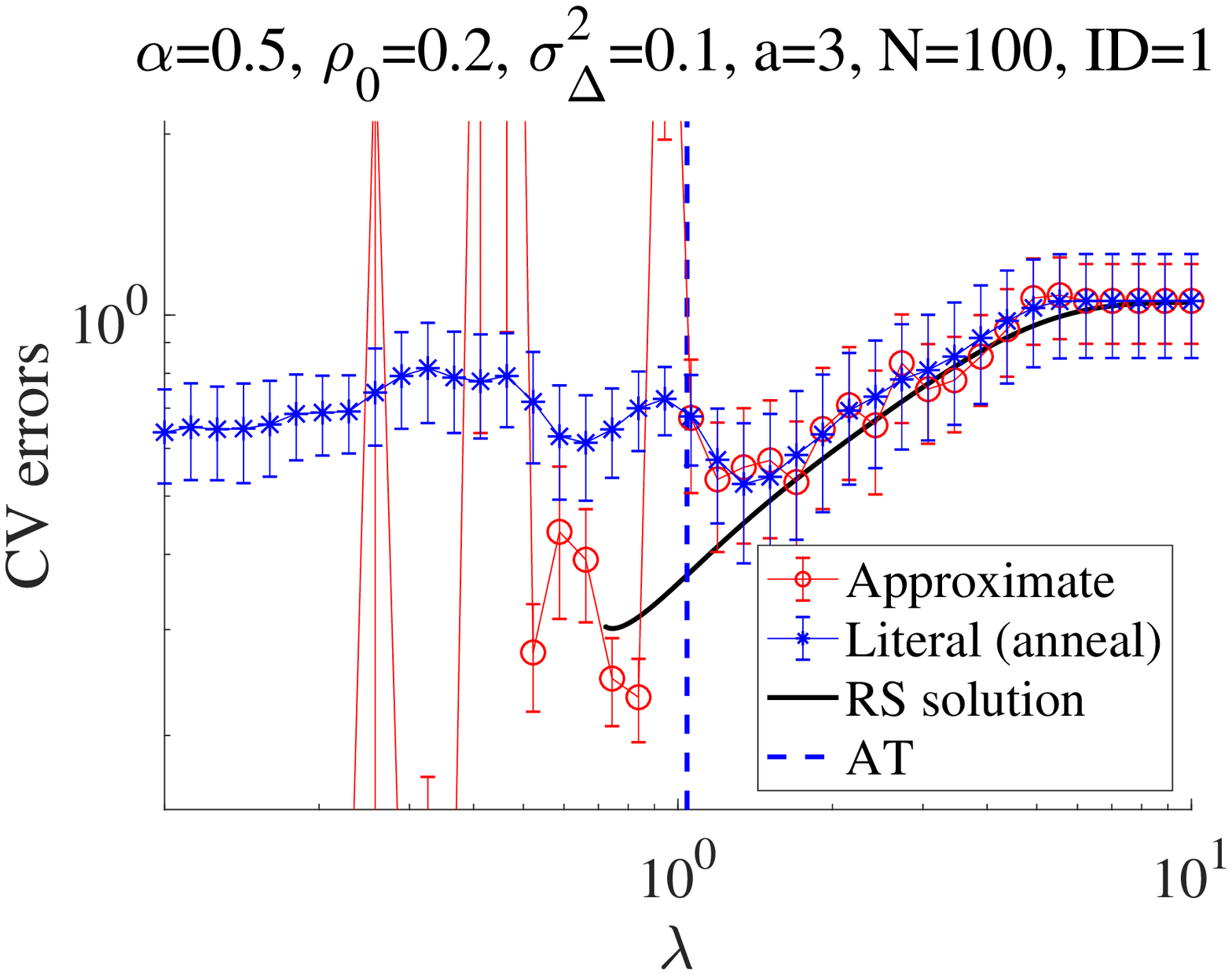}
\includegraphics[width=0.48\columnwidth]{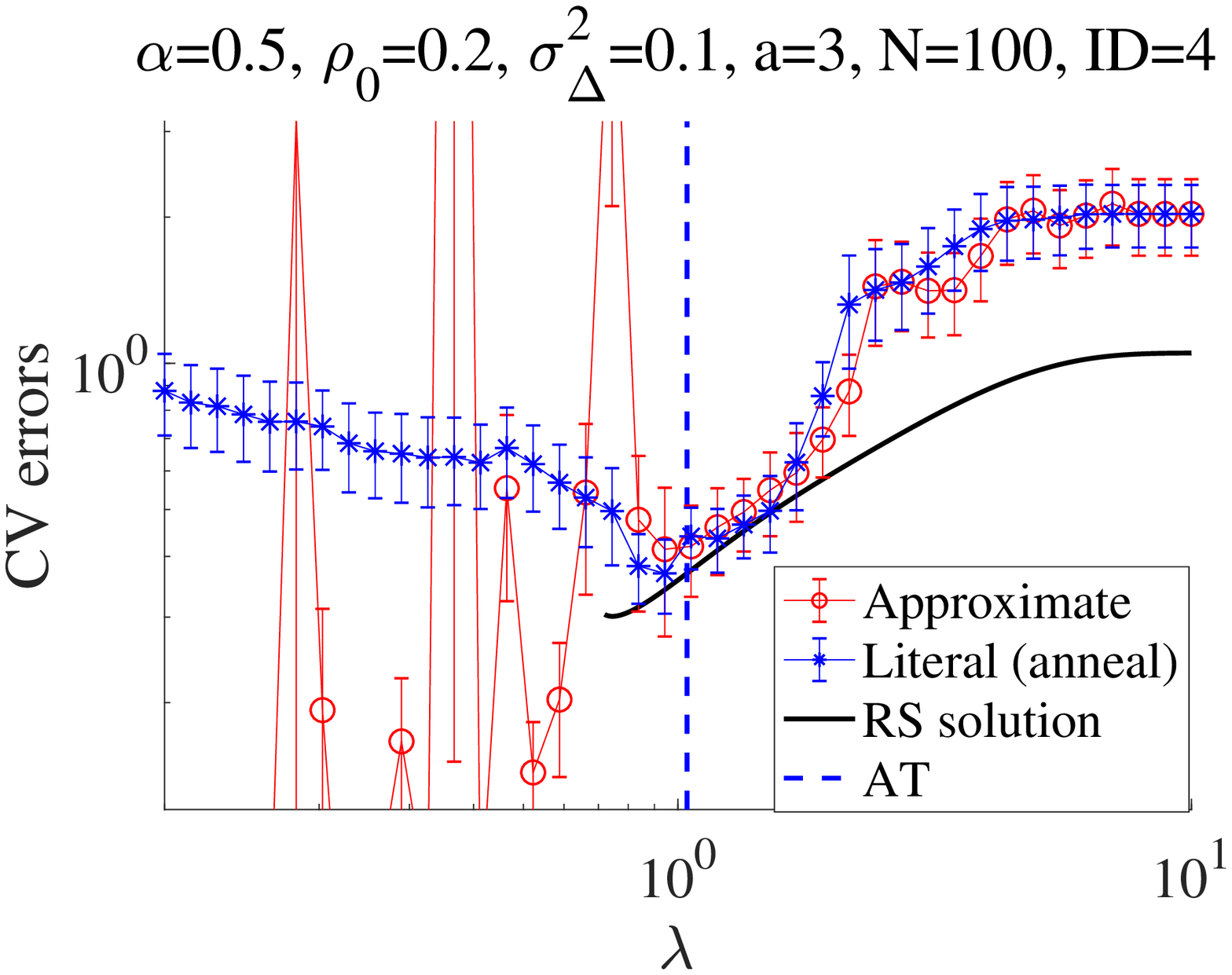}
\includegraphics[width=0.48\columnwidth]{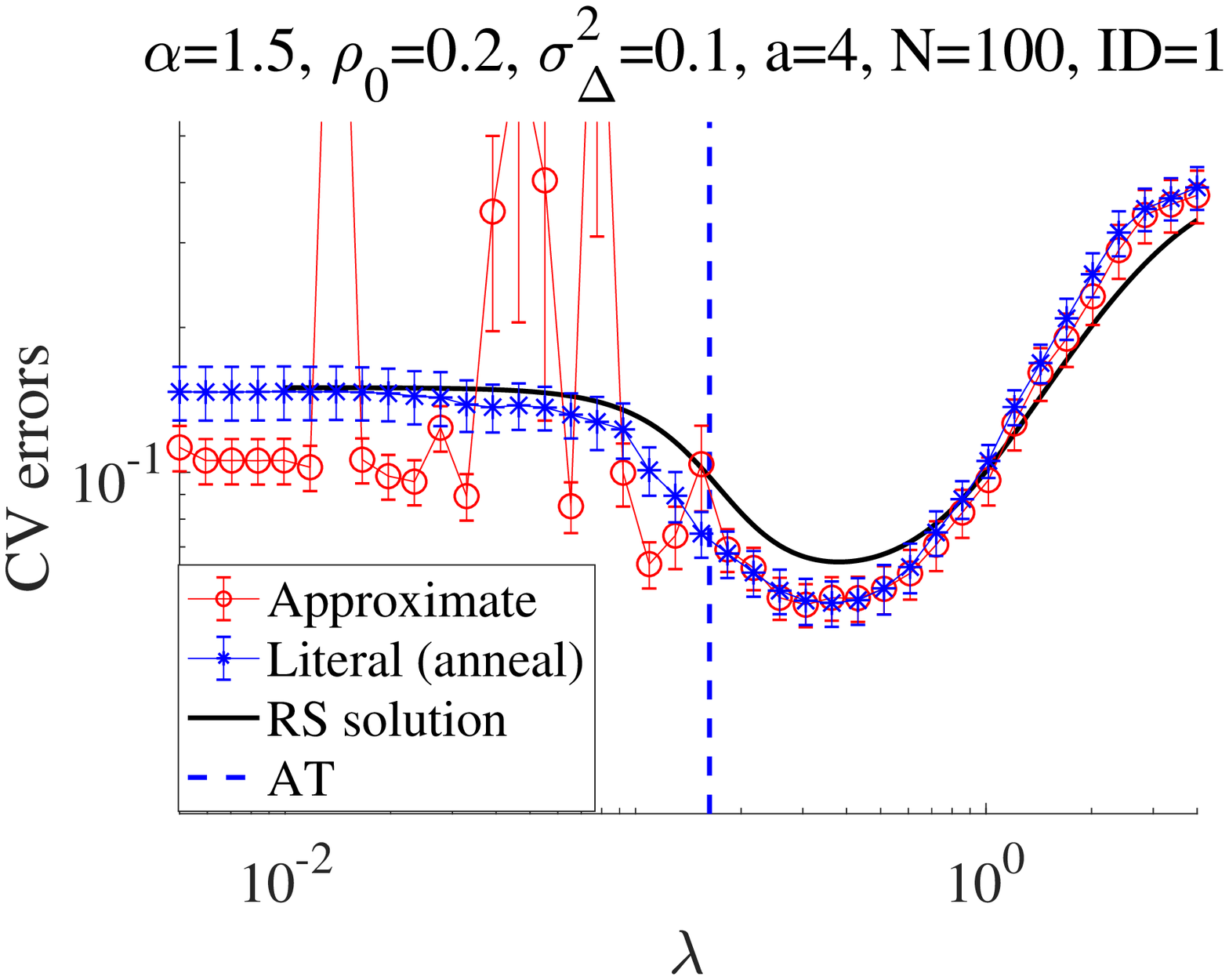}
\includegraphics[width=0.48\columnwidth]{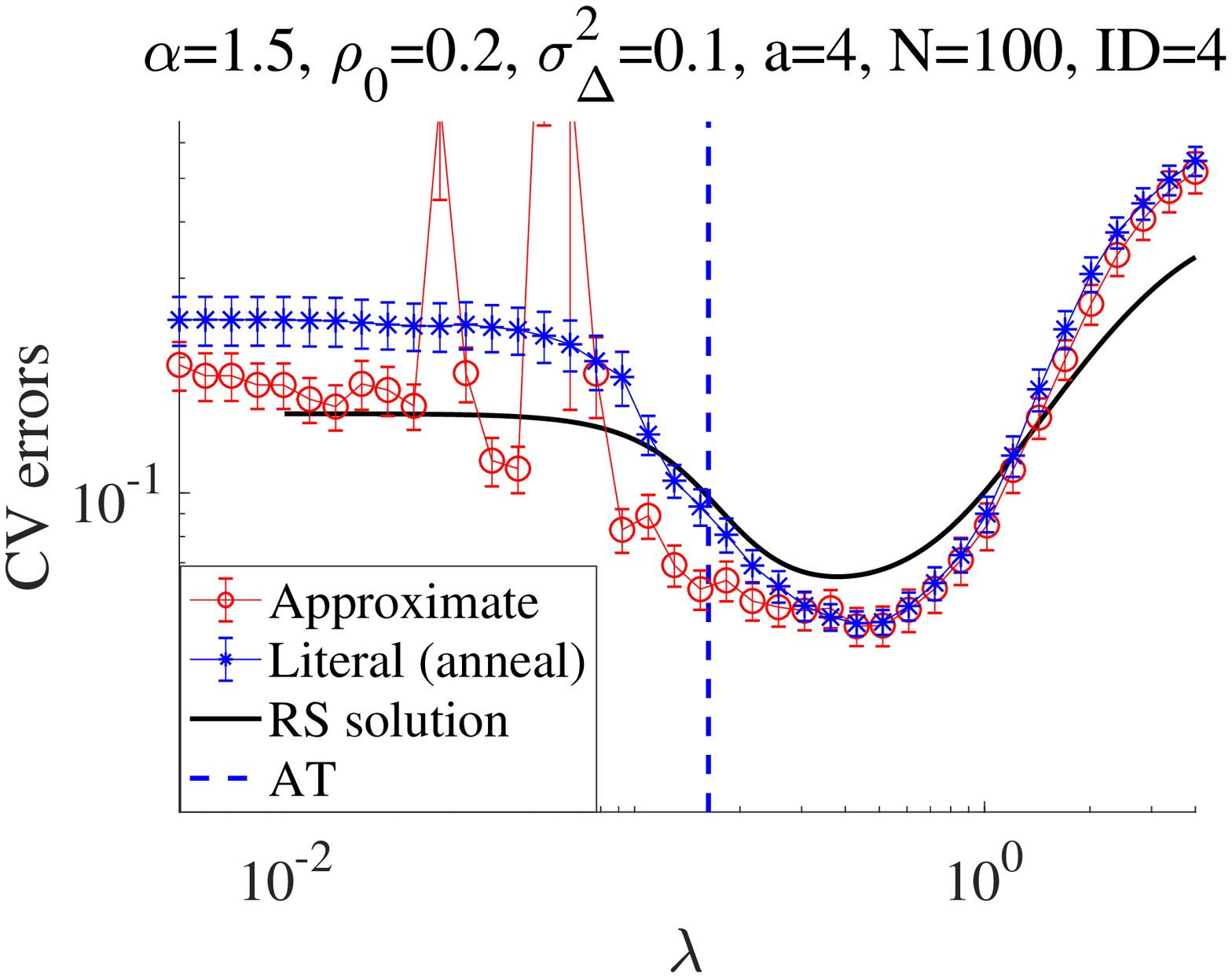}
\caption{
Plots of the literal (by \Req{LOOE}) and approximate (by \Req{CVformula}) CV errors against $\lambda$ at $N=100$ for given $(\alpha,\rho_0,\sigma_{\noise}^2,a)=(0.5,0.2,0.1,3)$ (upper) and $(\alpha,\rho_0,\sigma_{\noise}^2,a)=(1.5,0.2,0.1,4)$ (lower). The results of two specific different samples (left and right) are shown. 
The black thick curve and the vertical blue dashed line represent the RS solution and the AT point, respectively. The approximate results are well matching to the literal ones up to the AT point.
}
\Lfig{CV-directcomp}
\end{center}
\end{figure}
In all the cases, the approximate result reproduces well the literal one up to the AT point, even for the error bars given by the way explained at \Rsec{Derivation}. The uncontrolled behaviour of the approximate formula below the AT point is owing to the singular behaviour in the factor $\lb \lb  A^{\Bs \mu}_{\Wc S_{A}} \rb^{\top} A^{\Bs \mu}_{\Wc S_{A}}+ \lb \HESS  J (\hat{\V{x}}_{S_A};\eta)\rb_{S_{A}S_{A}} \rb^{-1}$ in \Req{d}. This is natural because this factor is nothing but the susceptibility, which is known to involve diverging modes when the AT instability occurs~\cite{mezard1987spin}. These considerations mean that our approximate formula is only applicable above the AT point or in the RS phase. 

Can we detect the instability point only from the approximate CV result without referring to the replica computation? \Rfig{CV-directcomp} speaks for this, because the approximate CV error tends to show uncontrolled behaviours at and below the AT point. As a trial, assuming a combination use with the $\lambda$ annealing, we examine the following procedures to detect the uncontrolled behaviours:
\begin{enumerate}
\item{Detect `irregular' datapoints by locally comparing each datapoint with neighbouring points along the $\lambda$ path (here datapoints mean the approximate CV result, red circles in \Rfig{CV-directcomp}).}
\item{Find the maximum value of $\lambda$ whose corresponding datapoint is irregular. Regard all the $\lambda$ region below it as `instability region'.}
\end{enumerate}
To obtain a concrete result, we need to implement the first step (i) as an algorithm. The actual implementation is:
\begin{description}
\item[(i)-1]{If the CV error difference between the irregular point candidate and the compared datapoint is larger enough in reference to the error bar of the compared point, then the candidate is regarded as ``irregular''.}
\end{description}
By these procedures, we can separate all the parameter regions into two parts: Stable and unstable regions corresponding to the RS and RSB phases, respectively. By employing this, in \Rfig{PD-sample} we draw `phase diagrams' of two specific samples, used also in \Rfig{CV-directcomp}. 
\begin{figure}[htbp]
\begin{center}
\includegraphics[width=0.48\columnwidth]{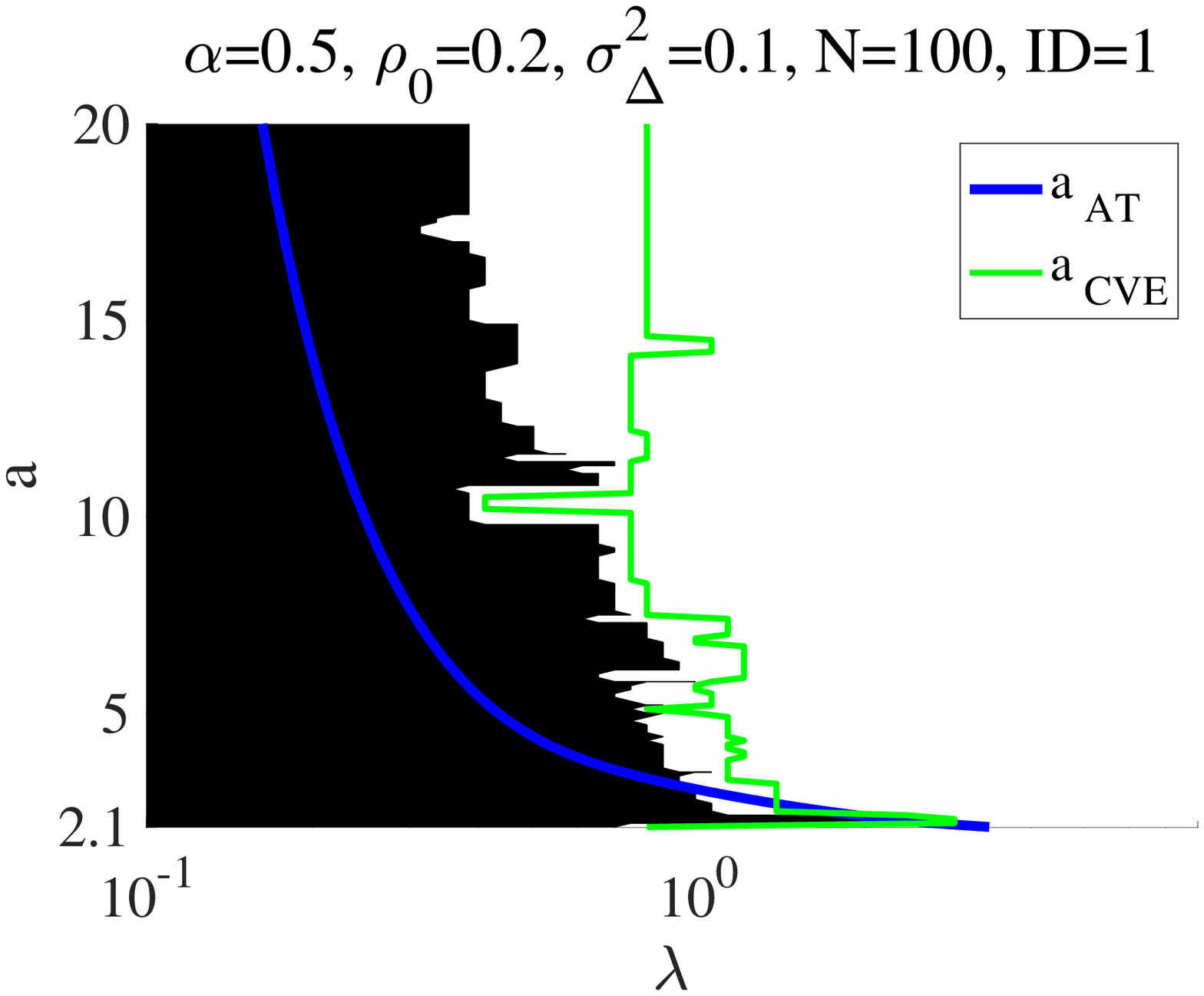}
\includegraphics[width=0.48\columnwidth]{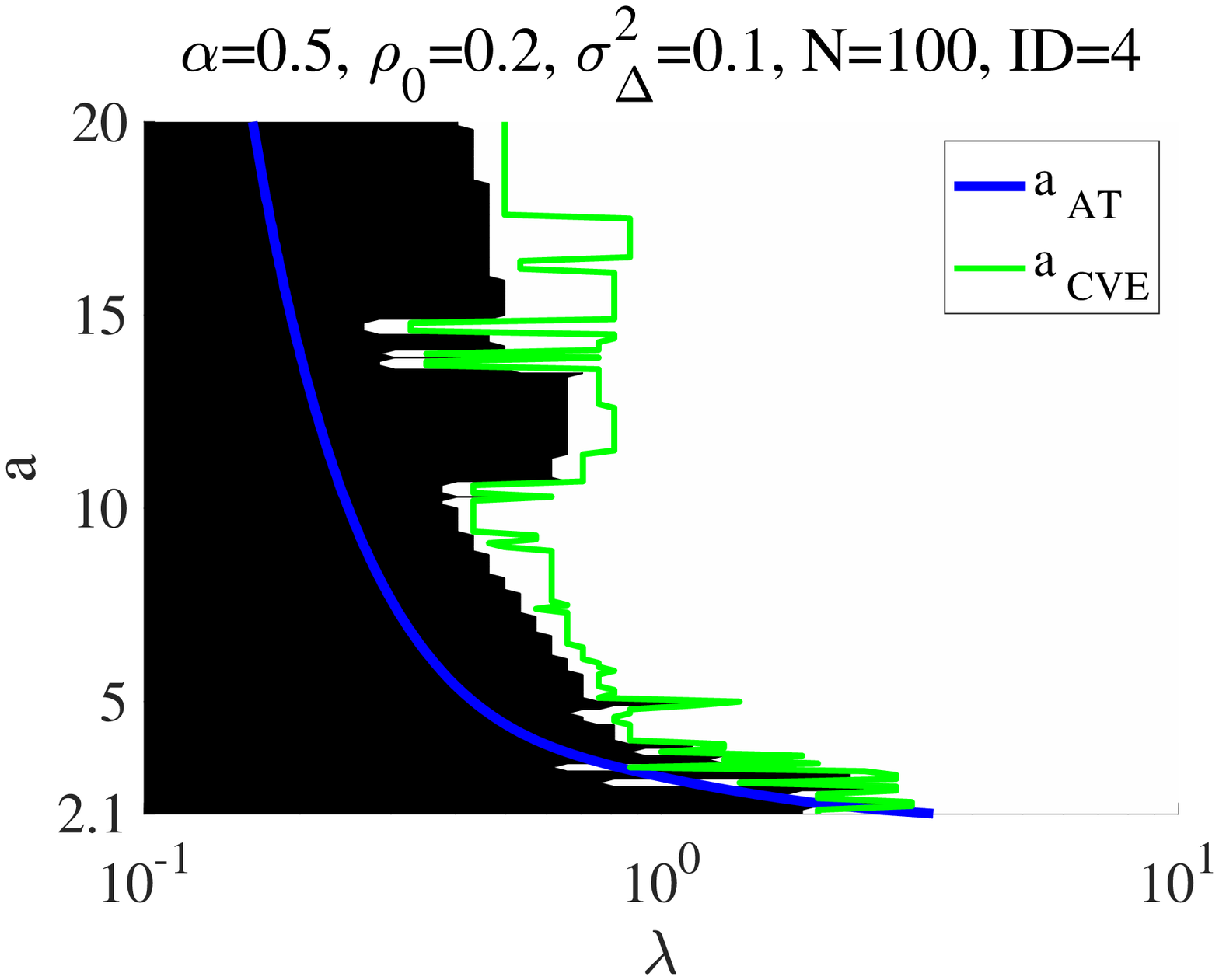}
\caption{
`Phase diagrams' drawn for two specific samples used in \Rfig{CV-directcomp} using the instability detection procedures described in the main text. The white and black regions represent the stable and unstable regions which correspond to the RS and RSB phases. The blue thick curve denotes the AT line $a_{\rm AT}$ computed by the replica method, while the green curve $a_{\rm CVE}$ shows the $\lambda$ location of the approximate CV error minimum given $a$ in the stable region which is supposedly related to $a_{\rm IMSE}$. 
}
\Lfig{PD-sample}
\end{center}
\end{figure}
This figure shows that the boundary between the white and black regions behaves similarly to the AT line $a_{\rm AT}(\lambda)$, although there is a gap supposedly owing to the sample fluctuation and the finite-size effect. As a reference, we also compute the minimum location of the approximate CV error with respect to $\lambda$ given $a$, defining $a_{\rm CVE}(\lambda)$, which is denoted by the green curve in \Rfig{PD-sample}. According to \Req{GE-LOOE}, this corresponds to $a_{\rm IMSE}$ appearing in the phase diagrams in \Rsec{Phase diagram}. Note that these procedures are somewhat overcautious, and can miss some stable regions possibly existing in the small $\lambda$ region. As typically seen in the left panel of \Rfig{PD-noise}, the re-entrant transition can emerge in the weak noise case but the present procedures cannot detect this re-entrancy, because these procedures detect the first RS-RSB transition corresponding to the rightmost branch of $a_{\rm AT}$ and all the region below this first transition point is regarded as `instability region'.  However, for practical use, it is more important to avoid giving wrong estimates by our approximate formula. Hence, we do not aim to improve the above instability detection procedures in this study. 

Apart from the re-entrancy, it is worthwhile to investigate the cause of the gap between the AT line and the instability points detected by our procedures. To this end, we compute the boundary value between the black and white regions in \Rfig{PD-sample} given $a$, $\lambda_c(a)$, for many samples and different system sizes. The results at $a=5$ and $a=10$ are shown in \Rfig{asym_lambda_c}.
\begin{figure}[htbp]
\begin{center}
\includegraphics[width=0.48\columnwidth]{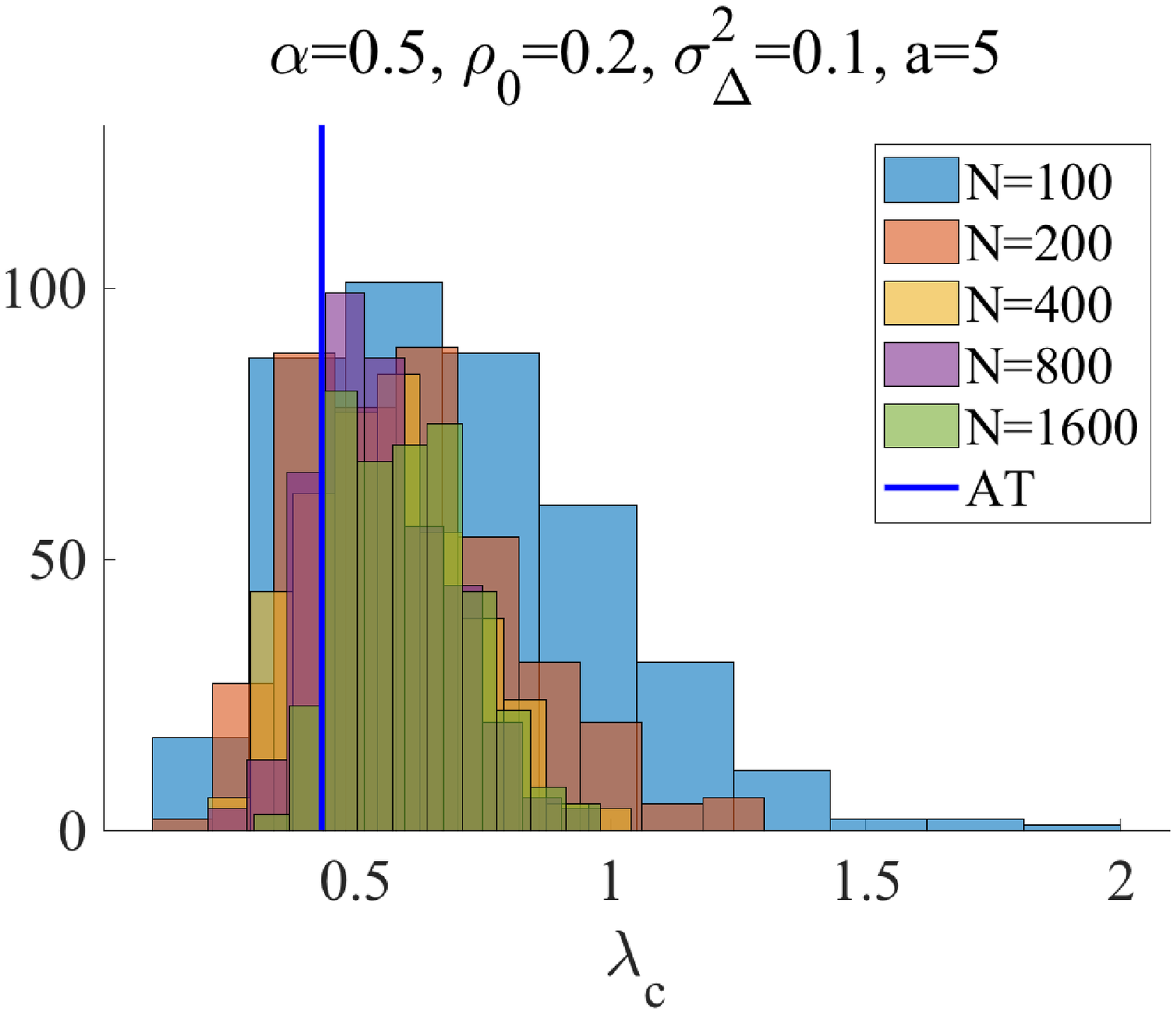}
\includegraphics[width=0.48\columnwidth]{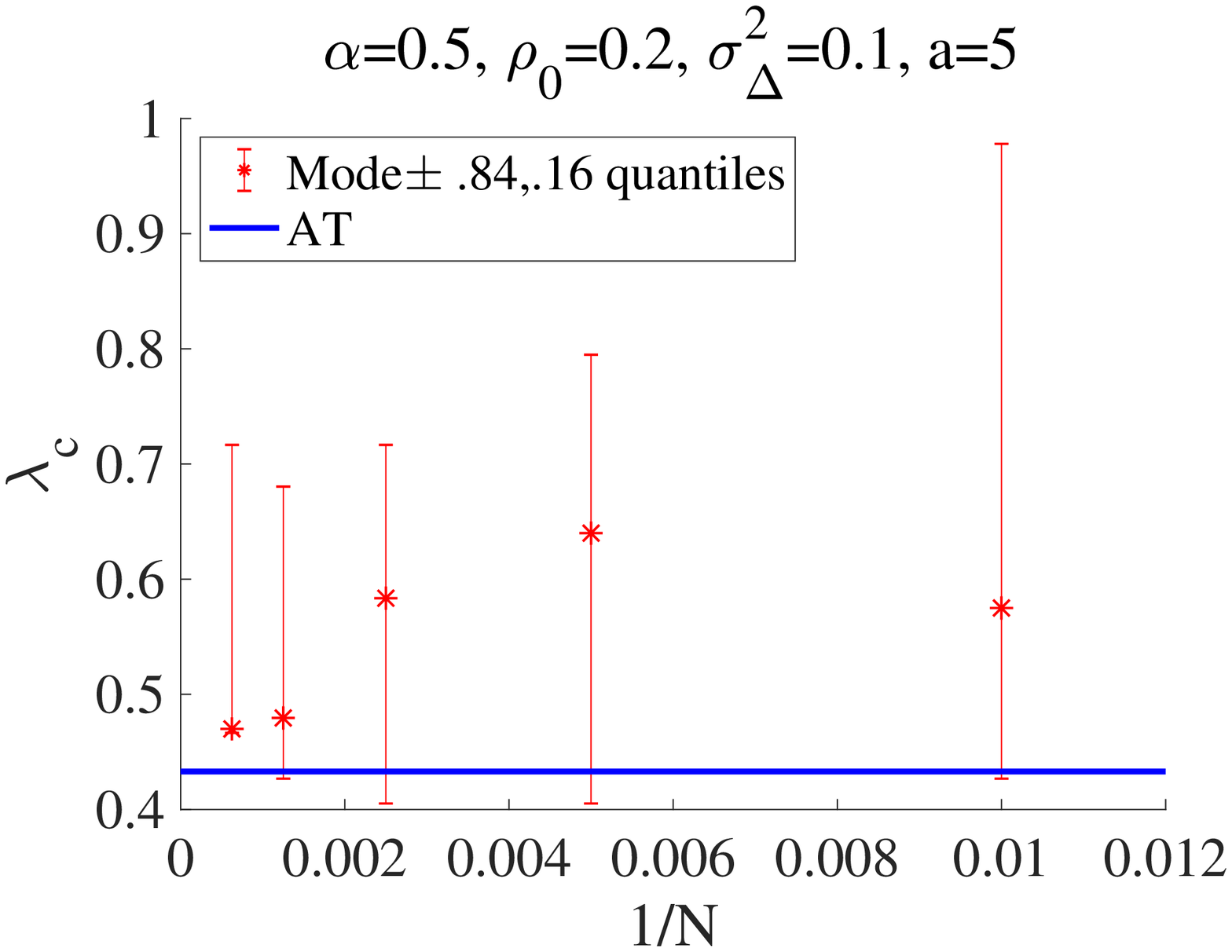}
\includegraphics[width=0.48\columnwidth]{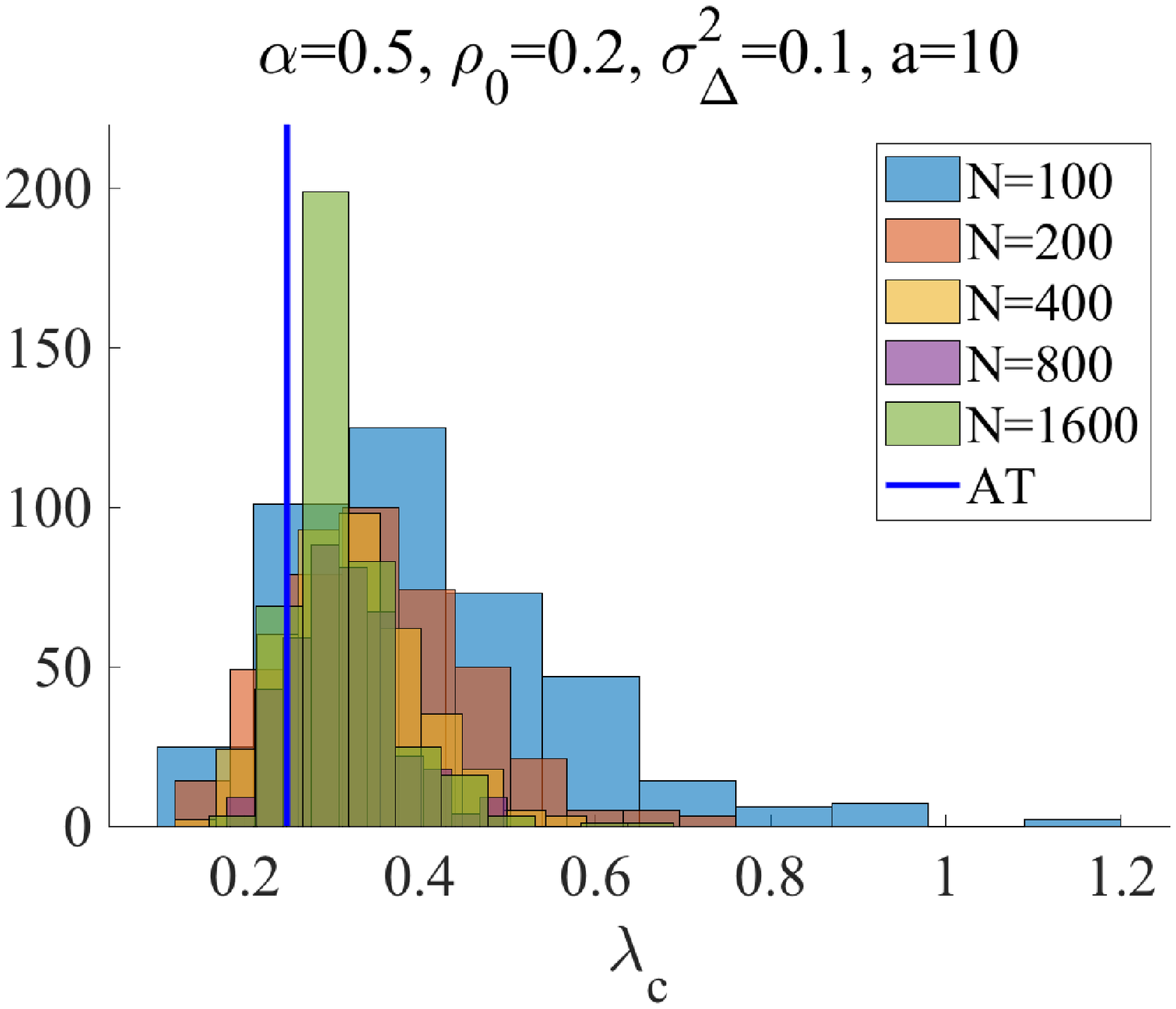}
\includegraphics[width=0.48\columnwidth]{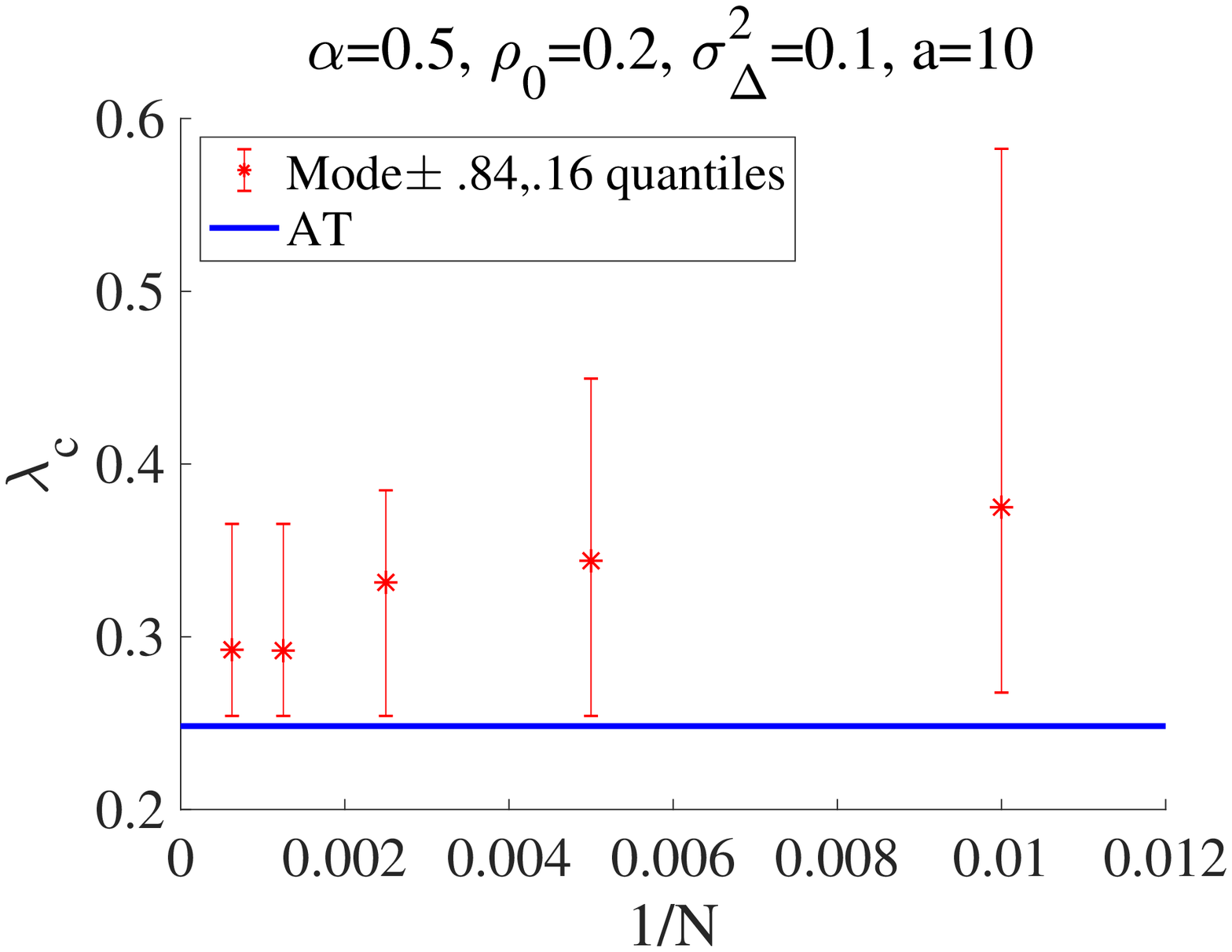}
\caption{
Histograms of $\lambda_c$ (left) and plots of the mode values against the inverse system size (right) at $a=5$ (upper) and $a=10$ (lower) computed from $N_{\rm samp}=400$ samples. The blue straight line represents the $\lambda$ value at the AT point commonly in all the panels. The error bar of the mode is put as the $0.86$ and $0.14$ quantiles of the histogram. The examined sizes are $N=100,200,400,800,1600$ and the different colors of the histograms correspond to different sizes as shown in the legend. The other parameters are set to $(\alpha,\rho_0,\sigma_{\noise}^2)=(0.5,0.2,0.1)$. To unambiguously define the mode value, we set the number of bins $N_{\rm bin}$ by Sturges rule as $N_{\rm bin}=\lceil 1+\log_2 N_{\rm samp} \rceil$. As the system size grows, the width of the histogram shrinks and the mode value approaches the AT point.
}
\Lfig{asym_lambda_c}
\end{center}
\end{figure}
The left panels provide the histograms of $\lambda_c$ from $N_{\rm samp}=100$ samples for different sizes $N=100,200,400,800,1600$ discriminated by different colors. As the system size grows, the width of the histogram shrinks and the mode value tends to approach the $\lambda$ value at the AT point. Here the number of bins $N_{\rm bin}$ for the histogram is determined by the so-called Sturges rule~\cite{sturges1926choice} as $N_{\rm bin}=\lceil 1+\log_2 N_{\rm samp} \rceil$, enabling us to define the mode value without ambiguity. To quantify the convergence behavior of the mode value, we plot the mode against the inverse system size $1/N$ in the right panels. Here the mode value shows a clear tendency of approaching to the AT value as $N$ grows. This indicates that the gap is actually due to the sample fluctuation and the finite-size effect, and also implies that our instability detection procedures are reasonably connected to the AT instability.  

The AT instability is known to be connected to the emergence of many local minimums~\cite{mezard1987spin}. To directly check this, we conduct the literal CV without the $\lambda$ annealing. For each point of $\lambda$, the estimator is computed from ten different randomly initialized $\V{x}$, each component of which is i.i.d. from $\mc{N}(0,1)$, by the CD algorithm. In \Rfig{CV-random}, the resultant CV errors are given as scatter plots in combination with the CV error using the annealing. The experimental setup of each panel is again identical to the corresponding one in \Rfig{CV-directcomp}. 
\begin{figure}[htbp]
\begin{center}
\includegraphics[width=0.48\columnwidth]{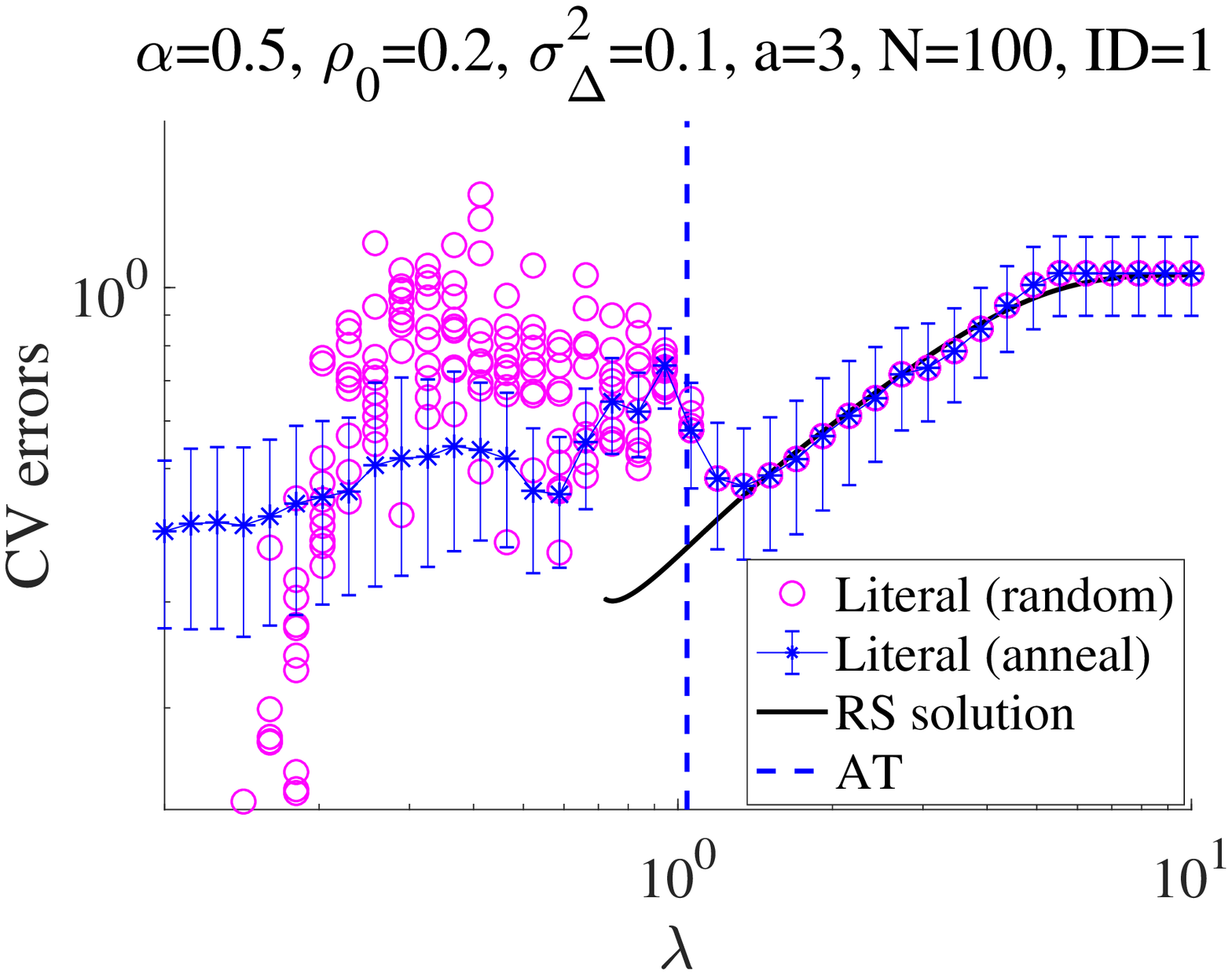}
\includegraphics[width=0.48\columnwidth]{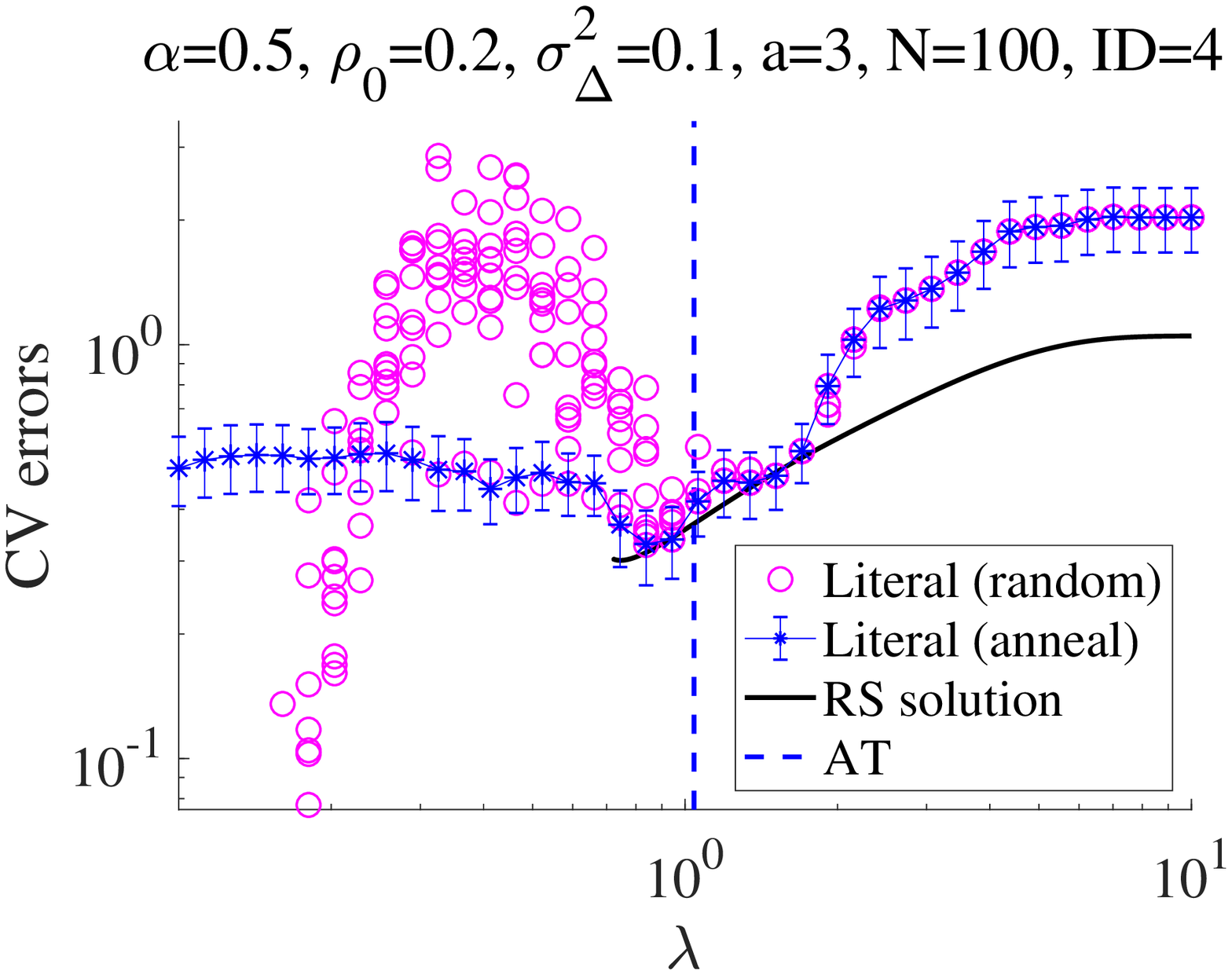}
\includegraphics[width=0.48\columnwidth]{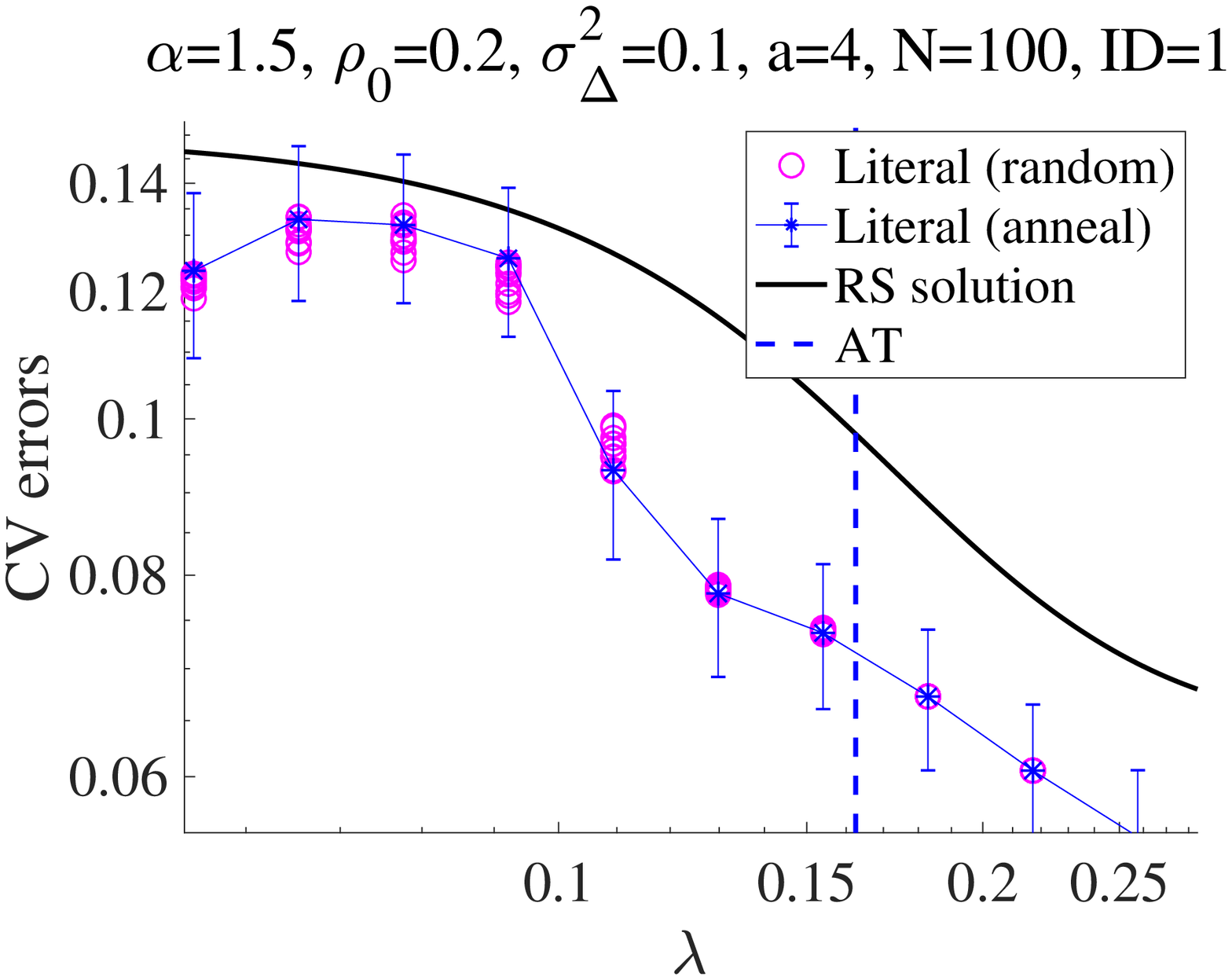}
\includegraphics[width=0.48\columnwidth]{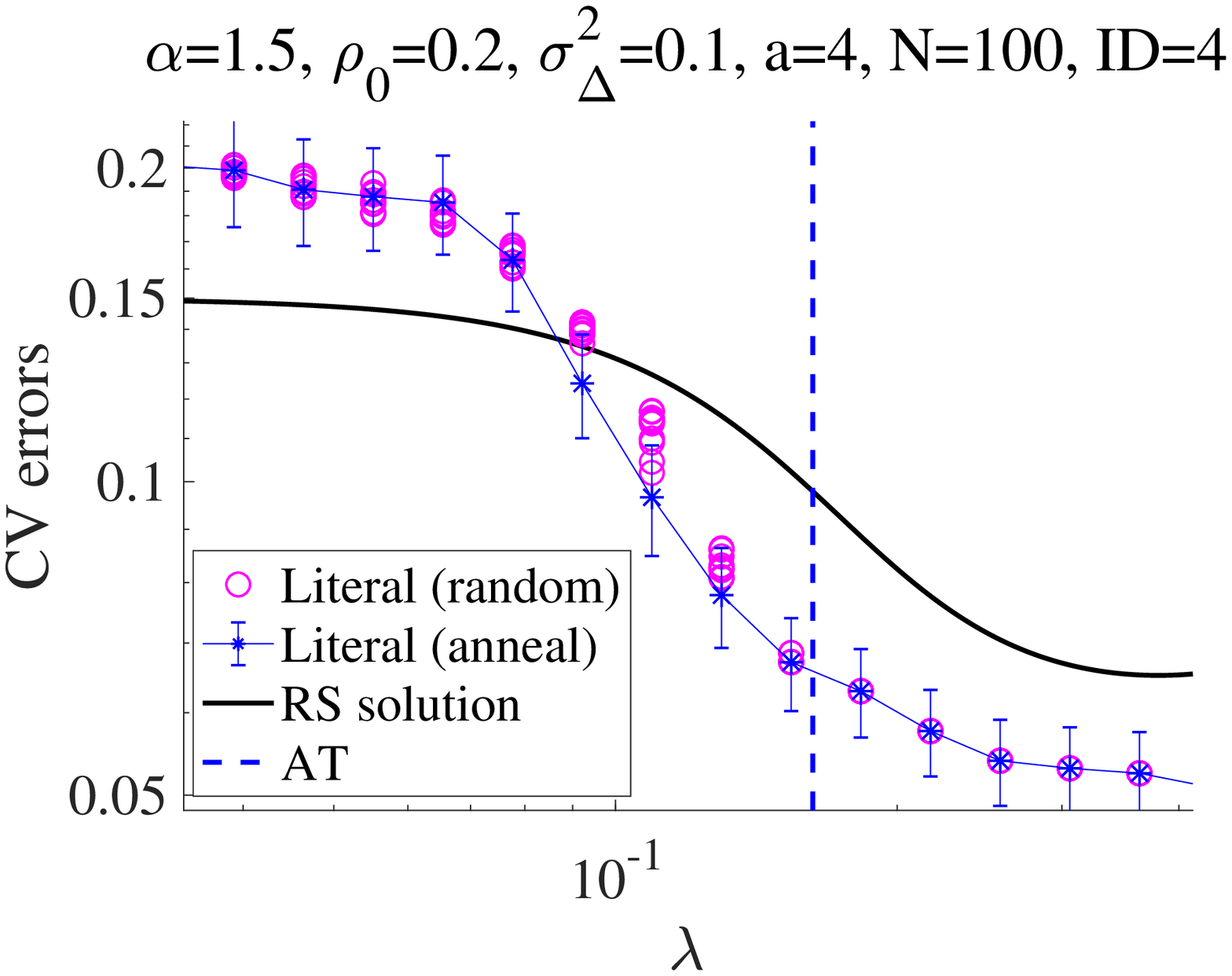}
\caption{
Comparison of the literal CV errors with and without the $\lambda$ annealing in the same experimental condition as the corresponding panel of \Rfig{CV-directcomp}. The result without the annealing is shown as scatter plots (magenta circles) for ten different random initial conditions and it exhibits visible differences from the annealing result (blue asterisks, identical result to \Rfig{CV-directcomp}) below the AT point, while no difference exists sufficiently above the AT point. For the lower panels, the region around the AT point is magnified because the difference is small, although it exists. 
}
\Lfig{CV-random}
\end{center}
\end{figure}
This figure gives a clear evidence of the multiple solutions below the AT point. \Rfig{CV-random} also implies that the solution obtained by the $\lambda$ annealing is rather atypical: Solutions obtained from random initial conditions tend to give rather different values of CV error from the annealed solution. To give a better theoretical background to this statement, we have to construct the full-step RSB solution and to figure out the characterisation of the annealed solution in the ensemble of all the local minimums. This is beyond the present purpose but will be an interesting future work.

Our present attitude to the multiplicity of the solutions is to avoid it. This is reasonable because the global minimum of the generalisation error is in the RS region without the multiplicity, as clarified by our analytical computation. Once accepting this attitude, we can use the proposed approximate formula efficiently estimating the generalisation error and, fortunately, the formula also enables us to avoid the multiple solution region by using the above instability detection procedures. This is the main outcome and contribution of this study.

Before closing this subsection, we check the computational time and the approximate accuracy of the proposed formula more quantitatively. Here we quantify the error difference between the literal and approximate CVs by a {\it normalised MSE} defined as:
\be
{\rm normalised~MSE}=
\lb \frac{
 \epsilon_{\rm CV,approx.}-\epsilon_{\rm CV,literal}
 }{
 \epsilon_{\rm CV,literal}
 }
 \rb^2,
 \ee
where $\epsilon_{\rm CV,approx.}$ and $\epsilon_{\rm CV,literal}$ are the CV errors evaluated by the approximate and literal CV procedures, respectively. According to the derivation of the formula in \Rsec{Approximate}, the accuracy is considered to be better as $N$ and $M$ increase. Thus, we plot the normalised MSE against $N$ as the left panel of \Rfig{comptime}. The parameters are set to $(\alpha,\rho_0,\sigma_{\noise}^2,a,\lambda)=(0.5,0.2,0.1,4,1)$ as an example. For each $N$, we compute the normalised MSE for several different samples of $\{ \DM,\V{x}_0,\V{\noise} \}$, and the marker (blue asterisk) denotes the median among the samples, and the upper and lower error bars correspond to the $0.86$ and $0.14$ quantiles, respectively. The number of samples is $\{1000,1000,200,200,100,50,10,10,10,2\}$ for the system size $\{50,100,200,400,800,1600,3200,6400,10000,20000\}$, respectively.  
\begin{figure}[htbp]
\begin{center}
\includegraphics[width=0.32\columnwidth]{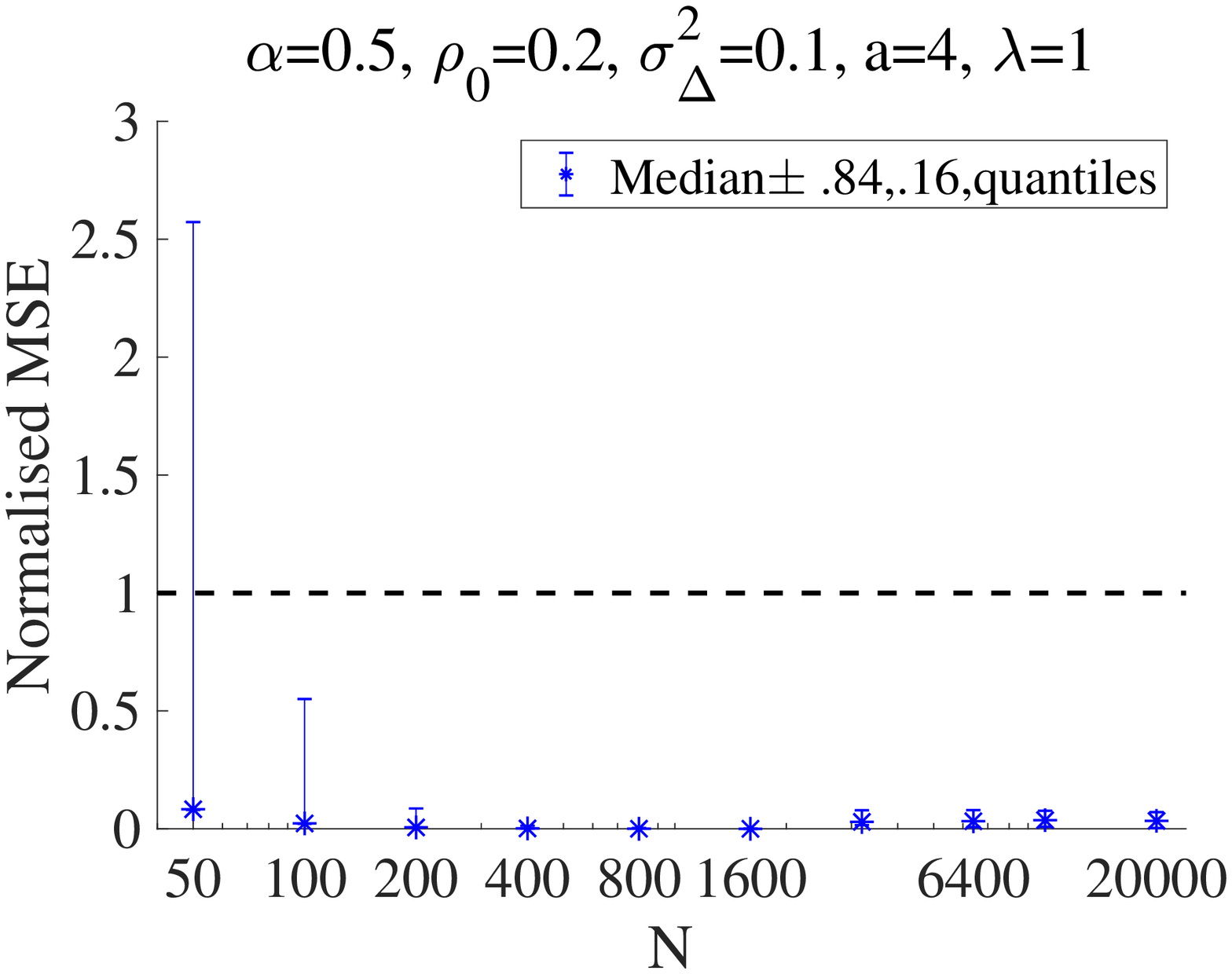}
\includegraphics[width=0.32\columnwidth]{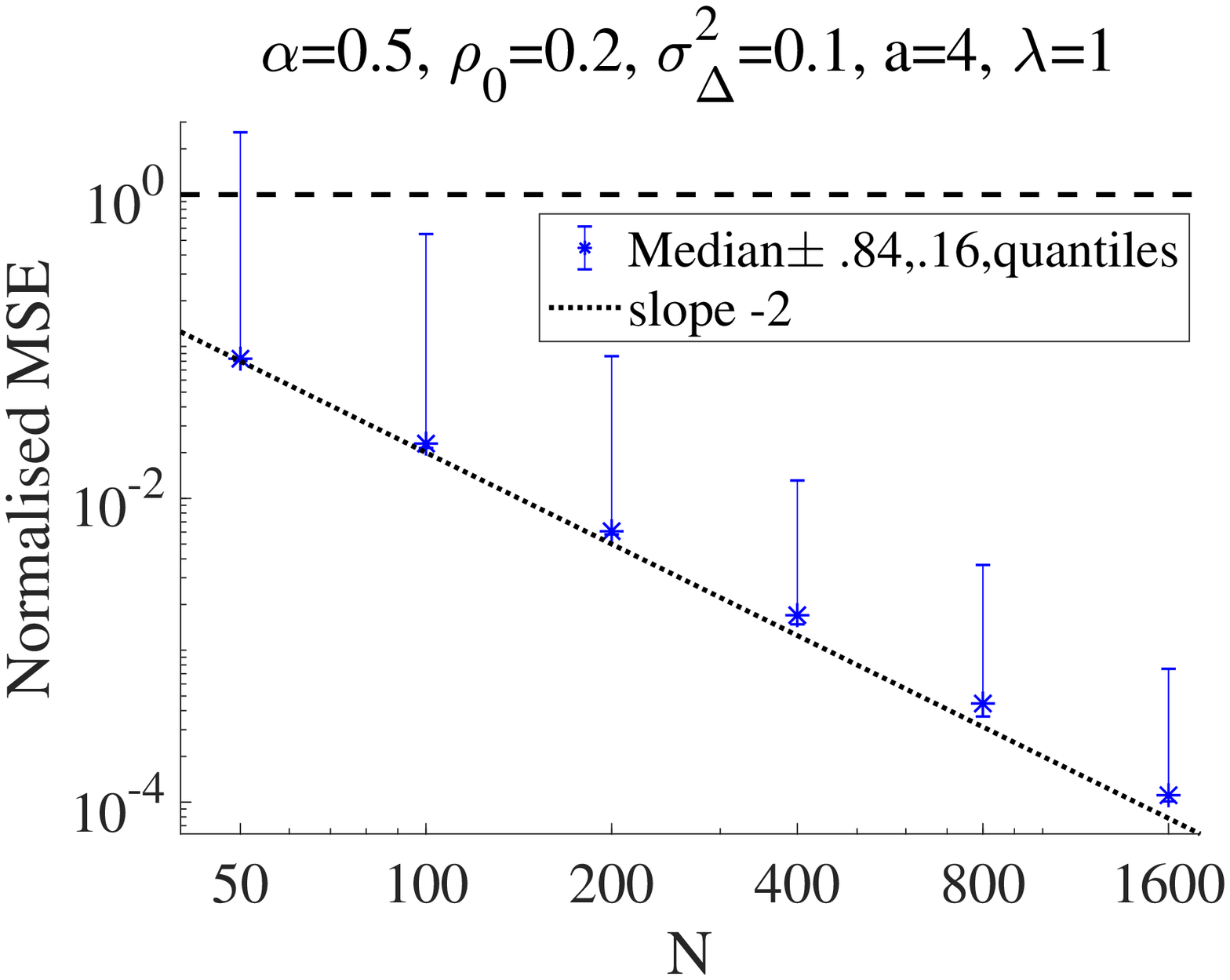}
\includegraphics[width=0.32\columnwidth]{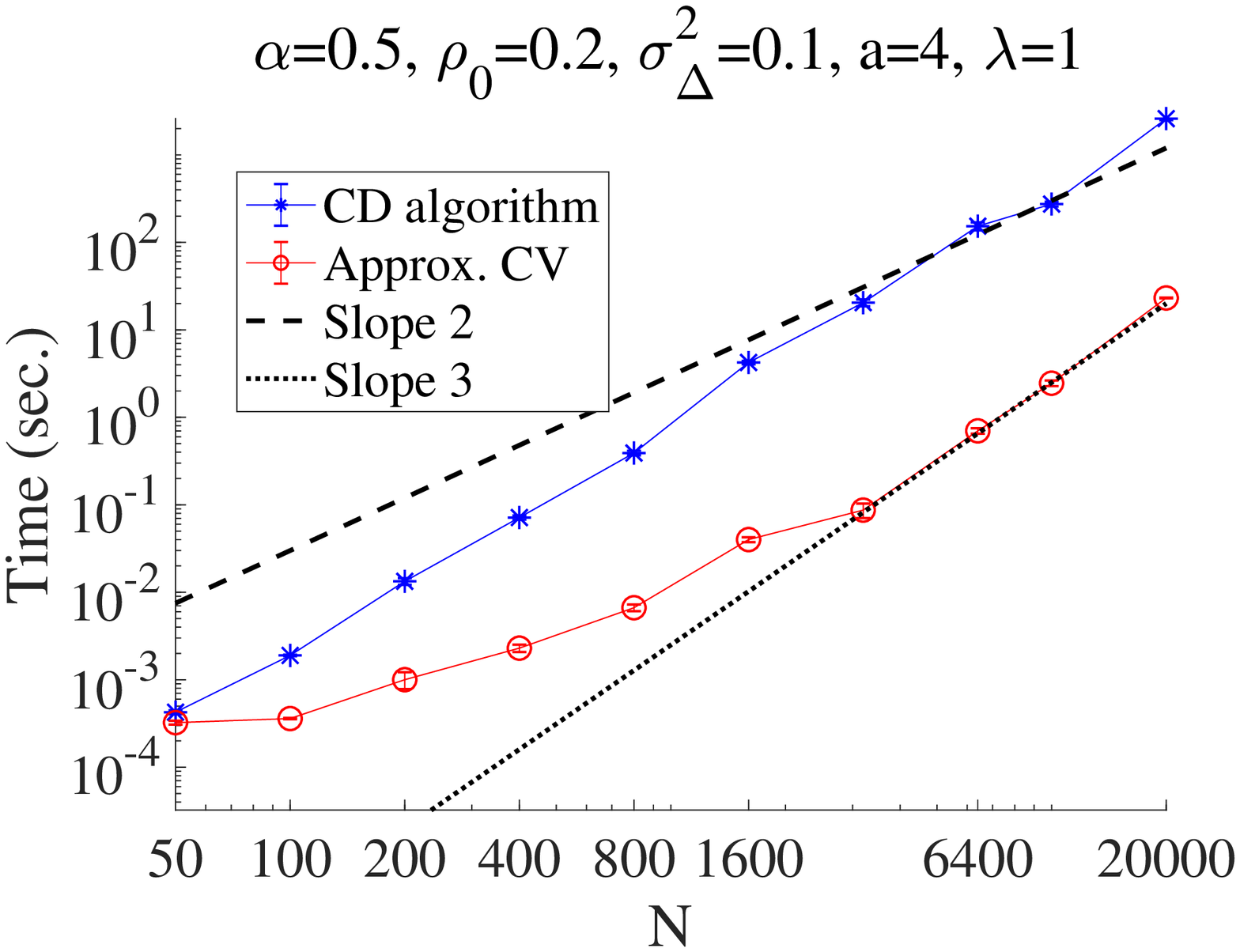}
\caption{
(Left) The normalised MSE of the CV error difference is plotted against the system size $N$ in the log-linear scale. The number of samples is $\{1000,1000,200,200,100,50,10,10,10,2\}$ for the system size $\{50,100,200,400,800,1600,3200,6400,10000,20000\}$, respectively. The marker denotes the median and the error bars consist of the $0.86$ and $0.14$ quantiles among the samples. The dashed horizontal line denotes unity, given as a reference. For $N\geq 3200$, the literal CV is conducted by the ten-fold CV instead of the LOO CV, to save the computational cost. (Middle) The same plot as the left panel in the double log scale for small sizes $N\leq 1600$. The normalised MSE decreases in the scale $N^{-2}$ as the system size grows, which is clearly indicated by the dotted line representing slope $-2$.  
(Right) The computational time for the CD algorithm convergence (blue asterisk) and for the approximate CV formula (red circle), in the same experiment as the left panel. The error bars are smaller than the marker sizes and hard to see. The dashed line denotes the slope $2$ while the dotted one represents the slope $3$, both of which are the expected size scaling of the computational time of the CD algorithm and the approximate CV formula, respectively.}
\Lfig{comptime}
\end{center}
\end{figure}
As expected, the normalised MSE quite small for large sizes of $N\geq 200$, although at the smallest size $N=50$ there is a non-negligible difference. This difference is dominated by a few percent of samples giving accidentally large values of $\epsilon_{\rm CV,approx.}$. The probability of the accidents seems to become smaller rapidly as the system size grows. 
In the middle panel, the same plot in the double log scale for small sizes $N\leq 1600$ is given, showing a clear decay of the normalized MSE in the scale $N^{-2}$ as the system size grows. This is naturally understood from the scaling argument presented in sec. 5.1.1 in \cite{obuchi2016cross}. 
The corresponding computational time of the CD algorithm convergence and the approximate formula are given in the right panel. The approximate formula requires to take the inverse of the Hessian, leading to the computational cost of $O(|S_{A}|^3)$, which is scaled as the third order polynomial of $N$ if $|S_{A}|=O(N)$. This computational cost can be more expensive than the optimisation cost by the CD algorithm in the large $N$ limit, because the total computational time of the CD algorithm is considered to be scaled as $O(N^2)$, under the assumption such that the convergence of the CD algorithm takes place in constant computational steps independent of the system size $N$, although the $O(N^2)$ behaviour is hard to see in \Rfig{comptime}. Despite this inconvenience in the limiting case, \Rfig{comptime} shows that the computational time of the approximate formula is much smaller than the CD algorithm convergence, in all the investigated range of the system size. We note that the computational time of the CD algorithm shown in \Rfig{comptime} is just for one-time optimisation, and hence, for conducting the literal $k$-fold CV, the required computational time becomes approximately $k$ times larger than that. Overall, although there is no superiority in the large $N$ limit, our approximate formula practically works very efficiently in a wide range of system sizes. 

\subsection{A real-world dataset: Type Ia supernovae}\Lsec{A real-world dataset:}
Here we apply the proposed approximate method to a dataset of Type Ia supernovae. Our dataset is a part of the data from \cite{silverman2012berkeley,Berkeley}, which is screened by a certain criterion~\cite{uemura2015variable}. This dataset was treated by a number of sparse estimation techniques recently, and a set of important variables, which is known to be empirically significant, has been reproduced~\cite{obuchi2016cross,uemura2015variable,kabashima2016approximate,obuchi2016sampling}. In those studies, the LASSO and $\ell_0$ cases are treated, and the CV is employed for hyperparameter estimation. We reanalyse this dataset by using the SCAD penalty and compute the CV error by using the approximate formula. The parameters of the screened data are $M=78$ and $N=276$, and an appropriate standardisation is employed as pre-processing.

Again, we use the $\lambda$ annealing to obtain the SCAD estimators for this dataset, and the CV error is computed by our approximate formula. The instability region is detected by the procedures explained in \Rsec{Accuracy of the}. As \Rfig{PD-sample}, the instability detection gives a phase diagram which is in \Rfig{PD-supernovae}.
\begin{figure}[htbp]
\begin{center}
\includegraphics[width=0.48\columnwidth]{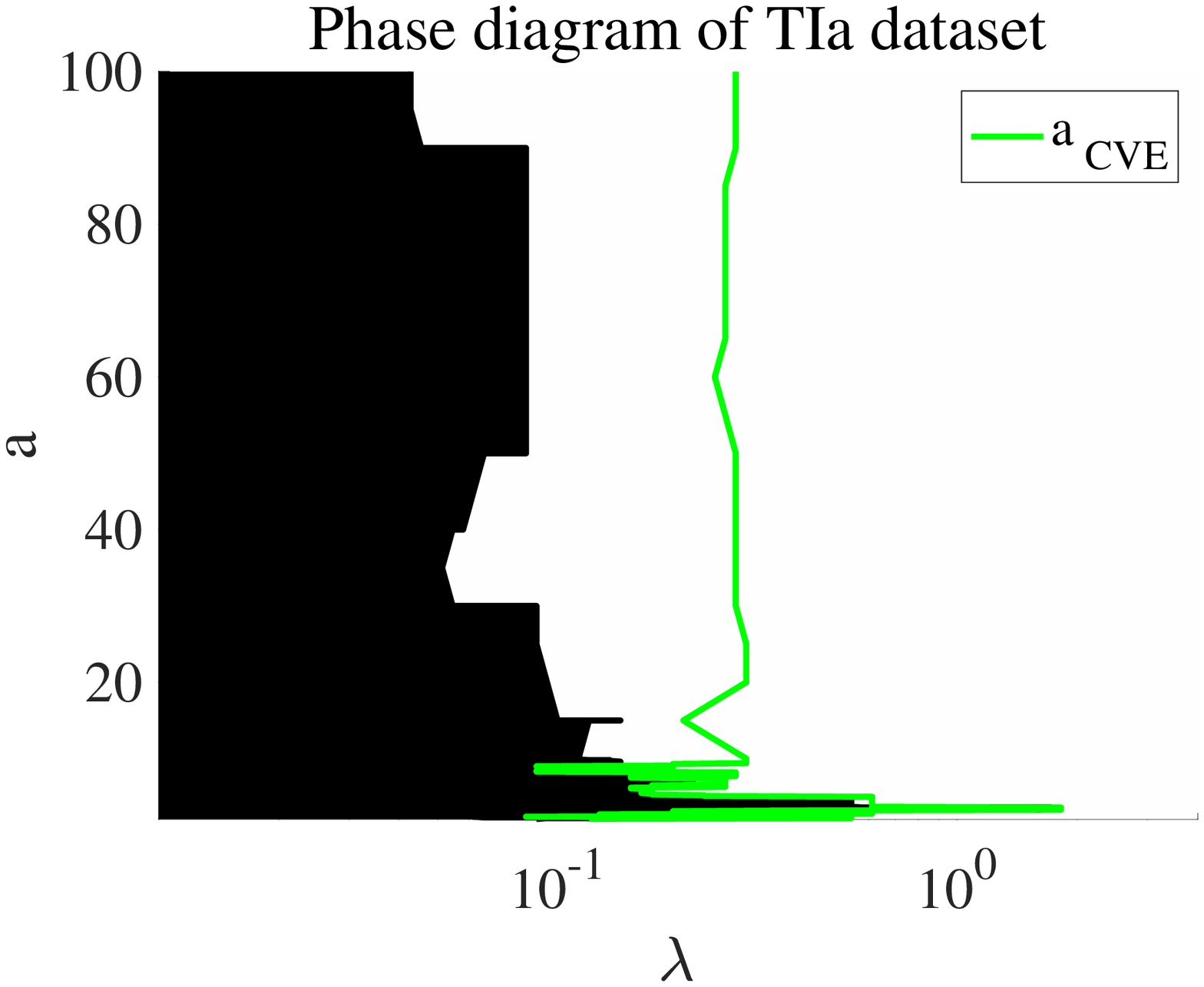}
\includegraphics[width=0.48\columnwidth]{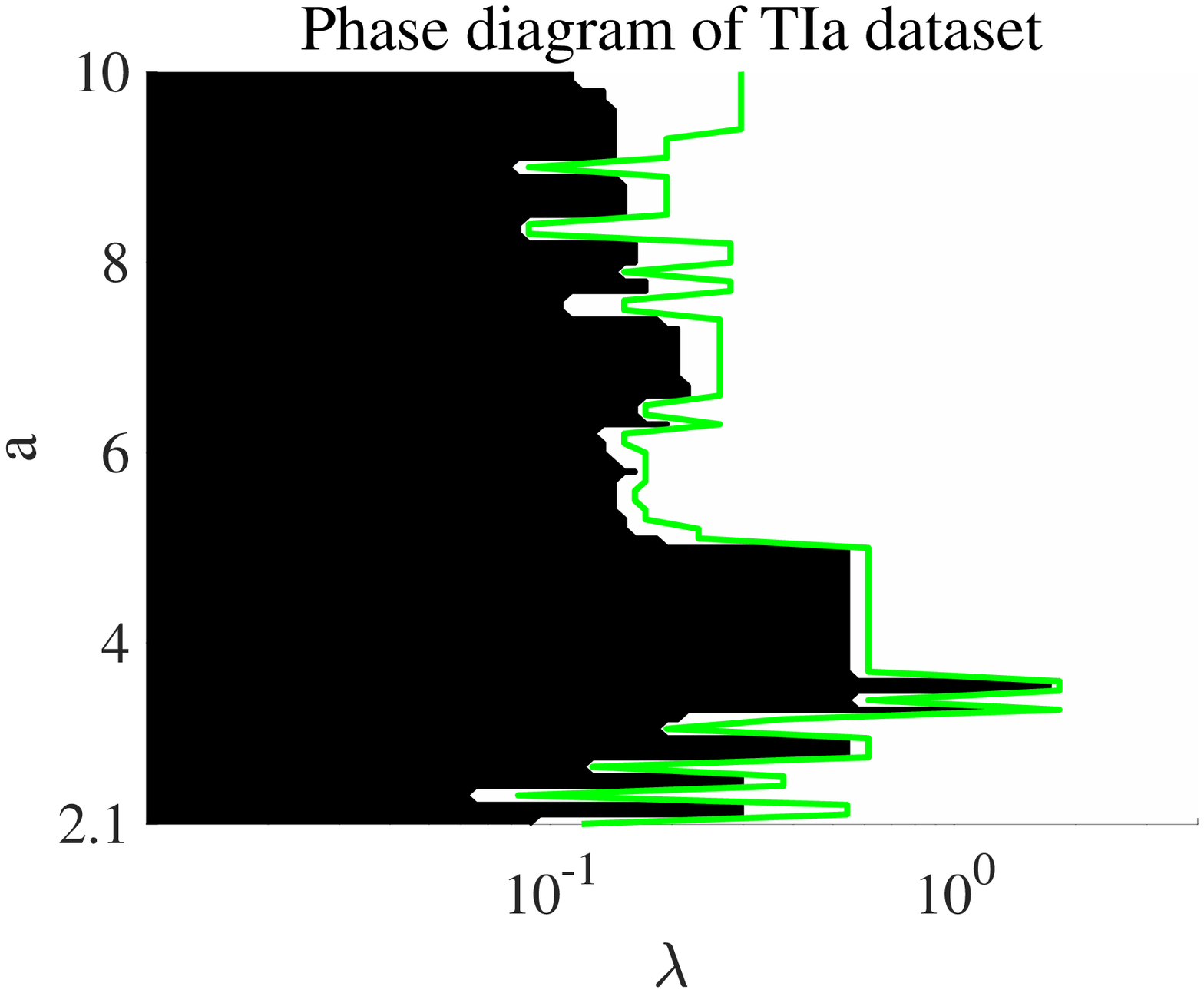}
\caption{
$\lambda$--$a$ phase diagram of the Type Ia supernovae dataset from~\cite{uemura2015variable}. The right panel is the magnified view of the left one in the small $a$ region. The black region represents the instability region for which the approximate formula cannot be applied, while the white one is the stable region in which the approximate formula gives a reliable estimate. The minimum point of the CV error in the stable region is given by $a_{\rm CVE}(\lambda)$, depicted by the green line.}
\Lfig{PD-supernovae}
\end{center}
\end{figure}
The overall shape of this phase diagram is similar to the ones in \Rsec{Phase diagram} or \Rfig{PD-sample}, supporting the practical relevance of our results so far. To directly check the approximation accuracy, we also conducted the literal CV at a number of values of $a$. The results for $a=4$ and $50$ are given in \Rfig{CVE-supernovae}.
\begin{figure}[htbp]
\begin{center}
\includegraphics[width=0.48\columnwidth]{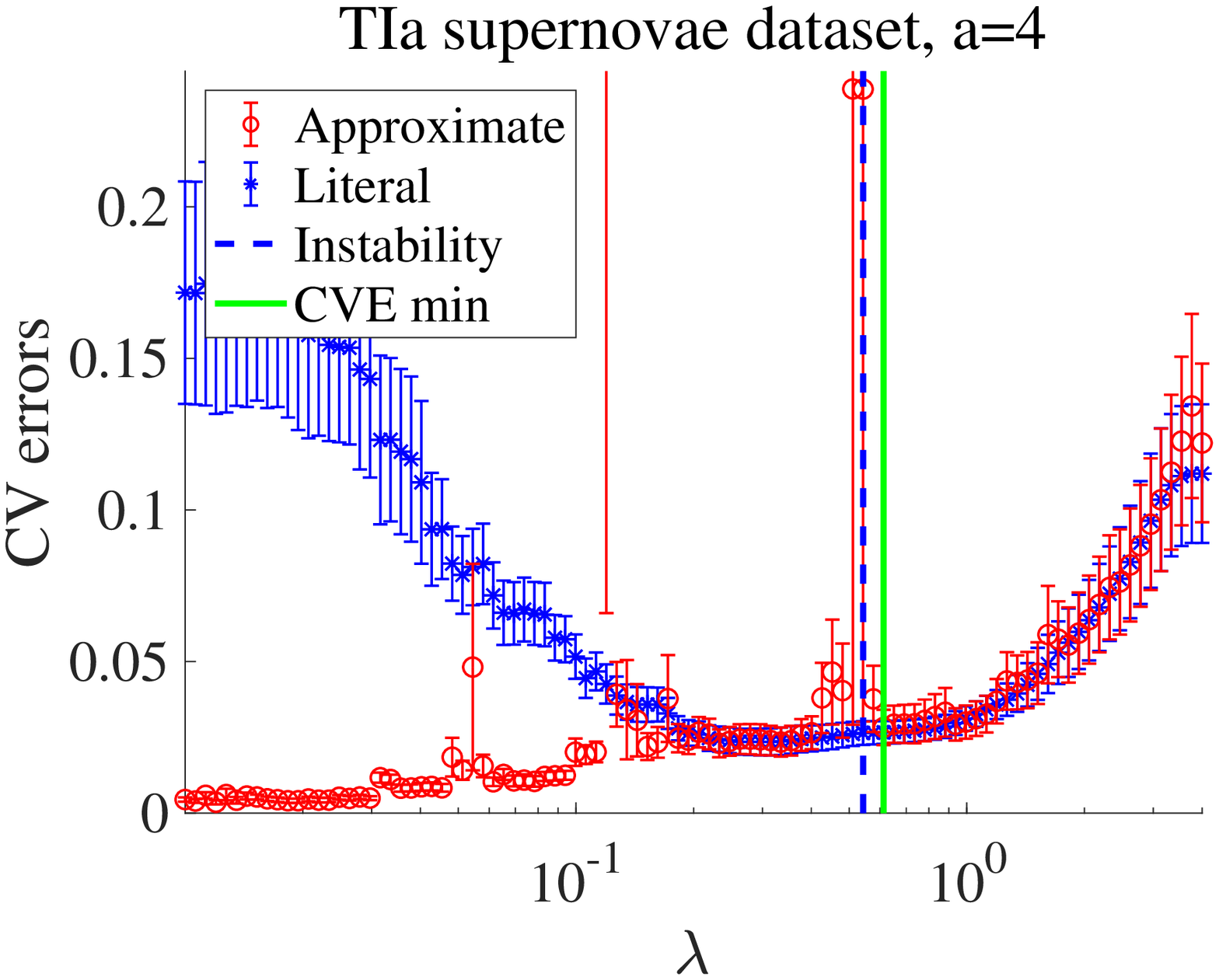}
\includegraphics[width=0.48\columnwidth]{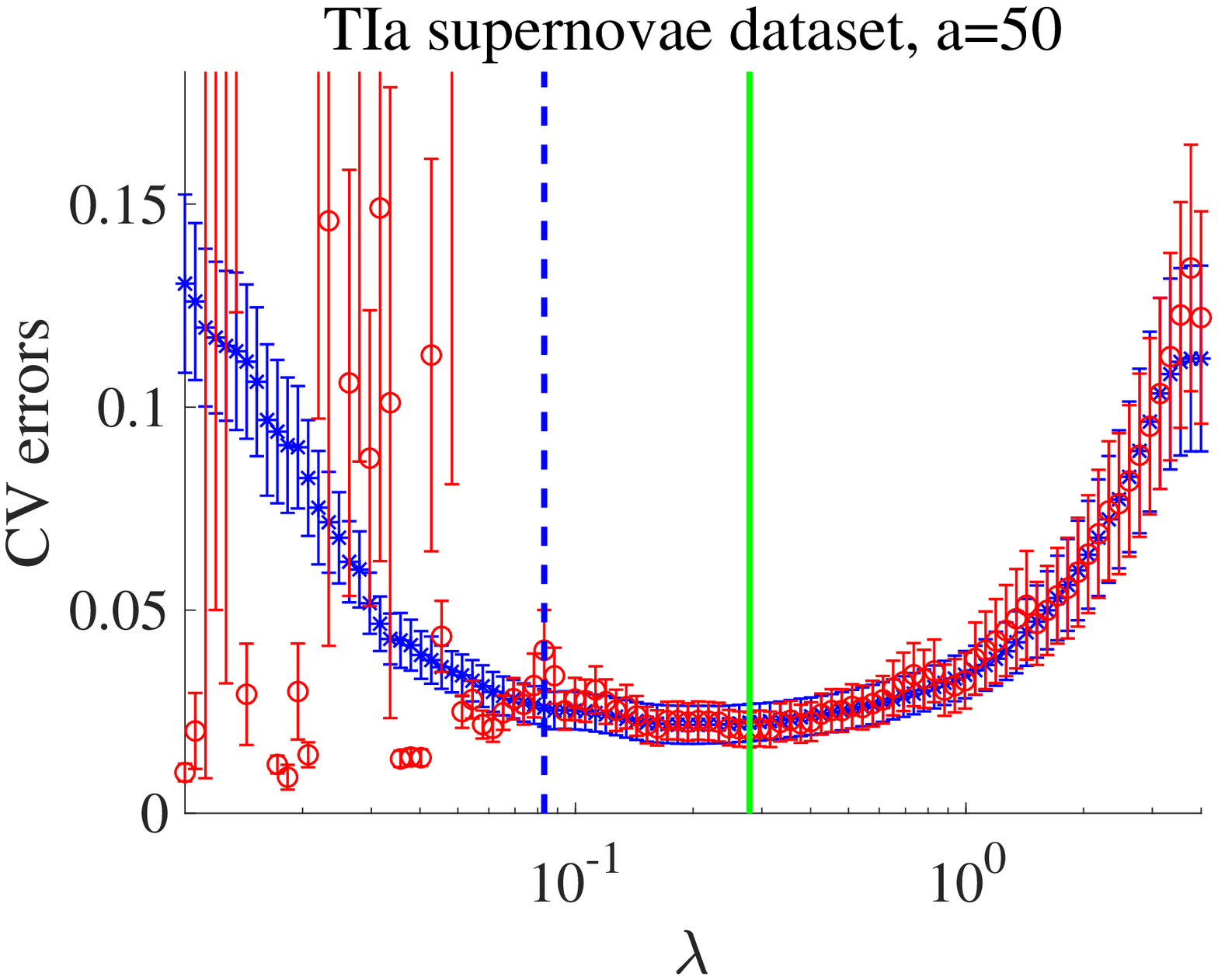}
\caption{Plot of the CV errors by the approximate (red circle) and the literal (blue asterisk) CV for $a=4$ (left) and $a=50$ (right) against $\lambda$ for the Type Ia supernovae dataset. The blue dashed vertical line indicates the instability point obtained by the procedures described in the main text and well matches to the point at which the literal and approximate CV errors deviate from each other.}
\Lfig{CVE-supernovae}
\end{center}
\end{figure}
The approximate error well matches to the literal one up to the instability point, determined by the procedures explained in \Rsec{Accuracy of the}, which justifies our instability detection procedures. For the left panel of $a=4$, however, even below the instability point, there exists a region in which two CV errors agree well. This implies the presence of re-entrant transition, and it is probably related to a protruding black region around $a\in (3,5)$ in the right panel of \Rfig{PD-supernovae}. As declared in \Rsec{Accuracy of the}, we do not try to detect the re-entrancy in the present study, but there must be some ways. For example, the annealing with respect to $a$ instead of $\lambda$ would be able to identify the re-entrancy with respect to $\lambda$. We found that this strategy can actually detect the re-entrancy, but the strategy itself is far from perfect. There are some reasons for this. One reason is that the switching parameter $a$ has no upper bound in contrast to $\lambda$ ($\lambda$ has an effective upper bound as explained in \Rsec{Numerical codes}) and hence the initialization becomes nontrivial. Another reason is that some instability ``islands'' seem to exist at unexpected regions on the parameter space for this specific dataset: Some compact parameter regions exhibiting the instability seem to be able to exist, in contrast to the theoretically derived phase diagrams in \Rsec{Phase diagram}, and hence isolating the instability regions becomes nontrivial even if the annealing with respect to $a$ is correctly performed. Due to these difficulties, we leave further exploration of better ways of nonconvexity control as a future work.

To extract relevant values of the parameters, we plot the approximate CV error and the number of non-zero components $K=||\hat{\V{x}}||_0$ along the $a_{\rm CVE}$ line in \Rfig{On_a_CVE}. Here, some outliers exhibiting extraordinary small CV errors are omitted.
\begin{figure}[htbp]
\begin{center}
\includegraphics[width=0.48\columnwidth]{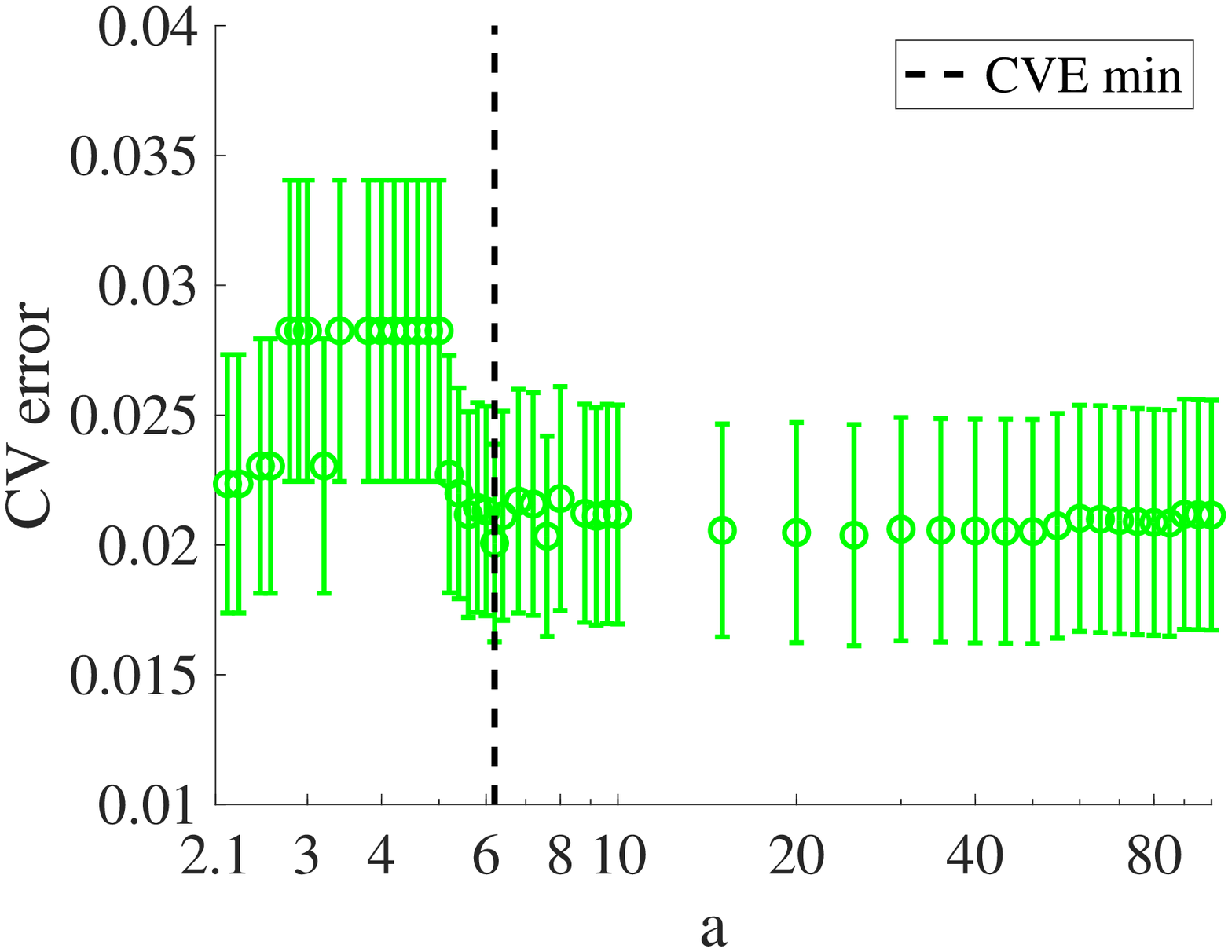}
\includegraphics[width=0.48\columnwidth]{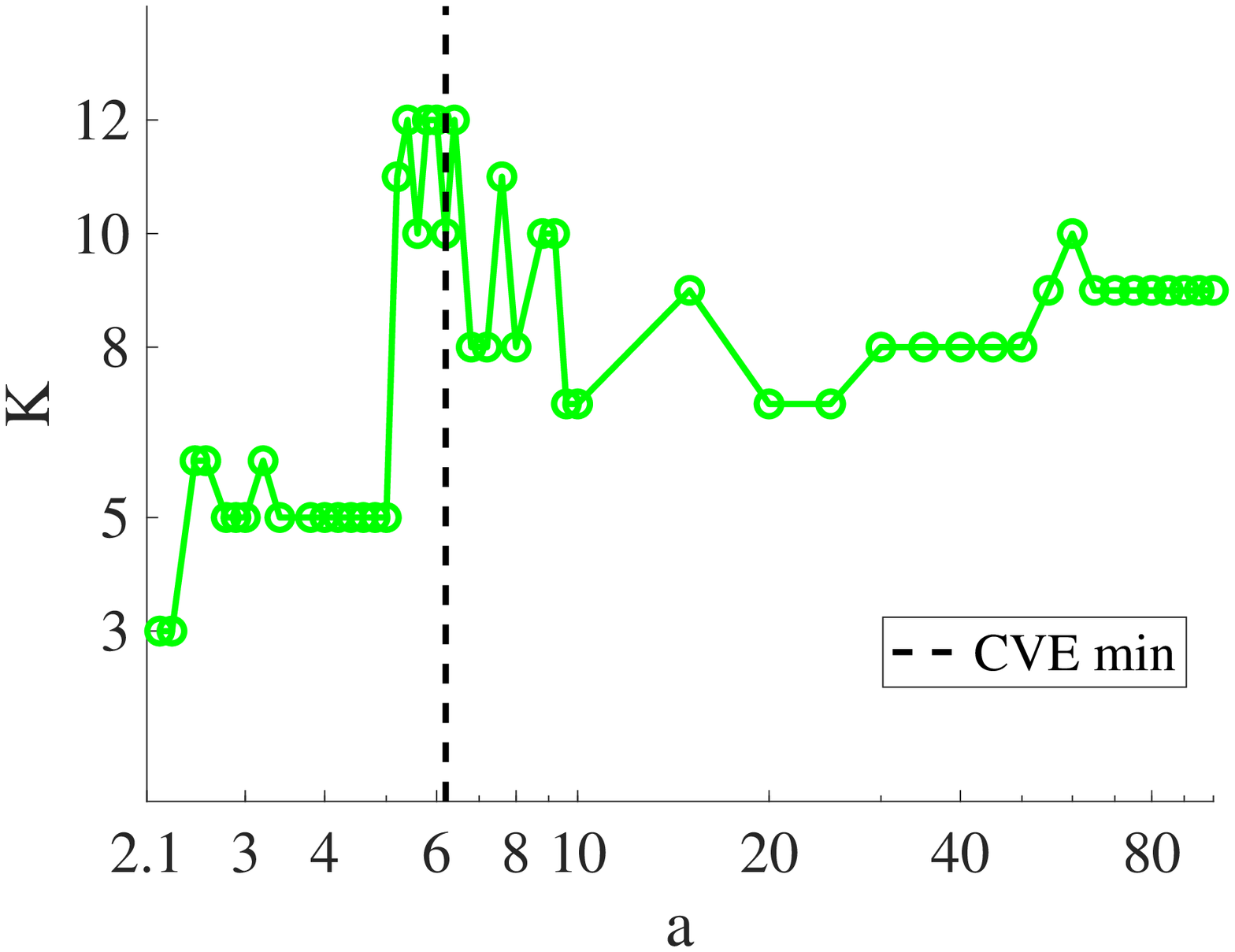}
\caption{
Plots of the approximate CV error (left) and the number of non-zero components (right) along the $a_{\rm CVE}$ line for the Type Ia supernovae dataset. The black vertical dashed line represents the minimum location of the CV error. Almost all datapoints are within the error bar of the minimum error point, and the sparsest solution within the one-standard error is obtained at $a=2.2$ and $2.3$ with $K=3$.}
\Lfig{On_a_CVE}
\end{center}
\end{figure}
At the CV error minimum, the solution with $K=10$ is obtained, which is comparable with $K=9$ of the LASSO solution at the minimum CV error~\cite{obuchi2016cross,uemura2015variable}. In the case of LASSO, it is common to select a sparser solution than the one at the CV error minimum according to the one-standard error rule~\cite{Hastie:2015:SLS:2834535,john2010elements}. Although it is unclear if the application of this rule to the SCAD estimators is appropriate or not, we here try to apply to our case. As a result, we have the solution with $K=3$ obtained at $a=2.2$ or $2.3$ as seen in \Rfig{On_a_CVE}. We globally examined all the datapoints within the one-standard error in the stable phase, and confirmed that the $K=3$ solution is the sparsest. This sparsest solution consists of variables whose IDs are $1$, $2$ and $233$. This is accurately matching to the result of~\cite{kabashima2016approximate,obuchi2016sampling,igarashi2018exhaustive}, in which the $\ell_0$ formalism is treated, while the LASSO estimator tends to give a denser solution with $K=6$ even under the one-standard error rule~\cite{obuchi2016cross,uemura2015variable}. These demonstrate the effectiveness of the SCAD estimator, and the presented analysis and approximate formula resolve its disadvantages of the multiplicity of solutions and the computational cost in hyperparameter estimation. The effect of the one-standard error rule on the SCAD estimator seems to be also good, though further exemplifications would be needed.

\subsection{Numerical codes}\Lsec{Numerical codes}
In \cite{obuchi:SLRpackage}, a MATLAB package of numerical codes implementing the estimation of the solution path using the $\lambda$ annealing in conjunction with the approximate CV formula is distributed; the optimization is performed by the CD algorithm as the experiments so far.  In the package three regularizations, LASSO, SCAD, and MCP, are treated in a unified manner. All the parameters are tunable in the codes, but the minimally required quantities to run the codes are the data vector $\V{y}$, the design matrix $A$ 
\footnote{In the package, the design matrix is denoted as $X$ and the regression coefficients are given as $\beta$, following the statistics convention.}, 
and the switching parameter $a$. In the default setting, the $L=100$ values of $\lambda$ are chosen as to be a descending order $\lambda_{1}> \lambda_{2}>\cdots >\lambda_{L}$, and the largest is set to be $\lambda_1=\lceil \max_{1 \leq j \leq N}\lb |\V{a}_{j}^{\top} \V{y}|\rb \rceil$ where $\V{a}_{j}$ is the $j$th column vector of $A$, because only the trivial solution $\hat{\V{x}}=0$ exists for $\lambda > \max_{1 \leq j \leq N}\lb |\V{a}_{j}^{\top} \V{y}|\rb$; the smallest is set to be $\lambda_{L}=\epsilon \lambda_1$ with $\epsilon=0.01$ and the intermediate values are given to interpolate these two values by the geometric progression with a constant rate. This way follows that of a commonly-used package {\em glmnet}~\cite{Hastie:2015:SLS:2834535,glmnet}. The $\lambda$ annealing is basically the same as {\em warm starts} explained in~\cite{Hastie:2015:SLS:2834535}, but it has a stronger meaning in the nonconvex penalties because it inevitably picks out a certain solution path as exemplified in the numerical experiments so far. On each point of $\lambda_{k}$, the CD algorithm finds the solution $\hat{\V{x}}(\lambda_{k})$ from the initial condition $\hat{\V{x}}(\lambda_{k-1})$ (for $k=1$ the initial condition is the zero vector), and hence $\hat{\V{x}}(\lambda_{k})$ and $\hat{\V{x}}(\lambda_{k-1})$ are expected to be close each other. In the default setting, after obtaining the whole solution path over $\{\lambda_{k}\}_{k=1}^{L}$, the approximate CV formula is subsequently applied and it is followed by the instability detection routine, yielding the approximate values of CV error and its reliable region. In the package, demonstration codes are also included and some experiments in \Rsec{Simulated dataset} can be easily re-obtained; readers who are interested in the experiments are thus encouraged to try to use them. The details of usage are more explained in \cite{obuchi:SLRpackage}.

\section{Conclusion}\Lsec{Conclusion}
In this study, using the replica method, we analysed the macroscopic properties of the SCAD estimator in the context of the signal reconstruction in the presence of noise, under the assumption that the design matrix is the i.i.d. random matrix. We derived the phase diagrams involving the RSB phase, and showed the superiority of the SCAD estimator to the LASSO one based on ROC curves. We also provided an analytical evidence that the global minimum of the input MSE or the generalisation error is located in the RS phase. Furthermore, we derived an approximate formula for the CV error, although it is applicable only for the RS phase. We implemented procedures detecting the AT instability or the approximation instability, enabling to clarify the applicable limit of the approximate formula and making the formula stand-alone. 

To examine the analytical results, numerical experiments on simulated datasets and a real-world dataset of Type Ia supernovae were conducted. On the simulated datasets, the replica prediction was well reproduced. The accuracy and the computational time of the approximate CV formula were examined, and its effectiveness was demonstrated in a wide range of the system size. For the real-world dataset, the application of the SCAD penalty reproduced the variables known to be empirically important. By using the approximate formula, we could globally search the parameters efficiently, and find that the SCAD estimator can provide a very sparse solution giving a reasonable value of the CV error. This solution is matching to the one of the earlier studies using the $\ell_0$ formulation~\cite{kabashima2016approximate,obuchi2016sampling,igarashi2018exhaustive}, and cannot be found by LASSO. These experiments demonstrate the effectiveness of the SCAD estimator, and the presented analysis and approximate formula resolve its disadvantages of the multiplicity of solutions and the computational cost in hyperparameter estimation.

As an efficient strategy to obtain a solution path, we proposed nonconvexity annealing as a part of nonconvexity control proposed in \cite{sakata2019perfect}, and especially focused on the usage of the annealing with respect to $\lambda$, termed $\lambda$ annealing in this paper. It was shown that this strategy works well also in combination with our approximate CV formula, but it further raised up a question related to RSB. In the RSB phase exhibiting the multiplicity of solutions, what solution is obtained by the annealing? Our numerical experiments showed that the annealed solution tends to give a smaller CV error compared to the solutions computed from random initial conditions, and in this sense the annealing is a nice strategy even in the multiple solution region. A similar observation was obtained in an inference in Gaussian mixture model~\cite{barkai1993scaling,barkai1994statistical}. To make a more accurate and quantitative analysis about these findings, it is needed to construct the full-step RSB solution and to figure out the characterisation of the annealed solution in the ensemble of all the solutions. This will be an interesting future work. 

The present instability detection procedures for the approximate CV formula are rather ad hoc and have some ambiguity, especially in specifying irregular datapoints along the solution path with respect to $\lambda$. This ambiguity is related to which points of $\lambda$ should be sampled when computing the solution path. In the case of LASSO, the change points of active set, usually called knots, can be efficiently computed~\cite{efron2004least}, which provides a clear criterion to the above ambiguity problem. It is expected that a similar technique computing knots for SCAD will be useful for improving the instability detection procedures.  

As a final remark, we mention about the MCP penalty defined by:
\begin{eqnarray}
J_{\rm MCP}(\theta;\eta)&=\left\{\begin{array}{ll}
\lambda|\theta|-\displaystyle\frac{\theta^2}{2a} & (|\theta|\leq a\lambda) \\
\displaystyle\frac{a\lambda^2}{2} & (|\theta|> a\lambda) 
\end{array}
\right.,
\end{eqnarray}
where $\eta=\{\lambda,a\}$. If we use this instead of the SCAD penalty, the effective one-dimensional estimator, \Req{x^*_SCAD} in the SCAD case, is replaced as:
\begin{equation}
x^*(h;\T{Q}^{-1})=V_{\rm MCP}(h;\T{Q}^{-1},\eta)S_{\rm MCP}(h;\T{Q}^{-1},\eta),
\Leq{x^*_MCP}
\end{equation}
where 
\begin{eqnarray}
S_{\mathrm{MCP}}(x;\sigma^2,\eta)&=\left\{\begin{array}{ll}
x-\mathrm{sgn}(x)\lambda & \mathrm{for}~a\lambda\sigma^{-2}\geq|x|>\lambda\\
x & \mathrm{for}~|x|>a\lambda\sigma^{-2}\\
0 & \mathrm{otherwise}
\end{array}
\right., 
\\
V_{\mathrm{MCP}}(x;\sigma^2,\eta)&=\left\{\begin{array}{ll}
(\sigma^{-2}-a^{-1})^{-1} & \mathrm{for}~a\lambda\sigma^{-2}\geq|x|>\lambda\\
\sigma^{-2} & \mathrm{for}~|x|>a\lambda\sigma^{-2}\\
0 & \mathrm{otherwise}
\end{array}
\right..
\end{eqnarray}
Replacing $x^*$ in eqs. \NReq{RS_chi_gen}--\NReq{RS_m_gen} by \Req{x^*_MCP}, we can get EOS for the MCP penalty, and the  AT condition \NReq{AT} can be replaced by the same way. Corresponding to \Req{RS feasible}, the RS existence limit of the MCP case is also given as 
\be
\T{Q}-\frac{1}{a}\geq 0.
\Leq{MCPlimit}
\ee
Using these replacements, it is easy to obtain the result for the MCP case. As far as we searched, the MCP result is qualitatively similar to the SCAD one. For illustration of this, we give some phase diagrams, $\MSEx$ plots, and plots of literal CV errors with and without $\lambda$ annealing in \Rfig{MCPresult}.
\begin{figure}[htbp]
\begin{center}
\includegraphics[width=0.48\columnwidth]{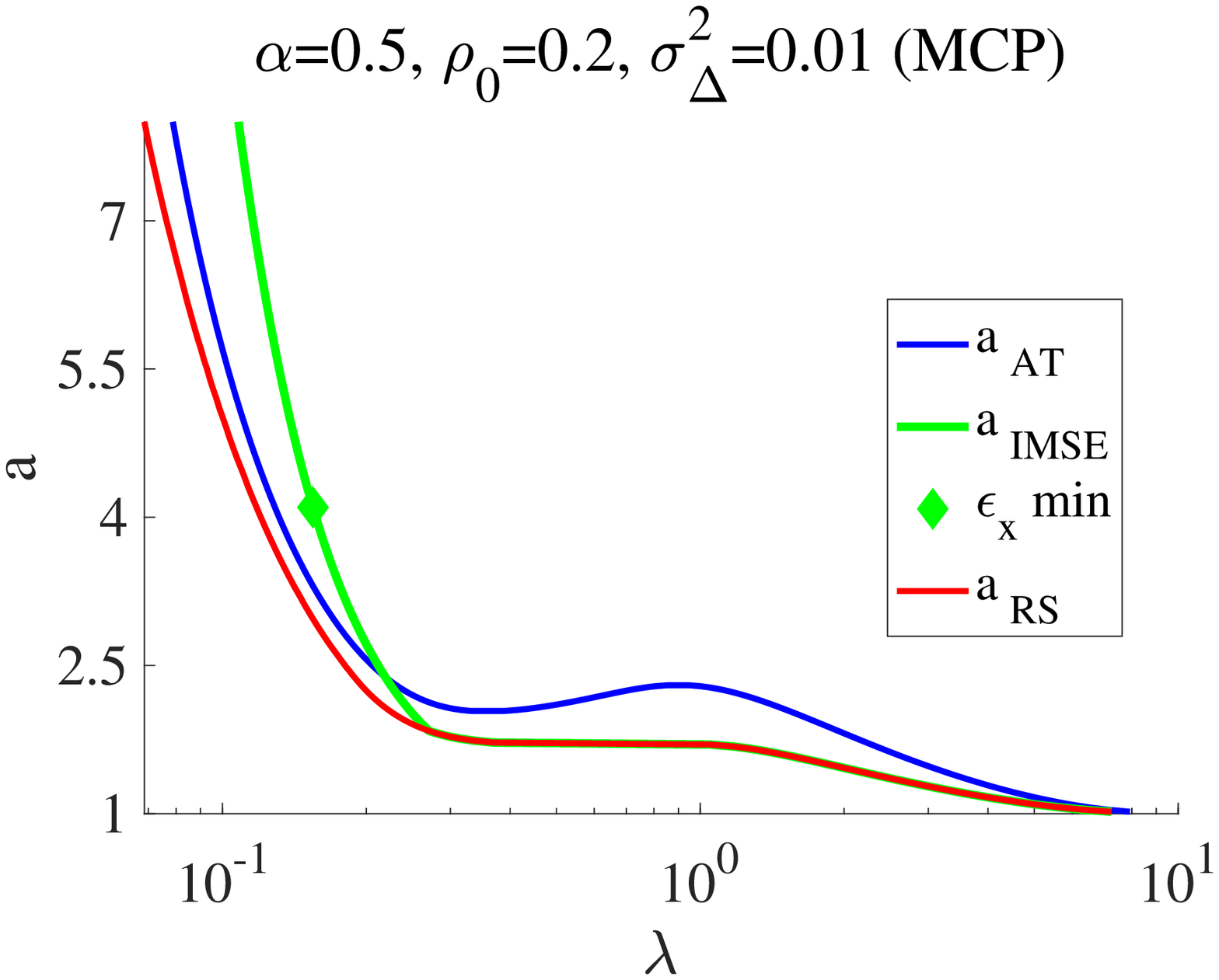}
\includegraphics[width=0.48\columnwidth]{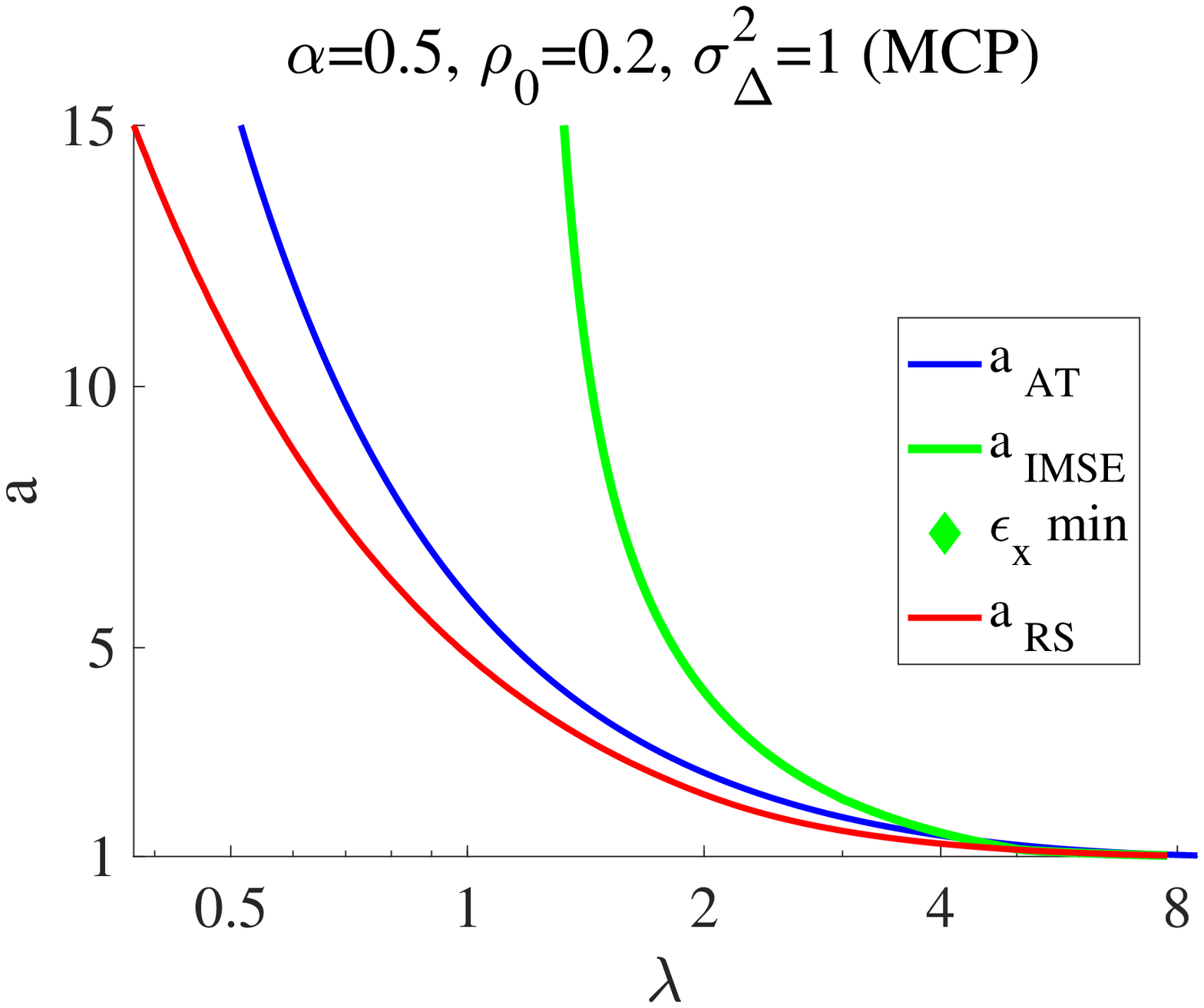}
\includegraphics[width=0.48\columnwidth]{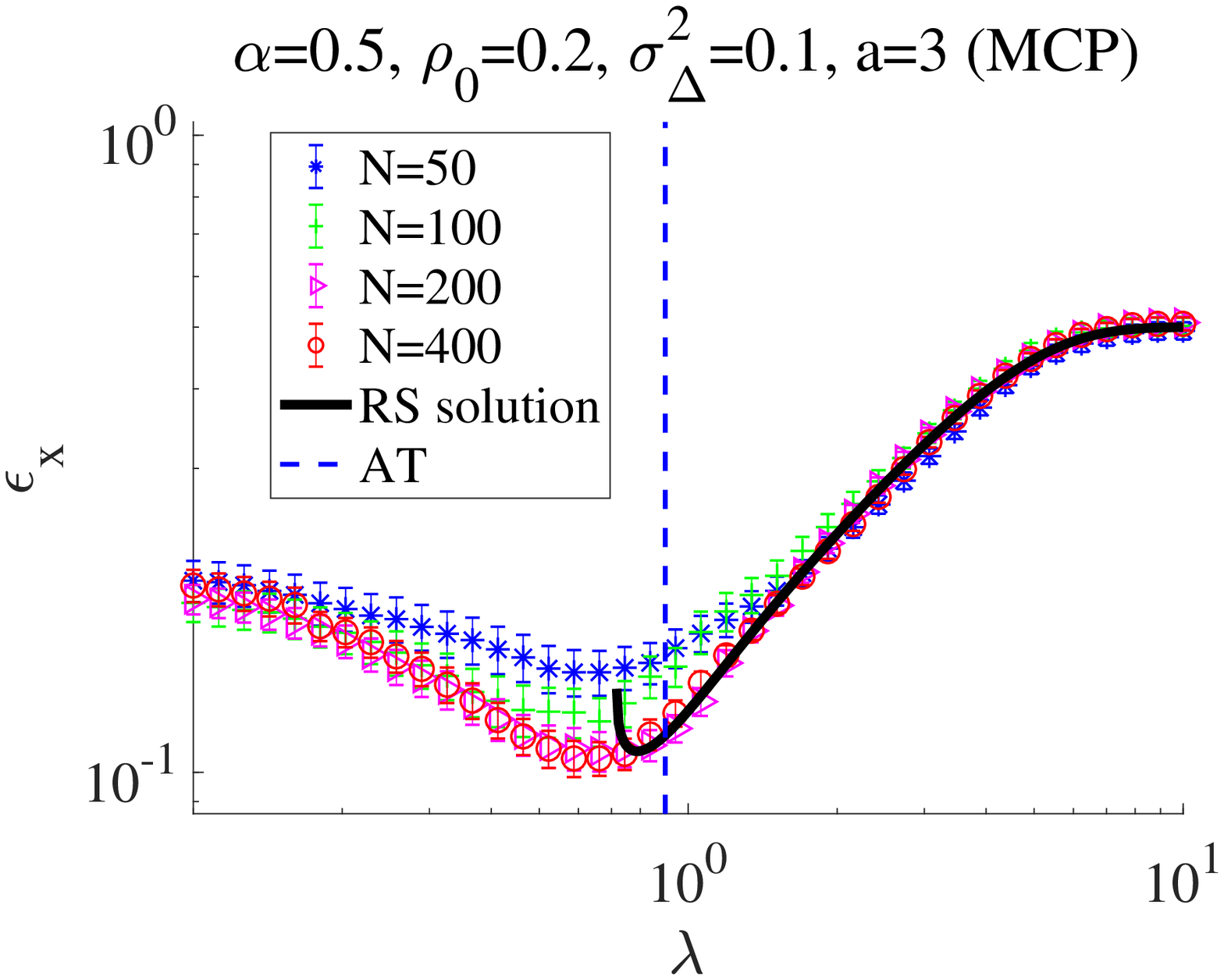}
\includegraphics[width=0.48\columnwidth]{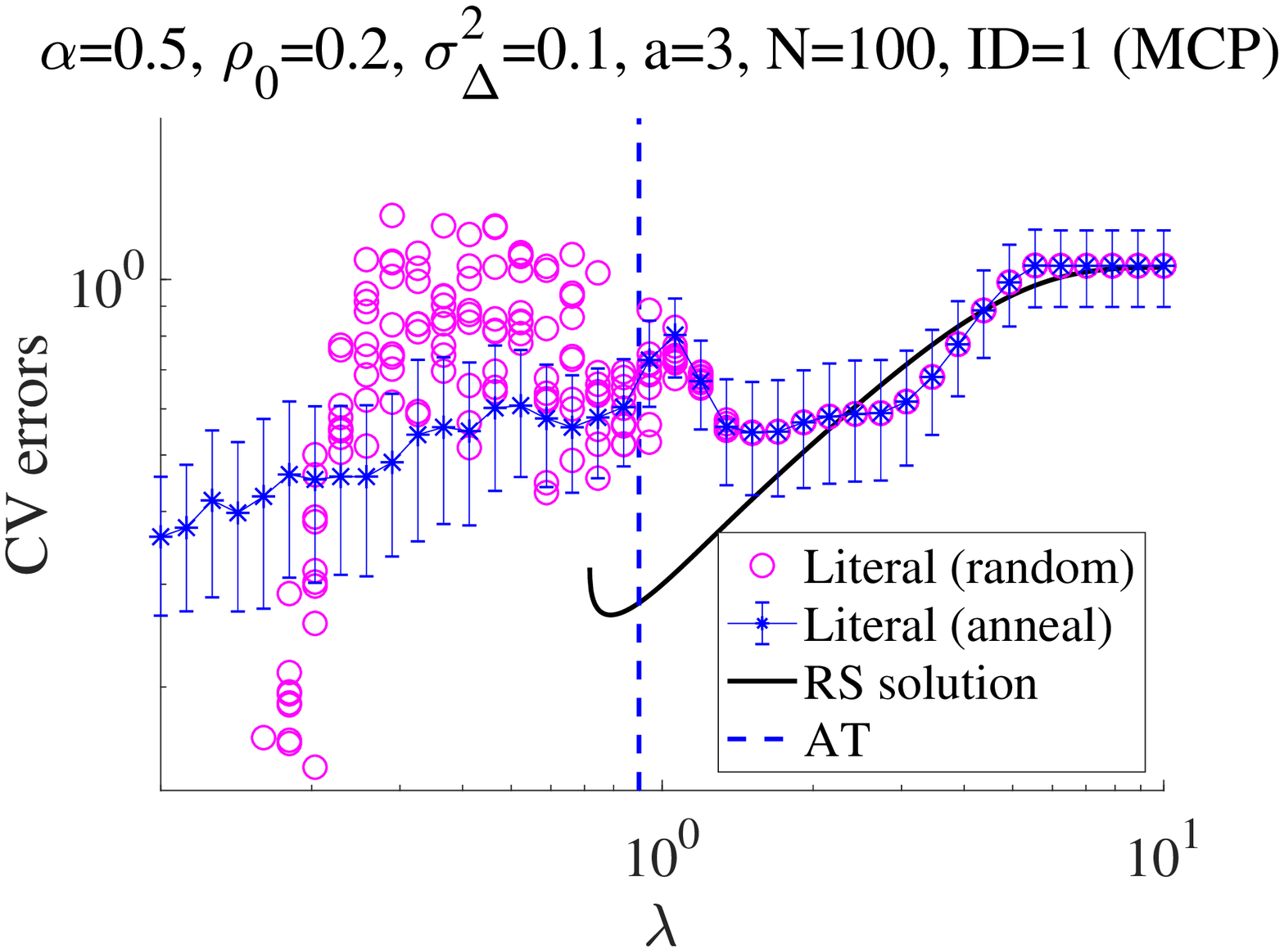}
\caption{
The result for the MCP case. (Upper) Phase diagrams for $\sigma_{\Delta}^2=0.01$ (left) $\sigma_{\Delta}^2=1$ (right) at $(\alpha,\rho_0)=(0.5,0.2)$, corresponding to the middle and right panels of \Rfig{PD-noise}. (Lower) Plots of the input MSE (left) and of the literal CV errors with and without the $\lambda$ annealing (right) against $\lambda$ at $(\alpha,\rho_0,\sigma_{\Delta}^2,a)=(0.5,0.2,0.1,3)$, corresponding to the left upper panel of \Rfig{numcheck-eps} and that of \Rfig{CV-random}, respectively.  Qualitatively similar results to the SCAD case are obtained. 
}
\Lfig{MCPresult}
\end{center}
\end{figure}
We see qualitatively similar results to the SCAD case: The re-entrancy for the weak noise region; the no global minimum of the input MSE in the RS phase at finite $a$ for the strong noise or dense signal cases; the accurate accordance between the RS and numerical results above the AT point; the solution multiplicity below the AT point. Although there can be a difference between the SCAD and MCP penalties in a quantitative level as reported in \cite{Breheny2011}, such a comparative study requires more detailed quantitative analyses and we also leave it as a future work. Note that the lower existence limit of the RS phase of the MCP case is given as $a=1$, which is derived from \Reqs{Qleq1}{MCPlimit}, and hence the RS stable region tends to be wider than the SCAD case. However, the direct comparison of two parameter spaces is not necessarily meaningful, and another systematic way of comparison is desired. 

\ack
The authors would like to thank Yoshiyuki Kabashima, Satoshi Takabe, Takashi Takahashi, and Yingying Xu for their helpful discussions and comments. This work is partially supported by JSPS KAKENHI No. 16K16131 (AS) and Nos. 18K11463 and 17H00764 (TO). TO is also supported by a Grant for Basic Science Research Projects from the Sumitomo Foundation.


\section*{References}
\bibliography{SCADreference}

\end{document}